\documentclass[journal]{IEEEtran/IEEEtran}

\usepackage{tikz}
\usetikzlibrary{backgrounds,positioning}
\usepackage{nrmac}
\usepackage{overpic}
\usepackage{siunitx}
\usepackage[caption=false,font=footnotesize]{subfig}
\usepackage{multirow}
\usepackage{xcolor}
\usepackage{bbm}
\newcommand{\reply}[1]{\textcolor{blue}{{#1}}}
\renewcommand{\reply}[1]{#1}
\newcommand{\id}{\reply{\mathbbm{1}}}
\newcommand{\Q}{\mathcal{Q}}
\newcommand{\X}{\mathcal{X}}

\newcommand{\G}{\mathcal{G}}

\newcommand{\Rq}{\R^{n}}
\newcommand{\Rx}{\R^{2n}}
\newcommand{\Rc}{\R^{2n+k+1}}
\newcommand{\Rv}{\R^{n_v}}
\newcommand{\Rp}{\R^{n_p}}
\newcommand{\Ru}{\R^{n_u}}
\newcommand{\Rmu}{\R^k}
\newcommand{\Rh}{\R^{n_h}}
\newcommand{\Ra}{\R^{n_a}}
\newcommand{\nc}{{2n+k+1}}
\newcommand{\npv}{{{n_p}+{n_v}}}
\newcommand{\npu}{{{n_p}+{n_u}}}

\newcommand{\xtu}{(x_0, \tau, \mu)}
\newcommandx{\flow}[3][1=\tau, 2=, 3=(x_0)]{#2{\varphi}^{#1}_{\mu}#3}
\newcommandx{\Flow}[1][1=\tau]{\flow[#1][][]}
\newcommand{\fun}[3]{#1: #2 \to #3}
\newcommand{\pre}[1]{#1^{-1}(0)}

\DeclareMathOperator{\flip}{flip}

\DeclareMathOperator{\dist}{dist}

\newcommand{\Space}{\mathcal{S}}
\DeclareMathOperator{\Null}{Null}

\newcommand{\ceq}{c_\text{eq}}
\newcommand{\xeq}{x_\text{eq}}
\newcommand{\xeqtu}{(\xeq, \tau, \mu)}
\newcommand{\qeq}{q_\text{eq}}
\newcommand{\Gm}{\G_{\text{mapped}}}
\newcommand{\Mu}{\mathcal{M}}
\newcommandx{\Bez}[1][1=\ ]{B\'{e}zier polynomial#1}
\newcommandx{\Bezs}[1][1=\ ]{B\'{e}zier polynomials#1}

\begin{document}

\title{A Topological Approach to Gait Generation for Biped Robots}
%\title{Using Equilibria to Compute an Equivalency Class of Bipedal Gaits}

\author{Nelson Rosa Jr. and Kevin M. Lynch, \emph{Fellow, IEEE}%
\thanks{N. Rosa and K.~M. Lynch are with the Department of Mechanical Engineering and the Center for Robotics and Biosystems; K.~M. Lynch
is also with the Northwestern Institute on Complex Systems, Northwestern University, Evanston, IL 60208.
{\tt\footnotesize \{nr\} at u.northwestern.edu, \{kmlynch\} at
northwestern.edu}}
\thanks{This work was supported by NSF grants 
IIS-0964665, IIS-1018167, and CMMI-1436297.}%
}

%\markboth{IEEE Transactionis on Robotics,~Vol.~14, No.~8, August~2015}% {Shell
%\MakeLowercase{\textit{et al.}}: A Topological Approach to Gait Generation for
%Biped Robots}

%%%%%%%%%%%%%%%%%%%%%%%%%%%%%%%%%%%%%%%%%%%%%%%%%%%%%%%%%%%%%%%%%%%%%%%%%%%%%%%%

\maketitle

%%%%%%%%%%%%%%%%%%%%%%%%%%%%%%%%%%%%%%%%%%%%%%%%%%%%%%%%%%%%%%%%%%%%%%%%%%%%%%%%
\begin{abstract}
This paper describes a topological approach to generating families of open- and closed-loop walking gaits for underactuated 2D and 3D biped walkers subject to configuration inequality constraints, physical holonomic constraints (e.g., \reply{closed-loop linkages}), and virtual holonomic constraints (user-defined constraints enforced through feedback control).   Our method constructs implicitly-defined
manifolds of feasible periodic gaits within a state-time-control space that parameterizes the biped's hybrid trajectories.  Since equilibrium configurations of the biped often belong to such manifolds, we use equilibria as ``templates'' from which to grow the gait families.  Equilibria are reliable seeds for the construction of gait families, eliminating the need for random, intuited, or
bio-inspired initial guesses at feasible trajectories in an optimization framework.  We demonstrate the approach on several 2D and 3D biped walkers.
\end{abstract}

%%%%%%%%%%%%%%%%%%%%%%%%%%%%%%%%%%%%%%%% Introduction
\section{Introduction}
A challenging problem in bipedal locomotion is the gait-generation problem:
given a model of a bipedal robot, generate periodic gaits subject to the biped's
hybrid dynamics and other constraints.  We present an approach to the
gait-generation problem where equilibria of the biped are used as templates to
find families of gaits.  Under certain conditions, these equilibria can be
continuously deformed into sets of walking gaits, including passive dynamic
walking gaits (unactuated gaits where a biped walks downhill under the influence
of gravity) and actuated gaits where the biped walks on
flat ground or uphill.

In this paper, we assume the biped is physically symmetric about its sagittal
plane, and we are interested in symmetric \emph{period-one} gaits:  periodic
gaits where each step by the right leg is identical and the mirror image of
steps by the left leg.\footnote{The approach can be extended to general
period-$n$ gaits, such as limping gaits~\cite{Gregg2012limping}, but in this paper we focus on
period-one gaits for simplicity.}  \reply{The foot-ground collisions at the end of a step are modeled as plastic, and our models do not have a double-support phase~\cite{Goswami1998, Garcia1998, Westervelt2007}}.

Given the hybrid dynamics of the biped, the
entire trajectory of a single step is represented by the finite-dimensional
tuple $c = (x_0,\tau,\mu) \in \Space = \X \times \R \times \Mu \subseteq
\R^{2n+1+k}$, where $x_0 = (q_0,\dot{q}_0) \in \X \subseteq \Rx$ is the initial
state of the biped with configuration $q_0 \in \Q \subseteq \R^n$; $\tau > 0$
is the duration of the step; and $\mu \in \Mu \subseteq \Rmu$ describes design
or control parameters, such as $k$ polynomial coefficients describing
feedback-control-enforced coupling between joints of the biped.  Our goal is to
find points in $\Space$ that correspond to period-one gaits.

To precisely define period-one gaits, we define the flow $\varphi$ such that
$\flow$ is the biped's state after time $\tau$ using the controls $\mu$
beginning from the state $x_0$.  We define the coordinate-flip operator $\flip:
\X \rightarrow \X$ that maps a state of the biped to its symmetric state (i.e.,
the equivalent state when the other leg is taking a step).  The flip operator
satisfies $\flip(\flip(x_0)) = x_0$.  With these definitions, a point $c = (x_0,
\tau,\mu) \in \Space$ corresponds to a period-one gait if and only if $\flow -
\flip(x_0) = 0$.

Said another way, the \emph{periodicity map} $P:\Space \rightarrow \X$ is
defined as
\[
P(c) = \flow - \flip(x_0),
\]
and the set of all period-one gaits, denoted $\G$, is the set of all points $c
\in \Space$ satisfying $P(c) = 0$, i.e., $\G = P^{-1}(0)$.  Since $P(c)=0$
specifies $2n$ constraints on the $(2n+k+1)$-dimensional space $\Space$, in
general we would expect $\G$ to be $(k+1)$-dimensional
($\operatorname{dim}(\Space) - \operatorname{dim}(P)$).

The goal of our work is not to find a single period-one gait (a single point in
$\G$), but to map out a ``large'' continuous family of gaits $\Gm \subset \G
\subset \Space$.  The set of gaits $\Gm$ may include walking downhill, walking
uphill, and even hand-to-hand gibbon-like swinging gaits (\emph{brachiation})
underneath a support.  \reply{We construct these gait families through the continuous deformation of reference gaits in $\G$.  A continuous deformation from one gait to another defines a homotopy equivalence between two gaits, a type of topological equivalence \cite{Crossley2005}.  In this respect, we consider our framework to be a topological approach to gait generation because we only consider the connectivity properties of gaits to each other in $\G$.  This is in contrast to other approaches which focus on other aspects such as the biped's dynamics \cite{Goswami1999, Bessonnet2005, Westervelt2007} to help simplify the search for periodic motions.}  A long-term goal is a full topological description of
$\G$ for a given state-time-control space $\Space$, but that is beyond the scope of this paper.

\begin{comment}
We further restrict ourselves to the set of homotopy maps of $\G$ onto itself.  In other words, if we were to stop the deformation partway, the resulting trajectory would still be periodic.  Stated differently, any small perturbation to a gait results in a gait.  
\end{comment}

The standard approach to finding a single gait in $\G$ is to formulate a
non-convex optimization problem (OP) in the parameters $x_0$, $\tau$, and $\mu$.
The convergence of non-convex OPs relies critically on the initial seed
value~\cite{Grizzle2014, Xi2015}, which is typically chosen randomly or by
applying domain-specific knowledge~\cite{Chen2007, Hereid2018, Posa2016,
Xi2015}.   No general guidelines exist for generic $n$-degree-of-freedom bipeds.

In our framework, however, any one-footed rest state $\xeq = (\qeq,0)$ which is
also an equilibrium (i.e., $\varphi_\mu^t(\xeq) = \xeq$ for some $\mu$ and all
$t \geq 0$) is trivially an ``equilibrium gait'' $\ceq = (\xeq,\tau,\mu)$.  \reply{The virtual ``impact'' after time $\tau$ does not change the state of the biped, as the biped is at equilibrium, so the ``hybrid'' motion is trivially periodic.\footnote{In this paper, the impact time for a gait is given explicitly by the \emph{independent} variable $\tau$; the slope of the ground is determined by $\tau$ (and $x_0$ and $\mu$), not vice-versa.}}
%An equilibrium gait may be unstable.}
%Unlike equilibria of a system with continuous dynamics, a collision must occur in order for the stationary point to satisfy the biped's hybrid dynamics \cite{Bainov1993} (i.e., the step duration $\tau$ must be finite).}

%\reply{Given the equilibrium gaits of a biped}, a subset of these equilibrium gaits are period-one gaits satisfying $P(\ceq) = 0$.  
The set of all such trivial, non-locomoting equilibrium period-one gaits is
denoted $E$, a subset of $\G$.  
An equilibrium gait $\ceq$ is often in the same connected component of $\G$ as useful locomoting gaits, and this motivates the use of numerical continuation methods (NCMs) to generate this connected component starting from $\ceq$.  In particular, branches of locomoting gaits intersect an equilibrium branch containing $\ceq$ on the connected component
at critical values of the step duration and fixed values of $(\xeq,\mu)$ where the rank of the Jacobian of $P$ at these values is not maximal.  In other words, the easy-to-find equilibria are seeds, or
``templates,'' which are continuously deformed to generate $\Gm$.

% ; a collision must occur in order for the motion to satisfy the biped's hybrid dynamics \cite{Bainov1993} (i.e., the step duration $\tau$ must be finite)

Having a continuous family of gaits $\Gm$, instead of one or a small number of
gaits, can be useful in a number of ways.  First, some high-level walking motion
planners rely on low-level gait-generation modules, or a pre-computed library of
gaits, that can be applied on different terrains~\cite{Gregg2012, Motahar2016,
Liu2012a,Saglam2014}.  A gait family $\Gm$ constructed using our approach is a
continuous version of a gait library.  Second, a gait family $\Gm$ allows the
possibility of design of control laws that drive the biped to $\Gm$ rather than
to a single specific gait $c \in \G$.  In general, it is easier to design a
controller to stabilize a manifold than to stabilize a point.  Most importantly,
$\Gm$ provides a global view of the possible gaits of a biped robot for the
given space of design and control parameters $\Mu$.

%%%%%%%%%%%%%%%%%%%%%%%%%%%%%%%%%%%%%%%% Introduction - Statement of Contrib
\subsection{Statement of Contributions}
\label{ssec:soc}
This paper describes a topological approach to generating families of walking
gaits for 2D and 3D underactuated biped walkers with point, curved, or flat feet
that are physically symmetric about their sagittal plane.  We use NCMs to map
out connected components of gaits in a state-time-control space $\Space$.  The
biped may be subject to configuration inequality constraints, physical holonomic
constraints (PHCs) such as closed chains, and virtual holonomic constraints
(VHCs), i.e., user-defined constraints enforced through feedback control.  Our
main contributions are:
\begin{enumerate}
\item \textbf{A topological approach to the gait-generation problem.}  We view
gaits as points in a space $\Space$ of parameterized trajectories, where we
\reply{characterize} a fundamental property of the periodic orbits of a biped's hybrid dynamics: their connectivity to each other in $\Space$ across variations in state, step duration, and design and control parameters.  \reply{We take advantage of this connectivity to design algorithms to numerically construct families of gaits in $\Space$.}

\item \textbf{The use of equilibria to generate a continuum of walking gaits.}
We prove that we can find families of locomoting gaits that transversally
intersect a family of equilibrium gaits in $E$ at points $\ceq = \xeqtu$ for a
given fixed pair $(\xeq, \mu)$.  We provide an algorithm for determining the
values of $\tau$ where the intersections occur.

%\item \reply{\textbf{A framework that can determine if it can find gaits.} We can determine up to numerical precision and in a user-specified finite number of steps whether our framework will find walking gaits from equilibria using the same algorithm for finding branches of locomoting gaits prior to applying our NCMs.  This is in contrast with other frameworks, which report a successful or failed attempt after searching the space $\Space$ for a gait.}

\item \textbf{A framework for generating open-loop periodic motions that satisfy the full hybrid dynamics.}  \reply{We provide a systematic approach using known seed values and a conceptual model of the solution space to the challenging} problem \cite{Grizzle2014} of generating open-loop periodic motions for the unactuated joints of a biped robot subject to PHCs and VHCs, including when all joints are unactuated (passive dynamic walking) and when a subset of joints track parameterized trajectories. 
\end{enumerate}

This paper builds on our conference paper~\cite{Rosa2014a} and the
abstract~\cite{Rosa2017}.  In this previous work, we introduced the concept of
using NCMs to generate gaits for bipeds, including those subject to virtual holonomic constraints.  This paper extends
our preliminary work in several important ways:  1) we provide a unified
framework for generating gaits from equilibrium templates for 2D and 3D
underactuated bipeds subect to configuration inequality constraints and physical
and virtual holonomic constraints; 2) we provide a new algorithm to find
specific types of gaits with desired properties \reply{(e.g., a gait that walks on level ground)}; and 3) we provide applications
of the framework to finding gaits for simulated complex 3D bipeds such as Atlas and MARLO.

%%%%%%%%%%%%%%%%%%%%%%%%%%%%%%%%%%%%%%%% Introduction - Related Work

\subsection{Related Work}

Equilibria and numerical continuation methods have a strong history in generating gaits for unactuated biped walkers and brachiators \cite{Gomes2005a}.  In particular, the use of equilibria for generating families of unactuated walking gaits can be
found in past works studying simple two- and three-degree-of-freedom passive
dynamic walking biped models \cite{McGeer1990a, Goswami1998, Garcia1998}.  The
solution families of walking gaits for these biped models converge to an
equilibrium gait because the solution families exhibit a vanishing step size---as the biped's walking slope approaches flat ground, the step size of the
corresponding gait becomes shorter.  In the limit, as the incline approaches
level ground, the state of the biped must approach an equilibrium gait
\cite{Chatterjee2000}, a ``gait'' with zero step size.  The work in
\cite{Garcia2000} explores this notion of finding periodic walking motions near
equilibria for simple walking models with vanishing step sizes.  The paper gives
necessary conditions on the physical parameters of planar two- and three-link
bipeds for walking at arbitrarily small but near-zero slopes.

We extend the work on unactuated, low-dimensional, planar bipeds with vanishing
step sizes to include powered high-degree-of-freedom 2D and 3D bipeds.  In
our previous work \cite{Rosa2012, Rosa2013, Rosa2014a}, we used
NCMs~\cite{Krauskopf2007} to generate families of open-loop walking and
brachiating gaits that utilize the ``natural'' or full dynamics of the biped
model.  In particular, \cite{Rosa2014a} demonstrates that equilibria of
representative point-feet planar bipeds can be continuously deformed into
families of passive dynamic walking gaits.  We extend this body of work to
include closed-loop gaits for underactuated bipeds using the hybrid zero
dynamics (HZD) framework \cite{Westervelt2007, Griffin2015, Hamed2016}.

The HZD framework is an experimentally-validated approach
to generating stable walking
gaits for underactuated bipedal robots subject to virtual constraints
(constraints on the biped that are imposed using feedback control) \cite{Chevallereau2003, Ramezani2013, Hereid2018}.  The notion
of virtual constraints, in particular virtual \textit{holonomic} constraints, has been a useful concept in the design and control of bipedal walking
gaits.  We enforce VHCs using an HZD controller, which can provably impose
the constraints under mild conditions \cite{Westervelt2007}.  Alternative
control schemes for enforcing a set of VHCs also exist \cite{Saglam2015}.

A common application of VHCs on a bipedal system is to couple the motion of a subset of joints
on an underactuated robot so that they evolve with respect
to a function of the biped's configuration as opposed to time.  The resulting
motion is then synchronized to, for example, the motion of a biped's center of
mass projected onto its tranverse plane when the constraints are properly
enforced through feedback control.  The net effect is that the biped's joints move only
if the center of mass moves, irrespective of time.  In such a case, the
motions are said to be self-clocking \cite{Chevallereau2003}.

\begin{comment}
A common application of VHCs on a bipedal system is to have a subset of joints
on an underactuated robot track polynomial trajectories that evolve with respect
to a function of the biped's configuration as opposed to time.  The resulting
motion is then synchronized to, for example, the motion of biped's center of
mass projected onto its tranverse plane when the constraints are properly
enforced through feedback control.  The net effect is that the biped's joints move only
if the center of mass moves, irrespective of time.  In such a case, the
trajectories are said to be self-clocking \cite{Chevallereau2003}.
\end{comment}

Given a biped subject to physical and virtual holonomic constraints, we generate
gaits using NCMs, which originate from results in topology and differential
geometry \cite{Rheinboldt2000}.  In this context, our application is similar to
tracing the points on a differentiable manifold (e.g., a curve or a
higher-dimensional surface) represented as a set of equations that are
continuously differentiable.  Applications of continuation methods for
generating dynamic motions can be found in \cite{Liu2012, Gan2018}.

%An application of continuation methods for generating dynamic motion sequences
%of animated individuals executing parkour maneveurs can be found in
%\cite{Liu2012}.  The parkour moves are based on a human motion database.

NCMs are also present in optimization solvers, which many gait-generation
libraries rely on to generate gaits.  NCMs are typically used to find feasible
solutions (e.g., elastic mode in SNOPT \cite{Gill2002}) or to solve a series of
related optimization problems (e.g., interior-point methods \cite{Betts2010},
like IPOPT).

The standard approach to solving the gait generation problem is to formulate it
as an optimization problem (OP) \cite{Hereid2018, Posa2016, Bessonnet2005}.  The
idea is to specify the decision variables, constraints, and objective function
used in the optimization in such a way that the underlying solver (often SNOPT,
IPOPT, or fmincon) can quickly and robustly converge from an arbitrary seed
value \cite{Hereid2018, Posa2016, Xi2015}.  Recent approaches use direct
collocation methods as part of the problem formulation, where the biped's
equations of motion are discretized into a set of algebraic constraints using a
low-order implicit Runge-Kutta scheme with fixed step size.  A comparable
optimization-based framework to our work is \cite{Hereid2018}.  In
\cite{Hereid2018}, direct collocation methods are used to generate gaits for
bipeds subject to VHCs using an HZD feedback controller to enforce the VHCs.

Our use of NCMs to find gaits differs from methods in the literature that rely
on OPs in that these works attempt to find the ``best'' gait while we use NCMs
to find many gaits without having to guess an initial seed value.   \reply{However, with some effort it is possible to modify OPs to generate a continuum of gaits and NCMs to find optimal gaits.}

%%%%%%%%%%%%%%%%%%%%%%%%%%%%%%%%%%%%%%%% Introduction - Paper Outline

\subsection{Paper Outline}
After covering mathematical preliminaries in Section~\ref{sec:mp}, we describe
the gait space $\G$ and how to generate gaits from equilibria using NCMs in
Sections~\ref{sec:top}--\ref{sec:ext}.  In Section~\ref{sec:ex}, we give
examples of generating gaits for the planar compass-gait walker and the 3D
bipeds Atlas and MARLO.  In Section~\ref{sec:frost}, we compare our approach to FROST \cite{Hereid2018}.
%In Section~\ref{sec:conc}, we conclude the paper.

We also provide downloadable material~\cite{Rosa2020} consisting of
\begin{enumerate}
    \item an MP4 video of walking animations of all biped models used in this paper,
    \item a Mathematica \reply{v11.3.0} library of our framework and implementation details of the models, and
    \item a Node.js v12.17.0 visualization library for animating and creating video clips of the gaits.
\end{enumerate}

\begin{comment}
% For use in TRO submission
We also have supplementary downloadable material available at http://ieeexplore.ieee.org and GITHUB. The core material consists of
\begin{enumerate}
    \item an MP4 video of walking animations of all biped models used in this paper, and
    \item a Mathematica \reply{v11.3.0} library of our framework and implementation details of the biped models.
\end{enumerate}
The material is 23.3 MB in size.  The library available on GitHub also has a Node.js v12.17.0 visualization library for animating and creating video clips of generated gaits.
\end{comment}

%%%%%%%%%%%%%%%%%%%%%%%%%%%%%%%%%%%%%%%% Mathematical Preliminaries

\section{Preliminaries}
\label{sec:mp}
In this section, we specify the biped's hybrid dynamics, give the problem
statement, state assumptions, and formally define the space of parameterized
trajectories $\Space$, the gait space $\G$, and the connected components of
$\G$.

%%%%%%%%%%%%%%%%%%%%%%%%%%%%%%%%%%%%%%%% Introduction - The Space of Param
\subsection{The Hybrid Dynamics}
The \emph{hybrid dynamics} $\Sigma$ of an $n$-degree-of-freedom biped robot is
the tuple $\Sigma = (\X, f, \Delta, \phi)$, where 
\begin{itemize}
\item $\X$ is the biped's state space;
\item $f(x,u) \in T\X$ describes the continuous dynamics, where
$u \in \Ru$ is the robot controls;
\item $\fun{\Delta}{\X}{\X}$ is a jump map to model instantaneous impacts; and
\item $\fun{\phi}{\R \times \X}{\R}$ is a switching function to indicate when a
foot hits the ground.  If $\phi(t, x) = 0$, then $t \in \R$ is a \emph{switching
time}, $x \in \X$ is a \emph{pre-impact state}, and the foot is in contact with
the ground.
\end{itemize}

The motion of the biped can be subject to $n_p$ physical holonomic constraints and $n_v$ virtual holonomic constraints.  The physical
constraints, due to closed-loop linkages or kinematic constraints between the
foot and the ground, for example, give rise to $n_p$ constraint forces.  The
virtual constraints are enforced using feedback control.  We assume that the
biped has $n_u$ ($n_u \geq n_v$) control inputs $u(t) \in \Ru$ to enforce the
VHCs in the system.

The VHCs specify the configuration of certain degrees of freedom of the biped as
a function of a \emph{phase variable} $\theta$.  In real-time control, the phase
variable is often a function of the biped's state~\cite{Westervelt2007} (e.g.,
the swing leg's joints could be ``clocked'' by the angle from the stance foot to
the hip), but to plan a single step of a gait, time suffices as a phase
variable.

In this paper, a VHC takes the generic form 
\begin{equation}
\label{eq:qi}
q_i(t) - b_i^d(\theta(t), a) = 0, \quad t \in [0,\tau],
\end{equation}
where $\tau$ is the step duration, $\theta(t) = t/\tau \in [0,1]$ is the phase
variable, $q_i(t) \in \R$ is a joint angle ($1 \leq i \leq n$),
$b_i^d(\theta(t), a) \in \R$ is a \Bez of degree $d \in \mathbb{N}$, and $a \in
\Ra$ is a vector of polynomial coefficients.  Appendix~\ref{app:model} provides
further modeling details, including how to enforce a set of VHCs.

% \KL{In the equation below, why is $\mu$ in $f$ a second time (it is already
% there in $\varphi$)?} ---to emphasize f's dependence on \mu.

\subsection{The Space of Parameterized Trajectories}
Given $\Sigma$, we are interested in hybrid trajectories that correspond to a
step of a biped of the form
\begin{equation}
\label{eq:flow}
x(\tau) = \flow = \Delta(x_0; \mu) + \int_0^\tau f(\flow[t], u; \mu) \; dt,
\end{equation}
where $x_0 \in \X$ is the pre-impact state at $t = 0$, 
$\flow[t] \in \X$ is
the state of the robot at time $t$,
$\flow$ is the
next pre-impact state at $t = \tau > 0 \in \R$, 
$\mu \in \Mu$ is a vector of input
parameters, and $u(t) \in \Ru$ is a vector of control inputs that depend on
$\mu$.  The parameters $x_0$, $\tau$, and $\mu$ define the space of
parameterized trajectories.

\begin{myrem}
The input parameters $\mu$ can be used to specify design parameters
of the biped such as the center of mass position of a link, leg length, spring
coefficients, and moment of inertia.  It can also be used to define control
parameters, like feedback gains, magnitude of ankle push-off force, and spline
coefficients.
\end{myrem}

\begin{figure}[t]
\centering
\includegraphics[width=0.45\textwidth]{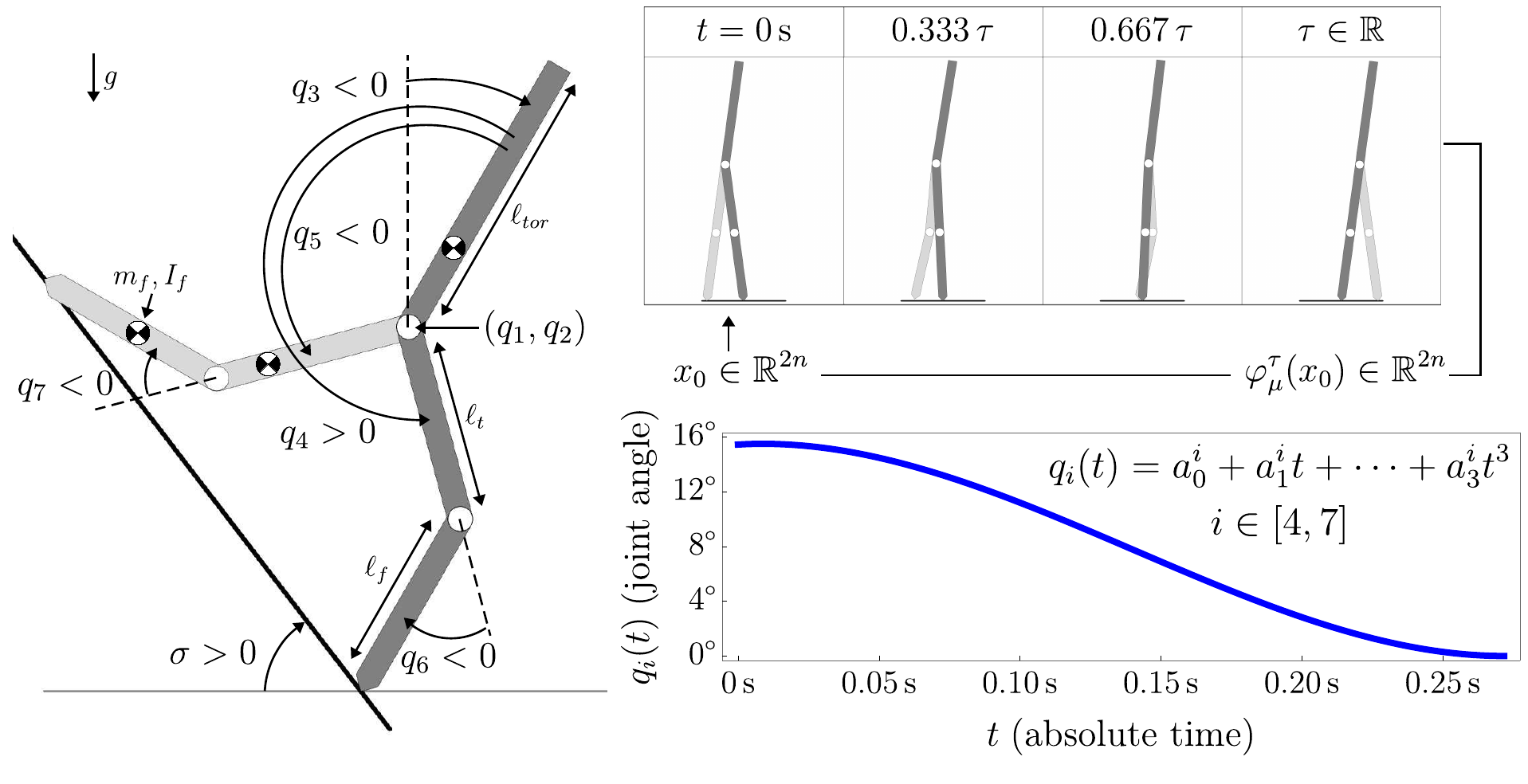}
\caption{A generic $n$-degree-of-freedom biped model (left).  We parameterize
motions that satisfy the model's hybrid dynamics (top right) with a pre-impact
state $x_0 \in \X \subseteq \Rx$, a switching time $\tau \in \R$, and a vector
of input parameters $\mu \in \Mu \subseteq \Rmu$ (top and bottom right).  We use
the vector $\mu$ to represent any parameter that is not a state or switching
time.  In this example, the vector $\mu$ consists of $k$ polynomial coefficients
(bottom right) used to define the trajectory of joint $q_i$ of the biped.}
\label{fig:generic-biped}
\end{figure}

\begin{mydef}
A biped's \emph{state-time-control space} $\Space$ is a finite-dimensional
vector space $\Space \subseteq \X \times \R \times \Mu \reply{\subseteq} \R^{2n+1+k}$.  A
point $c \in \Space$, where $c = \xtu$, defines a hybrid
trajectory $x(t) \in \X \subseteq \R^n$ given input parameters $\mu \in \Mu
\subseteq \R^k$ starting from $x_0 \in \X$ at switching time $t = 0$ until the
next switching time $\tau > 0$.
\end{mydef}
Figure~\ref{fig:generic-biped} shows how the parameters of the
state-time-control space $\Space$ can affect the motion of an
$n$-degree-of-freedom biped robot.
%%%%%%%%%%%%%%%%%%%%%%%%%%%%%%%%%%%%%%%%
%%%%%%%%%%%%%%%%%%%%%%%%%%%%%%%%%%%%%%%%
As defined earlier, the set of all period-one gaits in $\Space$ is defined
using the periodicity map $P$ as
\[
\G = \{c \in \Space \; : \; P(c) = 0\},
\]
i.e., the set of all points $c = (x_0, \tau,\mu)$ satisfying $\flow - \flip(x_0)
= 0$.  The set of equilibrium (stationary) gaits is defined as
\[
E = \{\ceq = \xeqtu \in \G \; : \; f(\xeq, u(t); \mu) = 0 \; \forall t \in \R
\},
\]
i.e., the set of all points $\ceq$ satisfying $P(\ceq) = 0$ and
$\flow[t][][(\xeq)] = \xeq$ for all $t$.

%%%%%%%%%%%%%%%%%%%%%%%%%%%%%%%%%%%%%%%% Mathematical Preliminaries - Connected

\subsection{The Connected Components of the Gait Space $\G$}
\begin{mydef}
Let $\G$ be the space of all gaits in $\Space$.
\begin{enumerate}

\item \label{def:path} A \emph{path} between two points $a$ and $b$ in $\G$ is a
continuous function $\fun{p}{[0, 1]}{\G}$ such that $p(0) = a$ and $p(1) = b$.

\item A set $X \subseteq \G$ is \emph{path-connected} if for all $a, b \in X$,
there exists a path $\fun{p}{[0, 1]}{X}$ with $p(0) = a$ and $p(1) = b$.

\item A set $X \subseteq \G$ is a \emph{connected component} of $\G$ if $X$ is
path-connected and $X$ is maximal with respect to inclusion.

\end{enumerate}
\end{mydef}

\begin{mythm}
\label{thm:impman}
\rm{\cite{Spivak1965}} \hspace*{0.05in} Let $D$ be an open set in $\G$ and the
periodicity map $\fun{P}{\Space}{\R^{2n}}$ be a class $\mathcal{C}^r$
differentiable function.  If for every $c \in D$, the Jacobian $J(c) \in \R^{2n
\times (2n+k+1)}$
\begin{equation} 
\label{eq:J} 
J(c) = \pd{P}{c}(c) = \begin{bmatrix} \pd{\varphi}{x}(c) - \id_{2n}, &
\pd{\varphi}{\tau}(c), & \pd{\varphi}{\mu}(c) \end{bmatrix}
\end{equation} 
has maximal rank $2n$, then $D$ is a $(k+1)$-dimensional ($\mathcal{C}^r$
differentiable) manifold in $\G$.
\end{mythm}

\begin{comment}
\begin{mydef} 
\rm{\cite{Choset2005}} \hspace*{0.05in} The \textit{tangent space} of $\G$ at
$c_0$, represented by $T_{c_0}\G$, is a $d$-dimensional vector space $\R^d$
($k+1 \leq d \leq 2n+k+1$) consisting of the tangents of all possible curves
$\fun{c}{\R}{\G}$ passing through $c_0$.
\end{mydef}
\end{comment}

For a point $c \in \G$, we have from \cite{Allgower1990, Krauskopf2007}
\begin{equation}
\label{eq:TcG}
T_{c}\G = \Null\left(J(c)\right), 
\end{equation}
where $\Null(J(c))$ is the null space of $J$ from Equation~\eqref{eq:J} and
$T_{c}\G$ is the tangent space of $\G$ at $c$ \cite{Choset2005}.

\begin{mydef} 
\label{def:sing}
A point $c \in \pre{P}$ is a \emph{singular point} of $P$ if $\rank(J(c)) < 2n$.
Points are \emph{regular} if they are not singular.  
\end{mydef}

The connected components of $\G$ generally consist of submanifolds of $\Space$
glued together at singular points of the periodicity map $P$.

%%%%%%%%%%%%%%%%%%%%%%%%%%%%%%%%%%%%%%%% Introduction - Problem Statement
\subsection{Problem Statement}
\label{ssec:prob}
Given
\begin{itemize}
\item a hybrid model $\Sigma = (\X, f, \Delta, \phi)$ of a biped,
\item a finite-dimensional space $\Space$ of parameterized trajectories, 
\item an implicit description of the set of all gaits $\G \subseteq \Space$ as
the points $c$ in $P^{-1}(0)$, and
\item a description of the set of equilibria $E \subset \G$,
\end{itemize}
use NCMs to approximately trace the connected components of $\G$ that contain
$E$.  The constructed set is denoted $\Gm$.

%%%%%%%%%%%%%%%%%%%%%%%%%%%%%%%%%%%%%%%% Mathematical Preliminaries - Assumption
\subsection{Assumptions}
\begin{myas}
Unless otherwise stated, we assume
\begin{enumerate}[label=A\theenumi]
\item \label{as:sym} Bipeds are physically symmetric about their sagittal plane.
\item \label{as:imp} Bipeds undergo exactly one collision per step, a plastic
impact between the pre-impact swing leg and the ground.  At impact, the
pre-impact stance leg breaks contact with the ground, and there is no free-flight \reply{or double-support}
phase (the stance leg instantaneously changes at impact).  No
slipping occurs at contacts between a foot and the ground. 
\item \label{as:phi} The next foot hits the support surface after a specified
period of time has elapsed, i.e., the impact is based on time, not state.  
\item \label{as:vhc} Bipeds may be subject to physical and virtual holonomic
constraints, but not nonholonomic constraints.
\end{enumerate}
\end{myas}

Assumptions~\ref{as:sym}--\ref{as:imp} allow us to take advantage of a biped's
symmetry to define a gait after one step with only one impact.  Additional
impacts would be needed to model the knees of a walker hitting a mechanical stop
to prevent hyperextension~\cite{Collins2005, Chen2007}.  Assumption~\ref{as:imp}
also rules out heel-toe collisions for bipeds with non-point
feet~\cite{Bessonnet2005, Hereid2018}.  This is a common assumption for point-,
curved-, and flat-footed walkers (i.e., all contact points on the bottom of a
foot impact the ground at the same time).

Given Assumption~\ref{as:phi}, the hybrid dynamics $\Sigma = (\X, f, \Delta, \phi)$ with fixed switching times is
\begin{equation}
\Sigma: 
\begin{cases} 
    \begin{aligned}
    \dot{x}(t) &= f(x(t), u(t)) & t \neq k \tau, \\
    x(t^+) &= \Delta(x(t^-)) & t = k \tau,
    \end{aligned}
\end{cases}
\label{eq:sigmatb}
\end{equation}
where $x(t) \in \X$, $f(x,u) \in T\X$, and $\Delta(x) \in \X$ are the state of
the robot at time $t$, a vector field, and a jump map, respectively (see
Equation~\eqref{eq:flow}), $x(t^-)$ and $x(t^+)$ are pre- and post-impact states,
respectively, and $k \geq 0 \in \mathbb{Z}$ is the $k^\text{th}$ impact.  The
switching function $\phi$ is $\phi(t, x) = t - k \tau$.

Finally, Assumption~\ref{as:vhc} precludes the use of virtual nonholonomic
constraints as discussed in~\cite{Griffin2015}.

%%%%%%%%%%%%%%%%%%%%%%%%%%%%%%%%%%%%%%%% Basics

\section{The Basic Gait-Generation Approach}
\label{sec:top}
We now present the core concepts and algorithms behind our framework.
Specifically, we
\begin{enumerate}
\item describe the connected components of $\G$ that include equilibrium gaits (EGs) (Section~\ref{ssec:cc}), 
\item describe how to find paths from an EG $\ceq \in E \subseteq \G$ to a set
of nonstationary gaits (nontrivial periodic orbits of the hybrid dynamics) in
$\G-E$ %given a set of switching times on a closed interval
(Section~\ref{ssec:I}),
\item present a continuation method for tracing curves (one-dimensional
manifolds) in $\G$ (Section~\ref{ssec:ncm}),
\item present an algorithm for constructing one-dimensional slices of $\G$ from
EGs (Section~\ref{ssec:e2g}), and
\item give an illustrative example of the basic approach using the compass-gait
walker (Section~\ref{ssec:cgw}).
\end{enumerate}

%%%%%%%%%%%%%%%%%%%%%%%%%%%%%%%%%%%%%%%% Topology - Connected Components

\subsection{Connected Components of $\G$ Containing Equilibrium Gaits}
\label{ssec:cc}
%%%%%%%%%%%%%%%%%%%%%%%%%%%%%%%%%%%%%%%% Figure: 2D State-Time-Control Space
\begin{figure}[t]
\centering
\includegraphics[width=0.48\textwidth]{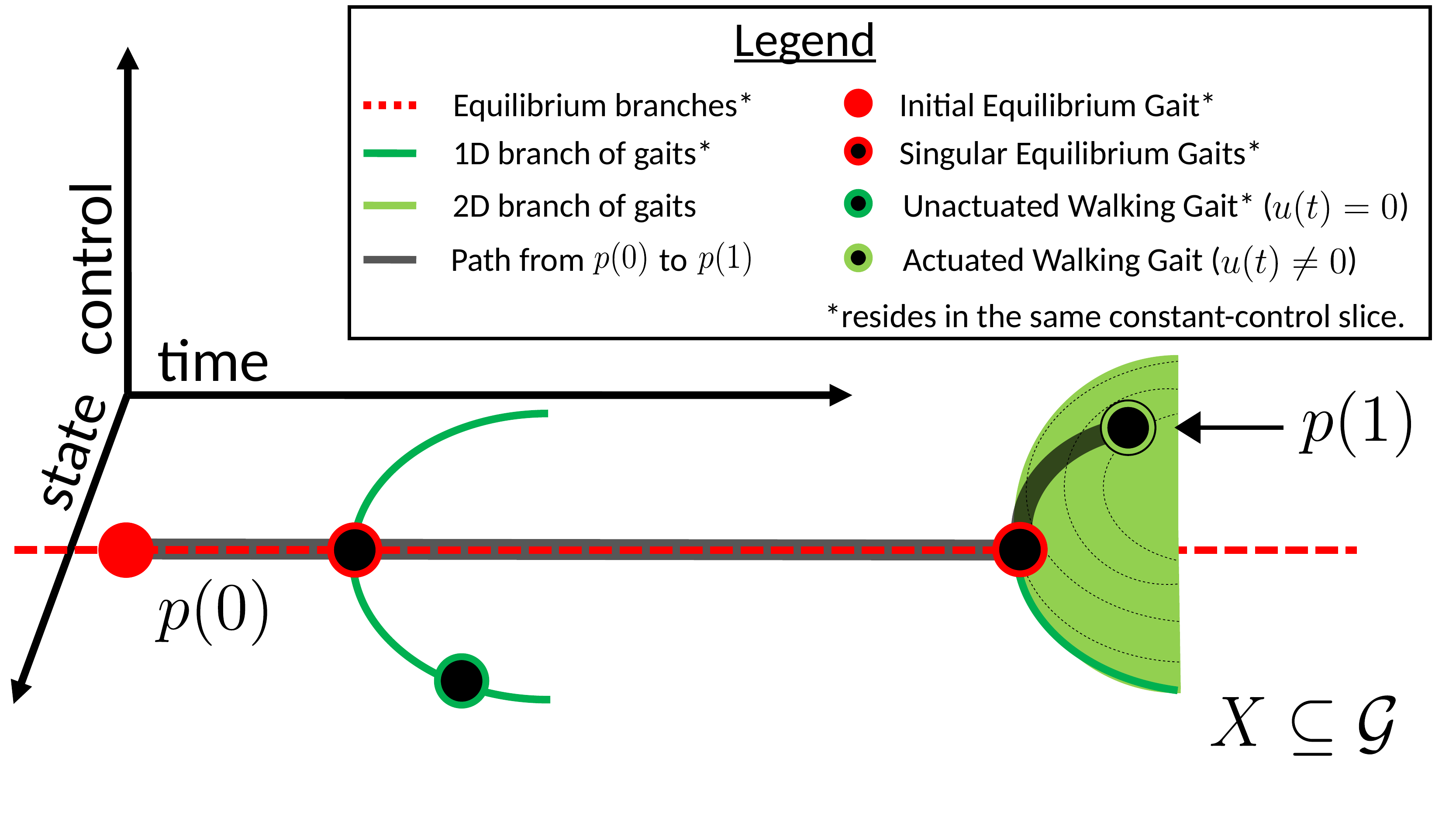}
\caption{\reply{A conceptual illustration of a portion of a connected component of gaits $X$ in $\G \subset \Space$ of a biped.  The branches of $X$ are manifolds of gaits joined together at singular points.  The manifolds depicted represent potential gait families of a connected component, including a 2D manifold of actuated gaits (right-most green surface) and a 1D (constant-control) slice of a larger 2D manifold consisting of unactuated (green curves) and equilibrium gaits (red-dashed curves).  This (user-defined) slice is used to search for singular equilibrium gaits and construct the initial set of gaits for $\Gm$.}}
\label{fig:S}
\end{figure}
\reply{The space of gaits $\G$ in the state-time-control space $\Space$ may consist of
multiple separate connected components \cite{Rosa2013}}.  We are particularly interested in those
connected components that include EGs.  A biped standing still on one foot is an example of an EG.

Figure~\ref{fig:S} is a conceptual depiction of a connected component of an EG
for an $n$-degree-of-freedom biped walker with one switching time and $k = 1$
design and control parameters.  The dimension of $\Space$ is $2n+k+1=2n+2$ and
the dimension of the manifolds in $\G$ are $\dim(\Space) - 2n = k + 1 =
2$-dimensional.  \reply{For illustrative purposes, we assume the connected component consists of manifolds of unactuated and actuated gaits.}  Examples of manifolds in the figure are the red-dashed and green curves (1D slices of a larger 2D manifold) and the green surface (a 2D manifold).  \reply{Specifically,
\begin{itemize}
    \item the green curves are 1D gait families of unactuated (passive dynamic walking) gaits,
    \item the green surface is a 2D branch of actuated gaits (we could, for example, expect $p(1) \in X$ to correspond to an actuated gait that walks on flat ground),
    \item the red dashed line consists of equilibrium gaits from which we would start to construct $\Gm$ (e.g., the EG $p(0) \in E$),
    \item the two singular points (black dots with thick red borders) divide the red dashed line into three distinct manifolds, and
    \item these singular points glue together the branches of unactuated and equilibrium gaits of $X$.
\end{itemize}
}

\reply{The connected component of an EG can be understood in terms of previous work with simple two-degree-of-freedom passive walkers \cite{McGeer1990a, Garcia1998, Goswami1998}.  While the hybrid dynamics in these works use state-based switching functions $\phi = \phi(x)$, the search for an initial set of gaits reduces to finding explicit values of the step duration $\tau$ that are roots of the linearized periodicity map $P$ with respect to state $x_0$; the linearized map is derived as a closed-form, Taylor-series approximation of $P$ about an equilibrium point $\xeq$.}

\reply{The connected component in Figure~\ref{fig:S} gives a geometric interpretation of this approach.  The search for roots of the linearized map is the same as the search being constrained to the red dashed line of $X$ in a constant-control slice of $\Space$.  The critical values of $\tau$ that are roots of the linearized map correspond to singular EGs on the line.  As can be seen in the figure, regular EGs cannot have critical values of $\tau$ because nearby walking gaits do not exist when we only consider perturbations in state in a constant-control slice.}

\reply{
In this paper, we generalize (and formalize) this analytical approach to multi-degree-of-freedom bipeds.  Given an EG, the task is to find paths from the EG in $E$ to a set of gaits in $\G - E$ on the EG's connected component.  Across biped models, EGs have two useful properties that we take advantage of in the construction of $\Gm$, which we prove in Section~\ref{ssec:I}:
\begin{enumerate}[label=P\theenumi]
\item \label{pr:branch} An EG in the gait space $\G$ is always a part of a
continuum of EGs of the form $\xeqtu$, where $\xeq$ and $\mu$ are fixed and
$\tau$ can take any positive value ($\tau > 0$), and
\item \label{pr:sing} In a constant-control slice of $\Space$, the same continuum of EGs often intersects with a branch of walking gaits residing in the slice.  The point of intersection can only happen at EGs that are singular points of $P$.
\end{enumerate}
}

\reply{
For Property~\ref{pr:branch}, the continuum of EGs only exists because we assume a robot's swing foot can
impact the ground at any time (Assumption~\ref{as:phi}).  For the biped standing still on one foot, Assumption~\ref{as:phi} implies that the state $\flow[\tau][][(\xeq)]$ at the end of the next step (the biped standing still on its opposite foot) does not change for any value of $\tau$ and fixed values of $\xeq$ and $\mu$.  In the robot's gait space, the resulting set of points $\xeqtu$ would correspond to the red dashed line of regular and singular EGs on the connected component in Figure~\ref{fig:S}.
}

\reply{Property~\ref{pr:branch} also places a constraint on an EG's tangent space (Definition~\ref{def:sing}).  For a regular EG in a constant-control slice, the constraint fully defines the EG's 1D tangent space such that it can only be a part of a single branch.}

\begin{mydef}
\reply{A path-connected set of EGs that are regular points of the periodicity map $P$ form an \emph{equilibrium branch} on the connected component.  Equilibrium branches reside in constant-control slices of $\Space$.
}
\end{mydef}

\reply{Property~\ref{pr:sing} implies that this is not the case for singular EGs.  At singular EGs in a constant-control slice, we can switch onto a non-equilibrium branch of gaits and, for example, trace a branch of walking gaits (dark green curves in Figure~\ref{fig:S}) for inclusion in $\Gm$.  These gaits can then be used to add gaits not in the constant-control slice (e.g., gaits on the 2D surface).}

We specify a constant-control slice of $\G$ with the map $M_0 : \Space \rightarrow \R^{2n+k}$
\begin{equation} 
\label{eq:M0}
\begin{aligned}
M_0(c) &= \begin{bmatrix} P^T(c), & \Phi_0^T(c) \end{bmatrix}^T, & \Phi_0(c) &=
\mu - \mu_0,
\end{aligned}
\end{equation}
where $\G_0 = \pre{M_0} \subset \Space$ is the set of gaits of $M_0$ and the $k$
auxiliary constraints $\Phi_0(c) = 0$ keep the controls constant at $\mu_0$.  In
other words, $\G_0$ lives in a constant-control slice of $\Space$ (the red and
dark green curves in Figure~\ref{fig:S}).  The set of equilibrium gaits of $M_0$
is $E_0 = E \cap \G_0$.  We specify how to find singular EGs in $E_0$ in the
next section.  In general, we use the subscript $0$, as in $M_0$, for variables
related to a constant-control slice $\G_0$ in $\G$.

\begin{myrem}
\reply{Under any state-based switching strategy, EGs are either isolated points in $\G$, or not in $\G$ at all, depending on whether conditions placed on the switching function $\phi$ allow for infinite impacts in zero time \cite{Bainov1993}.}
\end{myrem}

\begin{myrem}
The locations of singular EGs on a connected component are determined by the
kinematic and dynamic properties of the biped.  Any changes to the biped model,
for example, the addition of physical or virtual holonomic constraints,
modifications to the control inputs $u$, or changing the biped's total mass, can
shift the singular points or change the number of singular points on the
connected component.
\end{myrem}

\begin{comment}
In terms of the biped's hybrid dynamics, a regular EG indicates that small perturbations in state (which defines the walking surface) and step duration of the EG will not lead to gaits in $\G - E$.

\reply{
Furthermore, the red dashed line of regular and singular EGs exists because we assume a robot's swing foot can
impact the ground at any time (Assumption~\ref{as:phi}).  For the biped standing still on one foot, Assumption~\ref{as:phi} implies that the state $\flow[\tau][][(\xeq)]$ at the end of step does not change for any value of $\tau$ and fixed values of $\xeq$ and $\mu$.  In the robot's gait space, the resulting set of points $\xeqtu$ would correspond to the red dashed line on the connected component in Figure~\ref{fig:S}.}

\begin{myrem}
Equilibrium branches only exist because we assume a robot's swing foot can
impact the ground at any time.  Under any state-based switching strategy,
equilibria are either isolated points in $\G$, or not in $\G$ at all, depending on
whether conditions placed on the switching function $\phi$ allow for infinite
impacts in zero time \cite{Bainov1993}.
\end{myrem}

\begin{myrem}
We often choose constant-control slices where the gaits in $\G_0$ are open-loop
periodic motions.  In some cases, $\G_0$ is the set of passive dynamic walking
gaits for the biped model.
\end{myrem}
\end{comment}

%%%%%%%%%%%%%%%%%%%%%%%%%%%%%%%%%%%%%%%% Topology - Indicator Function

\subsection{Detecting Singular Equilibrium Gaits}
\label{ssec:I}
Given the characterization of equilibrium gaits in Section~\ref{ssec:cc},
consider the set of EGs $\{\ceq = (\xeq,\tau,\mu) \; | \; \tau>0\}$ for fixed
control inputs $\mu$.  For this set, the indicator function 
\begin{equation}
\label{eq:I}
I(\tau) = \det\left(\pd{P}{x_0}(\xeq, \tau, \mu)\right)
\end{equation}
can be used to identify singular EGs by searching along the switching-time axis
for values of $\tau$ that make $I(\tau)$ zero.

Given the map $M_0$ (Equation~\eqref{eq:M0}) and the Jacobian of $M_0$,
\begin{equation}
\label{eq:J0}
J_0(c) = \begin{bmatrix} \pd{P}{c}(c) \\ \pd{\Phi_0}{c}(c) \end{bmatrix}  =
\begin{bmatrix} 
\pd{P}{x_0}(c) & \pd{P}{\tau}(c) & \pd{P}{\mu}(c) \\
0 & 0 & \id_k
\end{bmatrix},
\end{equation}
the next two propositions prove that we can find a path from an equilibrium
point $\ceq \in E_0$ to a set of gaits in $\G_0 - E_0$ by searching for singular
EGs in $E_0$.  In the first proposition, we establish the existence of 1D
equilibrium branches in $\G_0$, which leads to a corollary that gives the
condition for when an EG is a regular point of $M_0$.

\begin{myprop}
\label{prop:reg}
Given
\begin{enumerate}
\item a biped's hybrid dynamics $\Sigma = (\X, f, \Delta, \phi)$, 
\item an equilibrium point $x_{{\rm eq}} \in \X$ of $f$, 
\item a switching time $\tau_0 \in \R$, and 
\item a vector of control parameters $\mu_0 \in \Rmu$
\end{enumerate}
such that $M_0(c_0) = 0$, where $c_0 = (\xeq, \tau_0, \mu_0) \in E_0$, if $c_0$
is a regular point of $\G_0$, then there exists a unique curve
$\fun{c}{(-\delta, \delta)}{E_0}$ of regular points contained in $E_0$ that
passes through $c_0$ at $c(0) = c_0$ for some $\delta > 0$.
\end{myprop}

\begin{proof}
See Appendix~\ref{app:ep}.
\end{proof}

\begin{mycor}
\label{cor:reg}
If $c_0 \in E_0$, then $\pd{P}{\tau}(c_0) = 0$ %for all $c_0 \in E_0$
and
$J_0(c_0)$ has rank of at most $2n+k$.  Furthermore, if $c_0 \in E_0$ is a
regular point of $M_0$, then the submatrix
\begin{equation*} 
\bar{J} = \begin{bmatrix}
\pd{P}{x_0}(c_0) & \pd{P}{\mu}(c_0) \\ 0 & \id_k \end{bmatrix} \in \R^{(2n+k)
\times (2n+k)}
\end{equation*}
of the Jacobian $J_0$ of Equation~\eqref{eq:J0} has full rank $2n+k$.
\end{mycor}

The next proposition states that $I(\tau)$ of Equation~\eqref{eq:I} can detect
singular EGs in $E_0$.

\begin{myprop}
\label{prop:sing}
Assume there exists a path $\fun{p}{[0, 1]}{\G_0}$ such that $p(0) \in E_0$ and
$p(1) \in \G_0 - E_0$.  If $p(0)$ is a regular point of $M_0$, then 
\begin{enumerate}
\item the path $p$ contains at least one singular equilibrium point $p(s) \in
E_0$, and
\item for each singular equilibrium point $p(s) \in E_0$ for $s \in (0, 1)$, 
$\det(\pd{P}{x_0}(p(s))) = 0$.
\end{enumerate}
\end{myprop}

\begin{proof}
  See Appendix~\ref{app:ep}.
\end{proof}

After identifying a singular EG using the indicator function, the next step is
switching onto a branch of gaits in $\G_0 - E_0$.  We can determine the correct
branch for the case where the singular EG, say $c_0$, is isolated in $E_0$ and
its tangent space $T_{c_0} \G_0$ is two dimensional.

\begin{myprop}
\label{prop:curves}
If $c_0 \in E_0$ is an isolated singular point in $E_0$ and $\dim(T_{c_0} \G_0)
= 2$, then taking a step in the direction of the tangent vector in $T_{c_0} \G_0$
orthogonal to the switching-time direction switches onto a branch of gaits in
$\G_0 - E_0$.
\end{myprop}

\begin{proof}
  See Appendix~\ref{app:ep}.
\end{proof}

\begin{comment}
A proof for a generalization of Proposition~\ref{prop:curves} with non-isolated
singular EGs with higher-dimensional tangent spaces would have similarities to
Proposition~\ref{prop:curves} in that the general idea is to remove the tangent
vectors in $T_{c_0} \G_0$ that are in $TE_0$.
\end{comment}

%%%%%%%%%%%%%%%%%%%%%%%%%%%%%%%%%%%%%%%% Algorithm: Finding roots of I(t)
\begin{algorithm}[t]
\caption{Detecting singular equilibrium gaits}
\begin{algorithmic}[1]
\Require an interval $[a, b] \subset \R$ and a step size $h \in \R$.
\State Define functions $\ceq(t) = (x_\text{eq}, \tau_0 + t, \mu_0)$ and
\State $\delta(t) = \det\left(\pd{P}{x_0}(\ceq(t)) - \id_{2n}\right)$
\State $N = \frac{b - a}{h}$
\For {$i := 1..N$}
\State $t = a + i \times h$
\If {$\delta(t) \times \delta(t - h) \leq 0$}
\State Solve for $\delta(t_0) = 0$ with $t_0 \in [t-h, t]$
\State Store $\ceq(t_0)$ as a singular equilibrium point
\State Store tangent vector $\sd{c_\text{eq}}{s}(t_0)$ such that 
\State $J_0(\ceq(t_0))\sd{c_\text{eq}}{s}(t_0) = 0$, $||\sd{c_\text{eq}}{s}(t_0)||
= 1$, and
\State $\sd{c_\text{eq}}{s}(t_0) \notin T_{\ceq(t_0)}E_0$. 
\EndIf
\EndFor
\State \textbf{return} singular EGs and tangent vectors $(\ceq(t_0),
\sd{c_\text{eq}}{s}(t_0))$
%\State \textbf{return} singular EGs $\ceq(t_0)$ and tangent vectors $\sd{c_\text{eq}}{s}(t_0)$
\end{algorithmic}
\label{alg:sfp}
\end{algorithm}
%%%%%%%%%%%%%%%%%%%%%%%%%%%%%%%%%%%%%%%%
Given Propositions~\ref{prop:reg}--\ref{prop:curves}, we can automate the search
for a singular EG along the switching-time axis (keeping the state and control
constant) using Algorithm~\ref{alg:sfp}.  Algorithm~\ref{alg:sfp} finds simple
roots of Equation~\eqref{eq:I} (i.e., if $I(\tau) = 0$, then $\sd{I}{\tau}(\tau)
\neq 0$) in a given interval by applying the intermediate value theorem to first
bracket a root and then switching to a root-finding algorithm to accurately find
the root.  The step size $h$ should be chosen with care to avoid skipping over
multiple roots in a given subinterval (we use $3 \times 10^{-13}$).  Alternative
univariate root-finding algorithms can be found in \cite{Press2002}.  In the
end, all singular EGs detected using Algorithm~\ref{alg:sfp}, their
corresponding tangent vector(s) that are orthogonal to the switching-time
dimension, and the map $M_0$ serve as inputs to the numerical continuation
method (NCM) of the next section.

\begin{myrem}
Conditions for when a singular EG is isolated can be found in
\cite{Allgower1990}.  In Proposition~\ref{prop:curves}, we assume isolated
singular points with $\dim(T_{c_0} \G_0) = 2$ because it is the most common type
of singular EG we encounter in practice.
\end{myrem}

%%%%%%%%%%%%%%%%%%%%%%%%%%%%%%%%%%%%%%%% Topology - Numerical Continuation Meth

\subsection{Tracing Branches with Numerical Continuation Methods}
\label{ssec:ncm}
Continuation methods are useful numerical tools for tracing the level set of a
continuously differentiable function.  While multi-dimensional continuation
methods exist \cite{Allgower1990, Krauskopf2007, Henderson2002}, we present an
NCM for tracing one-dimensional manifolds (curves) $\fun{c}{\R}{\G}$ in the
$(2n+k+1)$-dimensional state-time-control space $\Space$.  A curve in $\G$ is
implicitly defined such that every point on the curve $c(s) \in \G$ satisfies
\begin{equation}
M(c(s)) = [P^T(c(s)), \Phi^T(c(s))]^T = 0,
\label{eq:Meq0}
\end{equation}
where $\fun{M}{\Space}{\R^{2n+k}}$ is a continuously differentiable map, $P$ is
the periodicity map, $\Phi(c) \in \R^k$ is a set of $k$ auxiliary user-specified
constraints to select a one-dimensional curve to trace in the
$(k+1)$-dimensional set $\G$, and $s \in \R$ is the curve's arclength.  The
curve $c$ is as smooth as the map $M$ (Theorem~\ref{thm:impman}).
%%%%%%%%%%%%%%%%%%%%%%%%%%%%%%%%%%%%%%%% PSALC Algorithm
\begin{algorithm}[t]
\caption{Pseudo-arclength continuation method}
\begin{algorithmic}[1]
\Require $\fun{M}{\R^{\nc}}{\R^{2n+k}}$ and step size $h \in \R$.
\Function{cmstep}{$c$, $\dot{c}$, $h$}
\State \textbf{Assume}: $M(c) = 0$, $\pd{M}{c}(c)\dot{c} = 0$, and $||\dot{c}||
= 1$
\State \textbf{Prediction Step}:
\State $z = c + \dot{c} h$ \label{alg:en:pred}
\State \textbf{Correction Step}:
\State \label{alg:en:n1} Solve for $M(z) = 0$ and $\dot{c}^T(z - c) = h$
\State using Newton's method \label{alg:en:n2}
\State \textbf{return} $z$
\EndFunction
\Function{cmcurve}{$c_0$, $\sd{c_0}{s}$, $M$}
\State Set $c[0] = c_0$ and $\dot{c}[0] = \sd{c_0}{s}$.
\For {$i := 1..N$}
\State \label{alg:en:cmstep} $c[i] = $ \Call{cmstep}{$c[i-1]$, $\dot{c}[i-1]$, $h$}
\State \label{alg:en:cdot} Set $\dot{c}[i]$ such that $\pd{M}{c}(c[i])\dot{c}[i] = 0$ and $||\dot{c}[i]|| = 1$
\If {$\dot{c}^T[i]\dot{c}[i-1] < 0$}
\State $h = -h$
\EndIf
\EndFor
\State \textbf{return} the solution curve $c$
\EndFunction
\end{algorithmic}
\label{alg:en}
\end{algorithm}
%%%%%%%%%%%%%%%%%%%%%%%%%%%%%%%%%%%%%%%%

For a user-defined map $M$, the function $\Phi$ of Equation~\eqref{eq:Meq0}
defines the one-dimensional slice in $\Space$ that contains the curve $c$.  The
map $\Phi_0$ of Equation~\eqref{eq:M0} is an example of $\Phi$.  In general, the
map $\Phi$ can contain any entry that can be computed as a function of $c$,
including
\begin{enumerate}
\item desired values for pre- or post-impact states,
\item gait properties like step length, walking speed, step incline, or minimum
step height (which can be represented as an equality constraint using slack
variables \cite{Nocedal1999}), or
\item VHC boundary conditions, where, assuming $a$ is the polynomial coefficient
vector of the \Bezs of Equation~\eqref{eq:qi} and $\mu = [\ldots, a^T,
\ldots]^T$, then $x_0$ and $a$ have to satisfy periodic boundary conditions of
Equation~\eqref{eq:qi} at $t = 0$ and $t = \tau$.
\end{enumerate}

Algorithm~\ref{alg:en} describes the pseudo-arclength continuation method
\cite{Allgower1990} for tracing a curve in $\G$ using the map $M$.  The core
part of the algorithm is in \textproc{cmstep}; the function \textproc{cmcurve}
simply calls \textproc{cmstep} $N$ times.

\begin{figure}[t]
\centering
\includegraphics[width=0.45\textwidth]{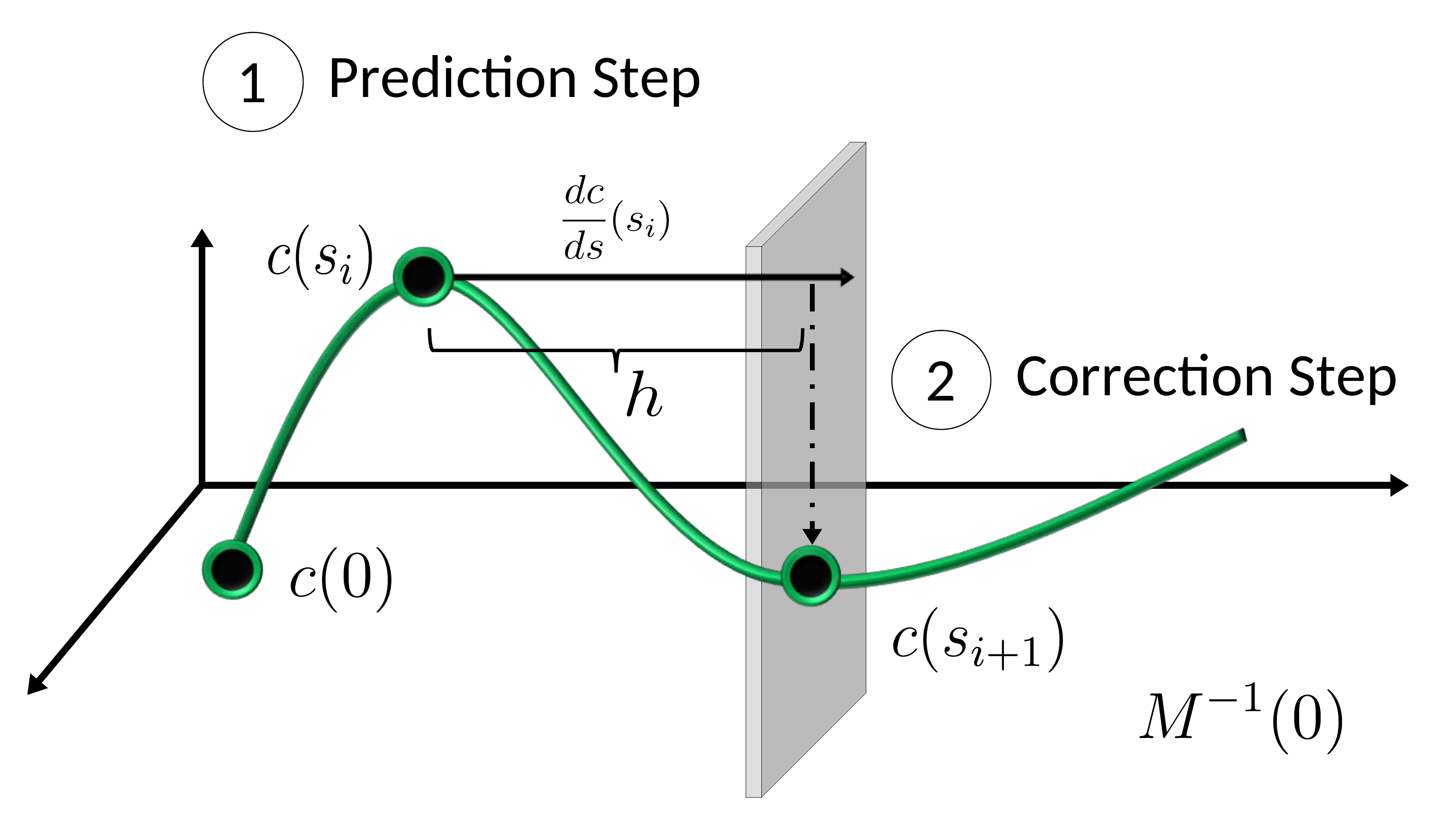}
\caption{An example iteration of the pseudo-arclength continuation method as
outlined in Algorithm~\ref{alg:en}.}
\label{fig:en}
\end{figure}

Geometrically, Algorithm~\ref{alg:en} defines a hyperplane a distance $h$ away
from the current point $c(s_i)$ on the curve where the search for the next point
on the curve $c(s_{i+1})$ takes place; the hyperplane is normal to the tangent
$\sd{c}{s}(s_i)$ at $c(s_i)$.  The algorithm's prediction step
(line~\ref{alg:en:pred} of the algorithm) selects a point on the plane as the
initial guess using an Euler-like integration step and then a root-finding
method iteratively refines the guess until the point is on the curve.
Figure~\ref{fig:en} illustrates this process.

In order to define the hyperplane, we can compute a tangent to the curve
$\sd{c}{s}(s)$ at $c(s)$ by solving for $\pd{M}{c}(c(s)) \sd{c}{s}(s) = 0$
(line~\ref{alg:en:cdot} of Algorithm~\ref{alg:en}).  In other words, the tangent
$\sd{c}{s}(s)$ is in the null space of the Jacobian of the map $M$
\begin{equation} T_{c(s)}\pre{M} =
\Null\left(\pd{M}{c}(c(s))\right),
\end{equation} 
where $T_{c(s)}\pre{M}$ is the tangent space of $\pre{M}$ at $c(s)$.  An
arclength parameterization of the curve leads to the constraint
$||\sd{c}{s}(s)|| = 1$ (see \cite{Allgower1990}).  If the point $c(s)$ is a
regular point of $M$, then $\dim(T_{c(s)}\pre{M}) = 1$ ($= \dim(\Space) -
(2n+k)$ constraints).
%%%%%%%%%%%%%%%%%%%%%%%%%%%%%%%%%%%%%%%%
%%%%%%%%%%%%%%%%%%%%%%%%%%%%%%%%%%%%%%%%

Algorithm~\ref{alg:pn} is the projected Newton's method \cite{Bertsekas1982,
Kelley1999}, a root-finding algorithm for use in
lines~\ref{alg:en:n1}--\ref{alg:en:n2} of Algorithm~\ref{alg:en}.  The projected
Newton's method is a variant of Newton's method that imposes box constraints $L
\leq c \leq U$ on the values of $c \in \Space$, where $L, U \in \Space \cup
\{\pm \infty\}$ specify the lower- and upper-bounds of $c$, respectively, and
the relational operators are applied elementwise.  An application of the
projected Newton's method is modeling inequality constraints, like the swing leg
of a biped staying above or on the walking surface, as equality constraints (see
Section~\ref{sec:ex} for an example).

\begin{myrem}
Part of the input to Algorithm~\ref{alg:en} is a vector $\sd{c_0}{s}$ tangent to
the initial point $c_0$.  If the manifold $\pre{M}$ is a one-dimensional
differentiable manifold, then the tangent space is one-dimensional and there is
no need to pass $\sd{c_0}{s}$ as an argument; the algorithm can compute the
tangent internally.  However, we use NCMs to generate branches of connected
components starting from a singular point, in which case we do need to specify
the tangent vector as the null space has dimension greater than one
\cite{Allgower1990}.  If we do not specify the tangent at a singular point, the
behavior of the algorithm is implementation dependent.
\end{myrem}

%%%%%%%%%%%%%%%%%%%%%%%%%%%%%%%%%%%%%%%% Topology - From Equilibria

\subsection{From Equilibria to One-Dimensional Sets of Walking Gaits}
\label{ssec:e2g}
Given an EG $\ceq \in E_0$, Algorithm~\ref{alg:gm} constructs $\Gm$ in a
constant-control slice $M_0(c) = 0$ in $\Space$.  For simplicity, we define
$\Gm$ as a two-dimensional array of gaits, but other data structures can be
used.  Most of the details of the algorithm are covered in Sections~\ref{ssec:I}
and \ref{ssec:ncm} (e.g., lines~\ref{alg:gm:search} and \ref{alg:gm:trace}).  In
particular, $N$ new gaits are added to $\Gm$ when we successfully return from
calls to Algorithm~\ref{alg:en} (line~\ref{alg:gm:trace}).  In the event that
Algorithm~\ref{alg:sfp} is not able to find isolated singular EGs, an error
message is printed along with a plot of the indicator function $I$
(lines~\ref{alg:gm:err1}--\ref{alg:gm:err2}).  Further analysis of the plot
provides potential directions for improving the model.  We list an informal list of steps in Appendix~\ref{app:err}.

%%%%%%%%%%%%%%%%%%%%%%%%%%%%%%%%%%%%%%%% Examples - Compass Gait Walker

\subsection{An Illustrative Example of the Basic Gait-Generation Approach Using
the Compass-Gait Walker}
\label{ssec:cgw}
\begin{figure}[t]
\centering
\subfloat[]{\includegraphics[width=0.3\textwidth]{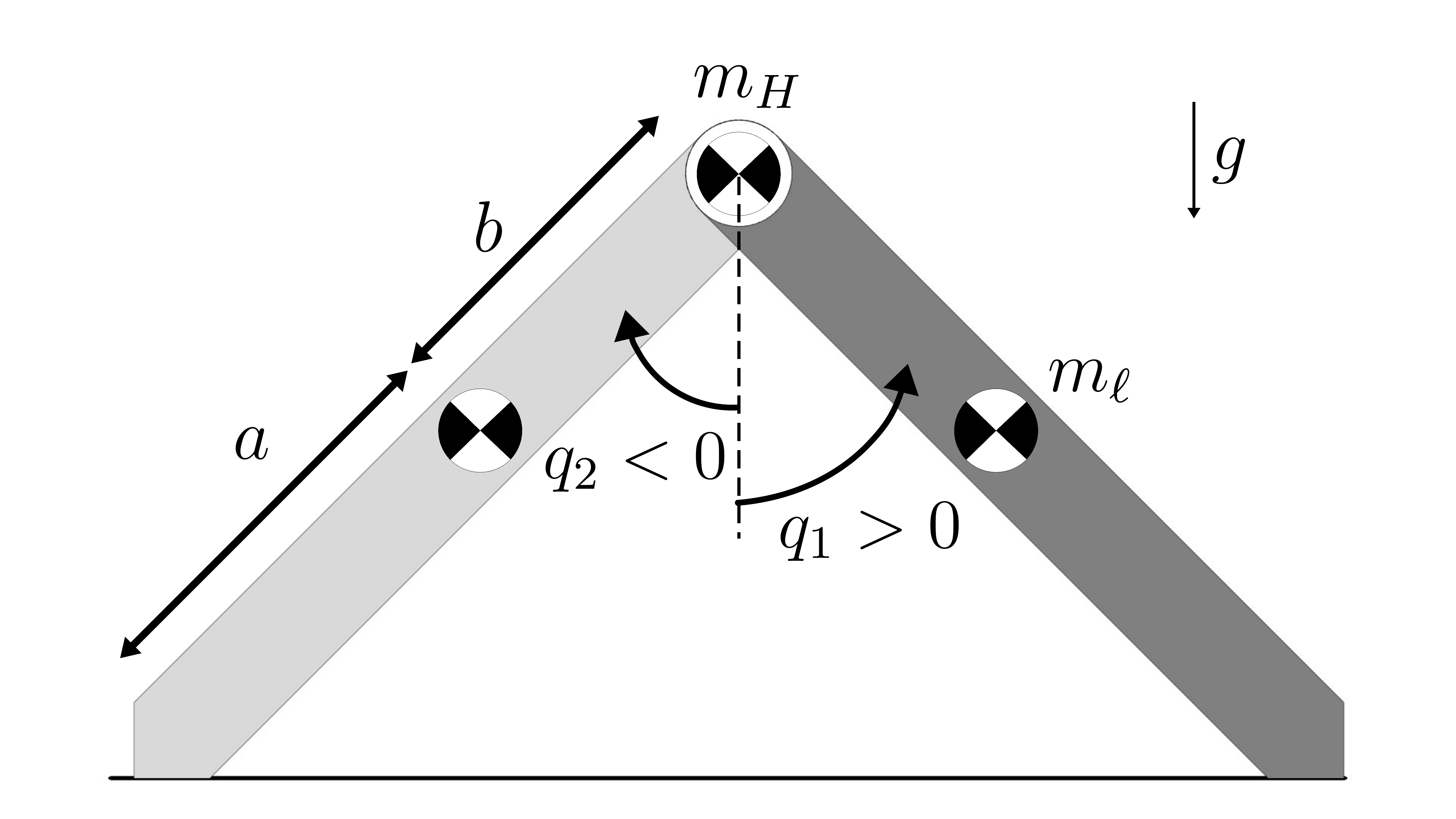}} \hfill
\subfloat[]{\includegraphics[width=0.02\textwidth]{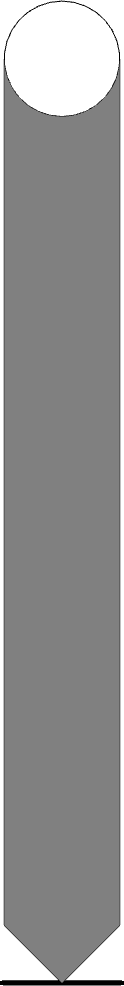}} \hfill
\subfloat[]{\includegraphics[width=0.02\textwidth]{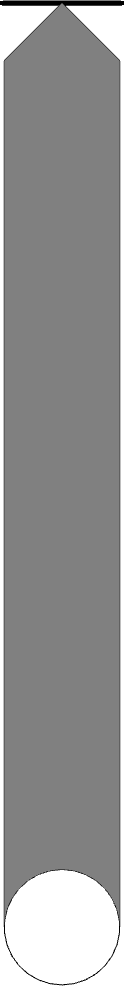}}
\caption{\reply{(a) The compass-gait model.  The legs in the model have the same physical parameters $a$, $b$, and $m_\ell$. (b)--(c)  The two equilibrium gaits in $E \subset \G$ of the compass-gait walker.  These non-locomoting gaits are used to generate connected components of walking and brachiating gaits for this model.  Note that only one leg is in contact with the surface as stated in Assumption~\ref{as:imp}.}}
\label{fig:cgw-model}
\end{figure}

We present an example of extending equilibria to periodic orbits for the passive
compass-gait walker \cite{Goswami1998}.  The compass-gait walker is a common
two-link model.  The model
(Figure~\ref{fig:cgw-model}(a)) consists of two legs each with a point mass $m$ and length $a+b$.  The biped
also has a large point mass $m_H$ at the hip.  We use the same values for the physical
parameters as in \cite{Goswami1998} with $\frac{m_H}{m} = 2$, $\frac{b}{a} = 1$,
and $g = 9.81$~m/s$^2$.  The state of the robot is $x = [q_1, q_2, \dot{q}_1,
\dot{q}_2]^T \in \R^4$, representing the two leg angles and their velocities.
With this minimal set of coordinates, we can directly compare our results to
those in \cite{Goswami1998}.  The angle of the walking surface $\sigma \in \R$
is implicitly defined according to the position of the swing leg's foot at $t =
0$: $\sigma = \frac{1}{2}(q_1(0) + q_2(0))$.  The biped has no motors ($n_u =
0$), no VHCs ($n_v = 0$), and two PHCs ($n_p = 2$) representing the no-slip
contact conditions between the stance foot and the ground.  We do not place constraints on the resulting ground reaction force arising from the PHCs.
%%%%%%%%%%%%%%%%%%%%%%%%%%%%%%%%%%%%%%%% Project Newton Algorithm
\begin{algorithm}[t]
\caption{Projected Newton's method}
\begin{algorithmic}[1]
\Require $\fun{r}{\Space}{\R^m}$, where $m \leq \dim(\Space)$.
\Function{projnewt}{$c$, $L$, $U$}
\State $z = c$
\Repeat
\State \textbf{Compute Newton Step}:
\State $d = \pd{r}{c}(z)^\dagger r(z)$, where $\left[ \cdot \right]^{\dagger}$ is the 
\State Moore-Penrose inverse
\State $\Delta = z - d$
\State \textbf{Project onto Box Constraints}:
\For {$1 \leq i \leq (\nc)$}
\If {$\Delta[i] \leq L[i]$}
\State Set to lower bound: $z[i] = L[i]$
\ElsIf {$\Delta[i] \geq U[i]$}
\State Set to upper bound: $z[i] = U[i]$
\Else
\State Take Newton step: $z[i] = \Delta[i]$
\EndIf
\EndFor
\Until {a stopping criterion is met}
\State \textbf{return} $z$
\EndFunction
\end{algorithmic}
\label{alg:pn}
\end{algorithm}
%%%%%%%%%%%%%%%%%%%%%%%%%%%%%%%%%%%%%%%%
\begin{algorithm}[t]
\caption{Constructing $\Gm$ from equilibria}
\begin{algorithmic}[1]
\Require $\ceq \in E_0$.
\State \textbf{Search for Singular Equilibrium Gaits}:
\State Call Algorithm~\ref{alg:sfp} with a search interval of $\tau \in [a, b]$. \label{alg:gm:search}
\State Store singular EGs and their tangent vectors in arrays
\State $A$ and $\dot{A}$, respectively.
\State \textbf{Generate Curves in $\G - E$}:
\If {$|A| > 0$}
\For {$i := 1..|A|$}
\State \label{alg:gm:trace} $\Gm[i-1][0..N] = $ \Call{cmcurve}{$A[i]$, $\dot{A}[i]$, $M_0$}
\EndFor
\Else
\State Print ``No isolated singular equilibrium gaits found.'' \label{alg:gm:err1}
\State Print plot of Indicator Function $I(\tau)$ for $\tau \in [a, b]$. \label{alg:gm:err2}
\EndIf
\State \textbf{return} $\Gm$
\end{algorithmic}
\label{alg:gm}
\end{algorithm}

After using this data to derive the biped's hybrid dynamics, the goal is to find
period-one walking gaits in a five-dimensional state-time space $\Space$.  A
point $c \in \Space$ consists of the pair $(x_0, \tau)$, where $x_0 \in \X$ is a
pre-impact state and $\tau \in \R$ is a switching time (in seconds).  There are
no control parameters $\mu$ ($k = 0$).  Given the five parameters that
define $\Space$ and the four periodicity constraints of the periodicity map $P$,
we expect to find one-dimensional manifolds of gaits in $\Space$.  The search
for a walking gait starting on a manifold of EGs is a two-step process:
\begin{enumerate}
\item Identify a subset of EGs of interest in $E \subset \G$,
\item Choose a $\ceq \in E$ and call Algorithm~\ref{alg:gm}, which
\begin{enumerate}
\item calls Algorithm~\ref{alg:sfp} to find all singular
EGs in a closed interval of switching times $\tau \in [a, b] \subset \R$ where
$0 \leq a < b$, then
\item calls Algorithm~\ref{alg:en} with the map $M_0 = P$, a singular EG $\ceq$,
and the correct tangent vector $\sd{\ceq}{s}(s) \in T_{\ceq}\pre{P}$.
\end{enumerate}
\end{enumerate}

The first step for the compass gait is straightforward.  The biped's state space
$\X$ has four one-stance-foot equilibrium points, but only two \reply{equilibria} in $\X$
correspond to fixed points of $\G$ because of the $\flip$ operator.  These are $\xeq = [0, 0, 0, 0]^T$ (standing
on a surface) and $\xeq^\pi = [\pi, \pi, 0, 0]^T$ (hanging below a surface), as
shown in Figure~\ref{fig:cgw-model}(b)--(c).  These points define the set of
equilibria $E \subset \G$, where \reply{$E = E_w \cup E_b$, $E_w = \{ \ceq \in \R^5:
[0, 0, 0, 0, \tau]^T\}$, and $E_b = \{ \ceq \in \R^5: [\pi, \pi, 0, 0,
\tau]^T\}$}.  We start our search using the equilibrium $\xeq$, which gives rise
to nearby walking gaits.  If we had started with $\xeq^\pi$, we would find
nearby (overhand) brachiating gaits.

\begin{figure*}
\centering
\subfloat[]{\includegraphics[width=0.95\textwidth]{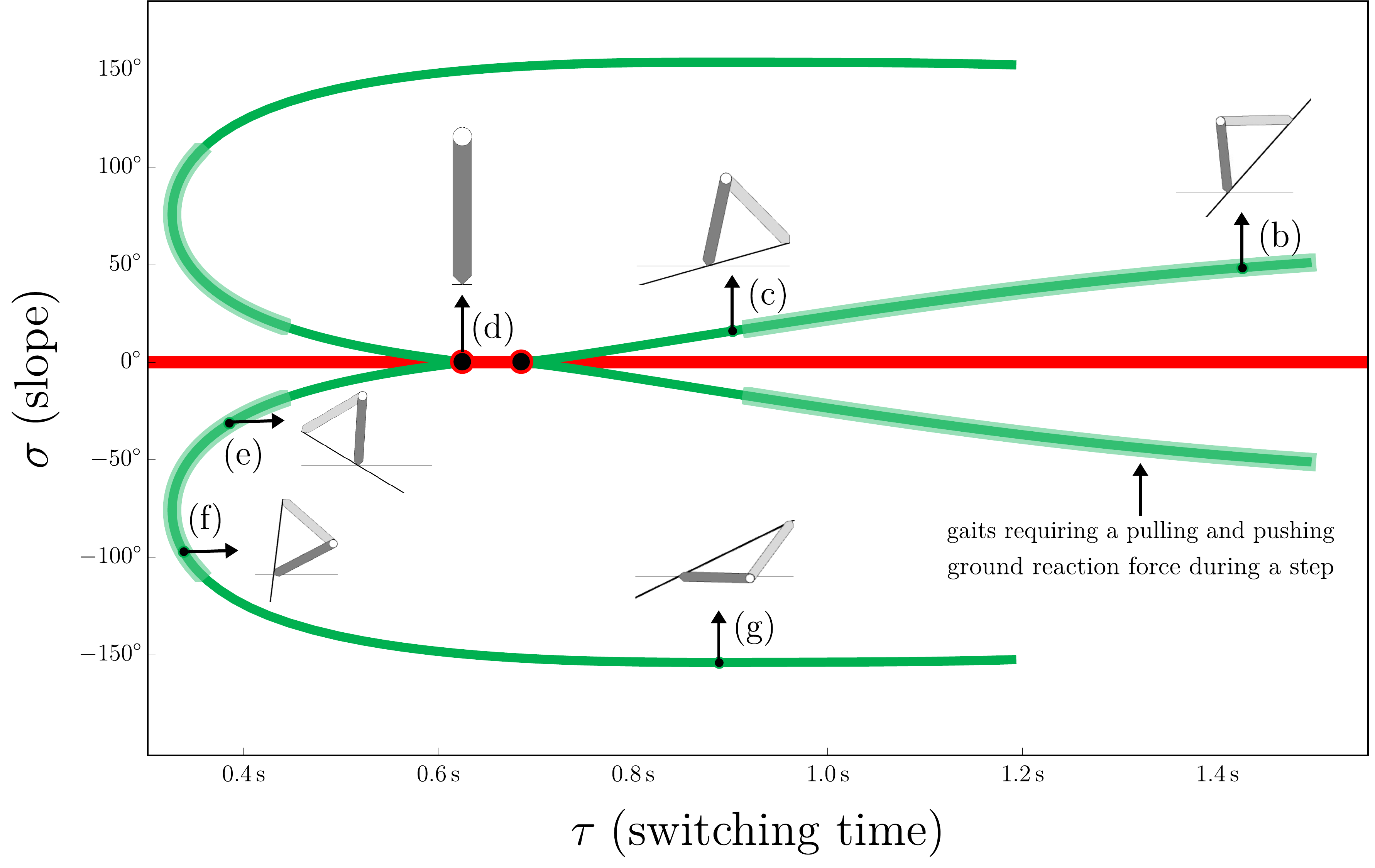}} \\
\subfloat[]{\includegraphics[width=0.45\textwidth,height=72.48032pt]{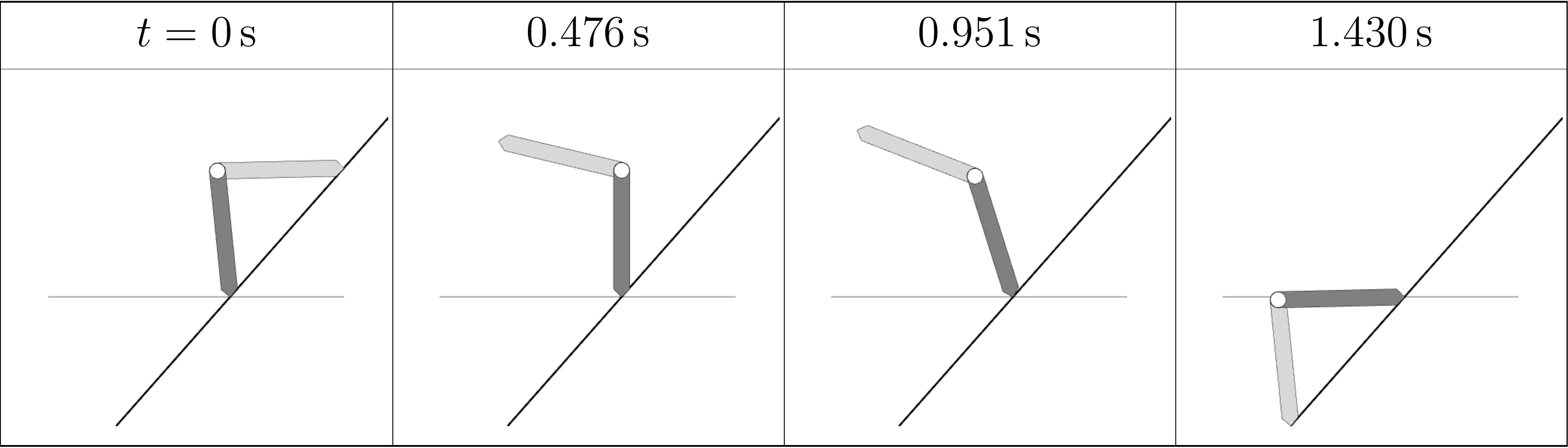}} \hfill
\subfloat[]{\includegraphics[width=0.45\textwidth]{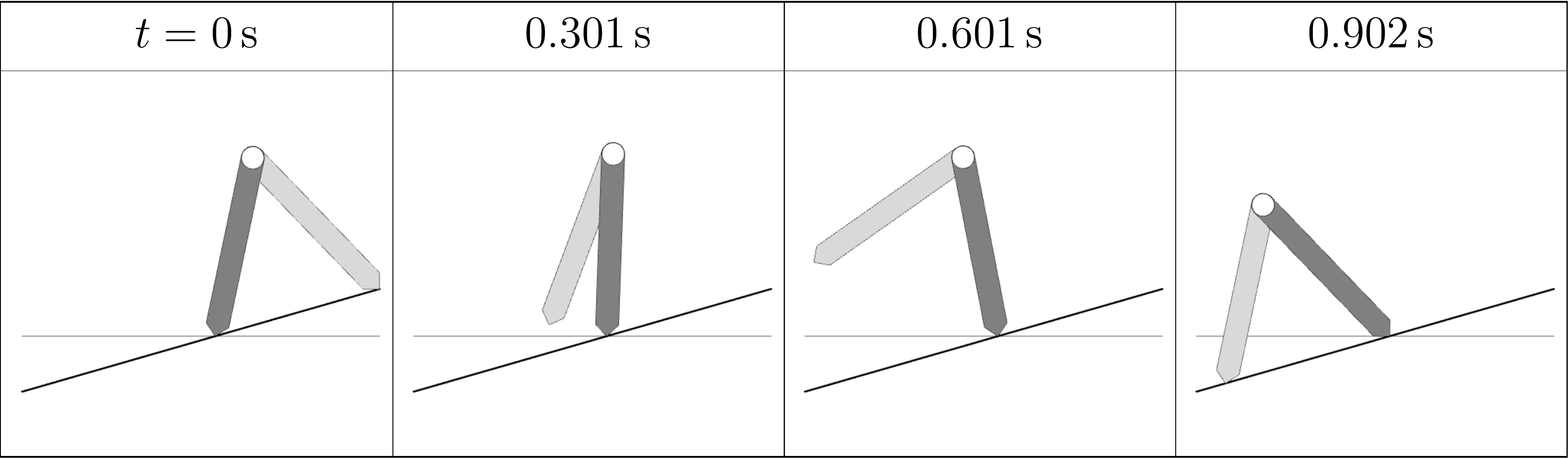}} \\
\subfloat[]{\includegraphics[width=0.45\textwidth]{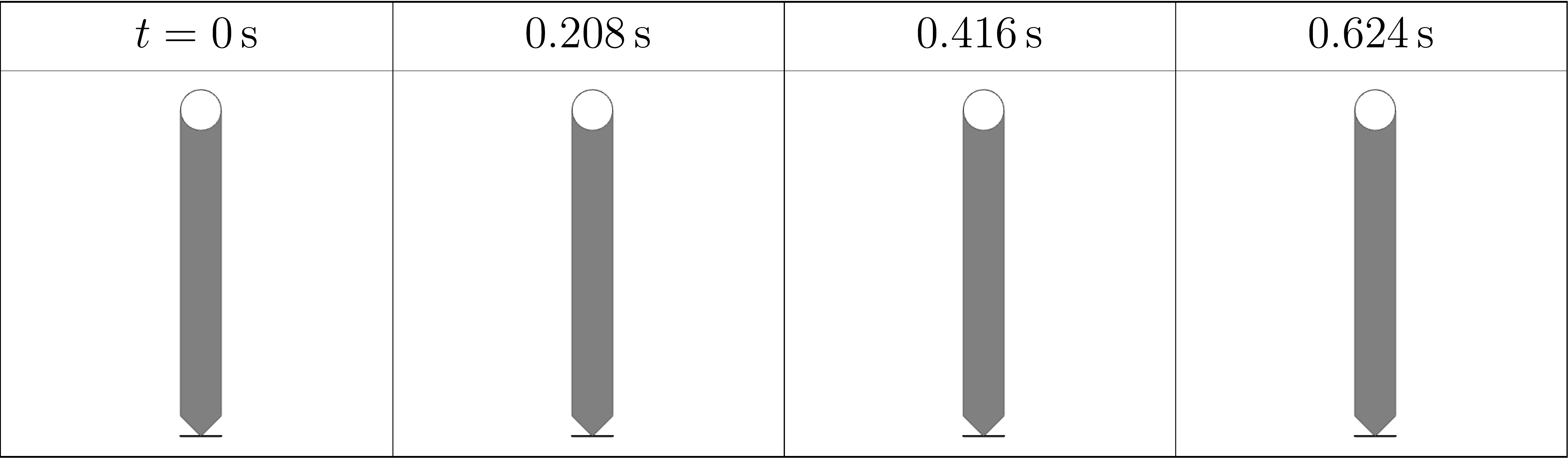}} \hfill
\subfloat[]{\includegraphics[width=0.45\textwidth]{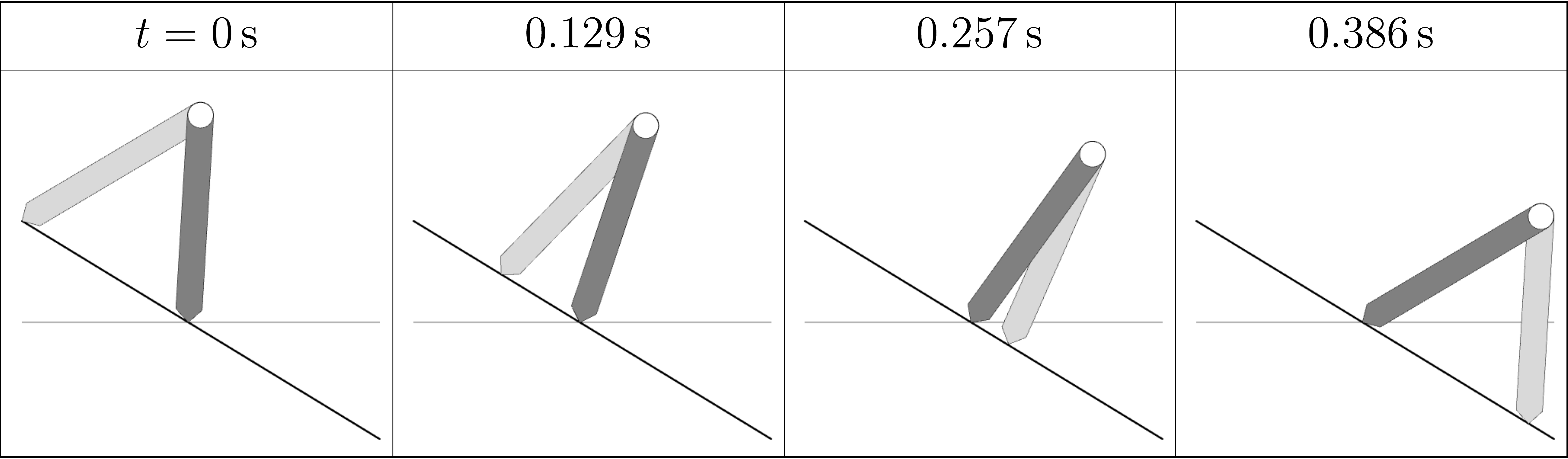}} \\
\subfloat[]{\includegraphics[width=0.45\textwidth]{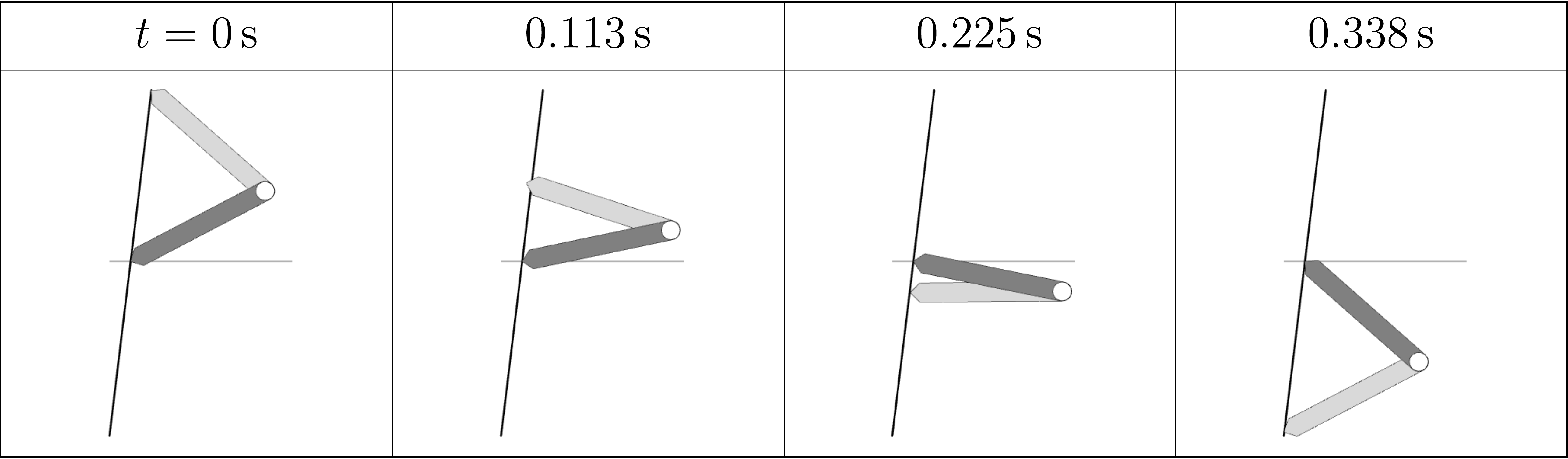}} \hfill
\subfloat[]{\includegraphics[width=0.45\textwidth]{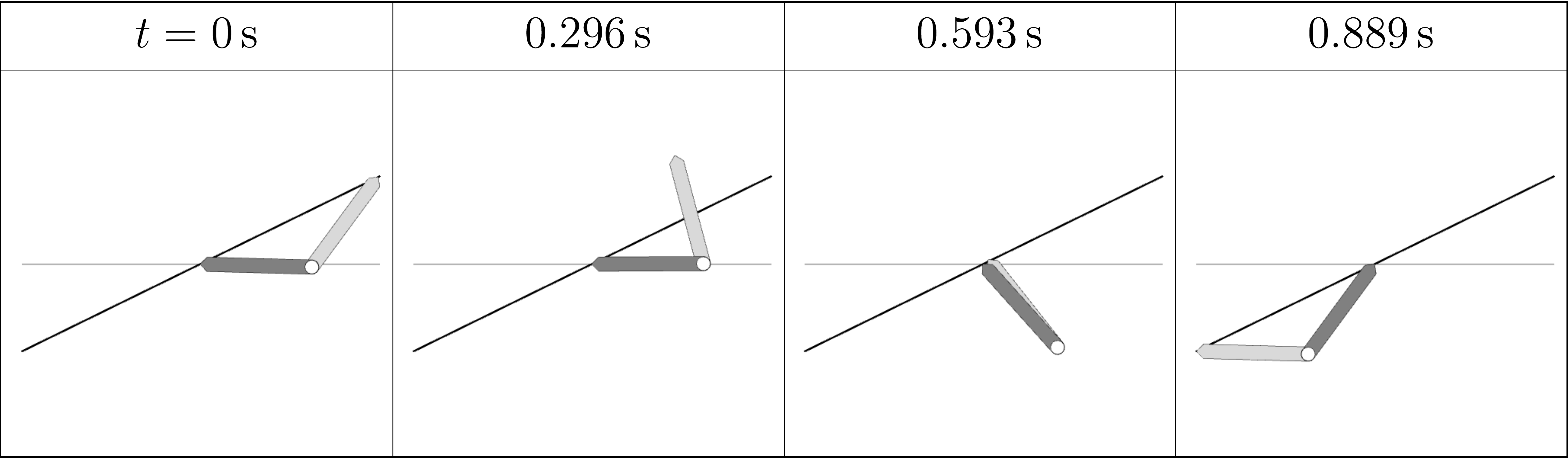}}
\caption{(a) A continuous set of unactuated periodic motions $\Gm$ that satisfy
the compass gait's hybrid dynamics as points in a parameter space $\Space$
projected onto a slope-switching-time ($\sigma$-$\tau$) plane (red and green
curves); the slope $\sigma$ is the biped's walking surface.  The plot consists of \reply{two singular equilibrium gaits \reply{(black dots with thick red borders)}, three
equilibrium branches (red curves separated by the two singular points), and four 1D manifolds of walking and brachiating gaits (green curves)}.  Specifically, each of the two leftmost green curves are
linear interpolations of 250 gaits computed with Algorithm~\ref{alg:gm} (the
maximum number of gaits we have Algorithm~\ref{alg:gm} compute), the two
rightmost green curves consist of the first 113 points we computed on each
branch, and the red line is added after the fact to represent the equilibrium
branches.  The thicker regions of the green curves correspond to walking gaits, where the ground pushes and pulls the foot of the robot in order to maintain the no-slip constraint.  The callout labels in the plot correspond to the animated
trajectories in (b)--(g);  the images in the plot are the pre-impact
configurations of the biped at $t = 0$.  (b)--(g) The motion of the gaits
depicted in (a) with respect to (absolute) time $t \geq 0$.  Gaits (b), (c),
(f), and (g) locomote from right to left and (e) goes from left to right.
Because of the biped's symmetry, every gait in (a) has a mirrored version of
itself about the $\tau$ axis.  In other words, trajectories in (b)--(g) that
locomote on a slope $\sigma$ have mirrored trajectories that walk or brachiate
downhill on a slope of $-\sigma$.}
\label{fig:cgw-cc}
\end{figure*}

Algorithm~\ref{alg:gm} performs the next step in two parts.  The algorithm first
calls Algorithm~\ref{alg:sfp} to find roots of the indicator function $I(\tau)$
of Equation~\eqref{eq:I} over the interval $\tau \in [0, 1]$.  The roots
correspond to the singular EGs of Figure~\ref{fig:cgw-cc} at $\tau =
0.62\,\text{s}$ and $\tau = 0.68\,\text{s}$, respectively \reply{(black dots with thick red borders)}.

At these singular EGs, there are two tangent vectors in the null space of $J(c)
= \pd{P}{c}(c)$ of Equation~\eqref{eq:J}.  Choosing an orthonormal basis, the tangent vector that leads to the branch of walking gaits
is orthogonal to the switching-time dimension in $\Space$.  Setting $e_0 =
[0,0,0,0,1]^T$, then an orthonormal basis for the null space evaluated at
$J(\xeq, 0.62)$ is $\{e_0, [0.13, -0.12, 0.72, 0.67, 0]^T\}$ and at
$J(x_\text{eq}, 0.68)$ is $\{e_0, [0.13, -0.13, 0.69, 0.69, 0]^T\}$.  The
desired tangent in each case is not $e_0$ (which points along the switching-time
dimension), as shown in the proof of Proposition~\ref{prop:curves}.

Algorithm~\ref{alg:gm} then calls the pseudo-arclength continuation method of
Algorithm~\ref{alg:en} starting from each of the singular EGs identified in the
previous step.  Figure~\ref{fig:cgw-cc} shows the resulting set of gaits $\Gm$
and animations of a selection of gaits from $\Gm$, respectively.
%%%%%%%%%%%%%%%%%%%%%%%%%%%%%%%%%%%%%%%%
The path-connected set of gaits in Figure~\ref{fig:cgw-cc}(a)
reflects the description given earlier in Section~\ref{ssec:cc} of the connected
components of EGs.  In this plot, we have seven different 1D manifolds of gaits
(red and green curves) joined together at two singular EGs (black dots).  In
particular, $(\xeq, 0)$ is part of a set of equilibrium branches of gaits with
zero net displacement (red line) such that $(\xeq, \tau)$ is a point on these
branches for all $\tau$.  The red line of EGs intersects with two green branches
of walking gaits.  The points of intersection at $\tau = 0.62\,\text{s}$ and
$\tau = 0.68\,\text{s}$ for the compass gait correspond to the start of the
``short'' and ``long'' solution branches of walking gaits as reported in
\cite{Goswami1998}.  The (symmetric) green branches that extend from each
singular point contain gaits that are mirror images of each other, i.e., if one
branch has state $x_0$, the other branch has state $-x_0$; the sign indicates
whether the gait walks downhill to the left or right of its initial stance.

In order to generate the gaits in Figure~\ref{fig:cgw-cc}(a), we do not constrain the ground reaction force generated during a step between the stance foot and ground \cite{Bhounsule2014a}.  For gaits located on the thicker portions of the green curves, the ground reaction force pushes and pulls the stance foot during a step.
%%%%%%%%%%%%%%%%%%%%%%%%%%%%%%%%%%%%%%%%

In Figure~\ref{fig:cgw-cc}(b)--(g), we see the existence of gaits in $\Gm$ over
a range of slopes that include walking and overhand brachiating gaits.  Each of
the gaits in Figure~\ref{fig:cgw-cc}(b)--(g) can be continuously deformed into
each other and are part of the same connected component of the EG $(\xeq, 0)$
depicted in Figure~\ref{fig:cgw-model}(b).

%%%%%%%%%%%%%%%%%%%%%%%%%%%%%%%%%%%%%%%%%%%%%%%%%%%%%%%%%%%%%%%%%%%%%%%%%%%%%%%%

\section{Extensions to the Approach}
\label{sec:ext}
In the previous section, we focused on constant-control slices in $\G$.  We now
present extensions to the approach for
\begin{enumerate}
\item constructing multi-dimensional manifolds in $\Gm$ from EGs
(Section~\ref{ssec:multi}), and
\item searching the manifolds of $\G$ for gaits with desired properties, for
example, specific walking speeds or values for $x_0$, $\tau$, or $\mu$
(Section~\ref{ssec:ghm}).
\end{enumerate}

These extensions expand our work beyond passive dynamic walkers.  In particular,
for biped models that cannot balance on one foot, EGs for use as
input to Algorithm~\ref{alg:gm} may not exist or may be difficult to find.  In
order to handle these types of models, we add control parameters that
continuously modify the physical parameters of the biped model.  For example, in
Section~\ref{sec:ex}, we introduce a control parameter $\omega \in [0, 1]$ that
parameterizes a family of MARLO biped models with different hip widths and
center of mass positions such that $\omega = 0$ corresponds to a planarized
version of MARLO and $\omega = 1$ corresponds to the 3D model used in
\cite{Griffin2015}.  The purpose of parameters such as $\omega$ is to start with
a model that has a simple set of EGs that can be used as input to
Algorithm~\ref{alg:gm} (e.g, the EGs for MARLO at $\omega = 0$) and to then
eventually connect these gaits to a family of walking gaits of the desired biped
model (for MARLO, walking gaits with control parameter $\omega = 1$).  Given
that the number of control parameters $k$ will necessarily be greater than zero,
this motivates the use of algorithms that can search higher-dimensional
manifolds for desired gaits.

\subsection{Constructing Multi-Dimensional Manifolds}
\label{ssec:multi}
Algorithm~\ref{alg:multi} is an example of a higher-dimensional continuation
method.  It uses the map $M_0$ and a collection of $k$ additional maps $M_1,
\ldots, M_i, \ldots, M_k$ such that the level set $M_i(c) = 0$ defines a
constant slice in $\Space$ where the switching time and all but the
$i^\text{th}$ control parameter are fixed:
\begin{equation}
\begin{aligned}
M_i(c) &= [P^T(c), \Phi_i^T(c)]^T \\
\Phi_i(c) &= [\tau - t, \mu_1 - \upsilon_1, \ldots, \mu_{i-1} - \upsilon_{i-1}, \\
          & \quad \mu_{i+1} -  \upsilon_{i+1}, \ldots, \mu_k - \upsilon_k]^T,
\end{aligned}
\label{eq:Mi}
\end{equation}
where $c = \xtu \in \Space$ is the input; $1 \leq i \leq k$ is the $i^\text{th}$
control parameter in $\mu$ that varies throughout the continuation; $P$ is the
periodicity map; $\fun{\Phi_i}{\Space}{\R^k}$ defines the slice in $\Space$; $t
\in \R$ is a switching time; and $\mu_j$ and $\upsilon_j$ are the $j^\text{th}$
element ($j \in [1, k] - \{i\} \subset \mathbb{N}$) of the control parameter
vectors $\mu$ and $\upsilon$, respectively, such that $\mu_j$ is held fixed at
the value $\upsilon_j$.
%%%%%%%%%%%%%%%%%%%%%%%%%%%%%%%%%%%%%%%%
\begin{algorithm}[t]
\caption{Constructing multi-dimensional manifolds}
\begin{algorithmic}[1]
\Require singular equilibrium gait $\ceq \in E_0$.
\Function{multi-dim}{$d$}
\If {$d = 0$}
\State \textbf{return} $\{\ceq\}$
%\State \textbf{return} \Call{cmcurve}{$\ceq$, $\sd{\ceq}{s}$, $M_0$}
\Else
\State $\G_{d-1}$ = multi-dim($d-1$)
\State \textbf{return} $\bigcup\limits_{g \in \G_{d-1}}$\Call{cmcurve}{$g$,
$\sd{g}{s}$, $M_{d-1}$}
\EndIf
\EndFunction
\end{algorithmic}
\label{alg:multi}
\end{algorithm}

The algorithm generalizes Algorithm~\ref{alg:gm} by using $M_0$ and the $k$ maps
of Equation~\eqref{eq:Mi} to recursively construct a $(k+1)$-dimensional
manifold.  The algorithm recurses on the dimension $d$ ($0 \leq d \leq k+1$) of
the manifold.  The base case of $d = 0$ returns a singular EG, which is the seed
value for constructing a curve for the case of $d = 1$.  The recursive step
generates gaits for a $d$-dimensional manifold using the gaits on a
$(d-1)$-dimensional submanifold as seed values.

While this algorithm allows for the control space to vary throughout the
continuation, it becomes impractical for large $k$.  The algorithm is a brute-force approach to searching higher-dimensional manifolds for
desired gaits as we have to continue to run the algorithm until it happens to
come across a gait we are interested in.

\begin{comment}
\begin{myrem}
It is possible to construct $\Gm$ starting from a different slice in $\Space$.
The explicit use of $M_0$ in Algorithms~\ref{alg:gm}--\ref{alg:multi} can be
replaced with a user-defined map, say $\fun{M_\text{user}}{\Space}{\R^{2n+k}}$.
However, the use of equilibria and the indicator function $I$ to identify
singular points may be of limited use.
\end{myrem}
\end{comment}

%%%%%%%%%%%%%%%%%%%%%%%%%%%%%%%%%%%%%%%% Topology - Global Homotopy Map

\subsection{Finding Desired Gaits Using the Global Homotopy Map}
\label{ssec:ghm}
In this section, we give an algorithm for searching $(k+1)$-dimensional gait
manifolds for gaits with desired properties, such as gaits for walking on flat
ground.  A core part of the algorithm is the use of the global homotopy map
(GHM) \cite{Keller1978, Allgower1990} to find these gaits.  The GHM
continuously deforms gaits found using our previous map $M_0$ into gaits that
satisfy the periodicity constraints of the map $P$, and up to $k$ additional
constraints, by varying all parameters in $\Space$ at the same time.  The GHM
$G: \Space \rightarrow \Rh$ is 
\begin{equation}
\begin{gathered}
G(c) = H(c) - p \, H(a), \quad a \in \Gm - \pre{H}, \\
p(c) = (H^T(a)H(c))/(H^T(a)H(a)).
\end{gathered}
\label{eq:ghm}
\end{equation}
The map $\fun{H}{\Space}{\Rh}$ ($n_h \leq k$) is
similar to $\Phi$ of Equation~\eqref{eq:Meq0} with the exception that $H$ can
specify fewer than $k$ constraints.  
The gait $a \in \Gm - \pre{H}$ serves as a template motion that we attempt
to continuously deform into a desired gait---a to-be-determined point $c \in
\Space$ satisfying $H(c) = 0$. The parameter $p(c) \in [0, 1]$ is the
homotopy parameter that continuously deforms the reference gait $a$ into a gait
on the $H(c) = 0$ slice in $\Space$.  Two important properties of the homotopy
parameter are the following:
\begin{enumerate}
\item for $c = a$, we have $p(a) = 1$, which makes $a$ a trivial root of $G$
(i.e., $G(a) = H(a) - p(a) H(a) = H(a) - H(a) = 0$), and,
\item for points $c_0 \in \pre{H}$, we have $p(c_0) = 0$ and $G(c_0) = H(c_0) -
p(c_0) H(a) = H(c_0) = 0$ thus making $c_0$ roots of $p$ and $G$ as well.
\end{enumerate}

The GHM is a type of auxiliary function that is meant to be used to query the
gait space.  As an illustrative example, let $\sigma(c) \in \R$ compute the
incline of a planar biped's walking surface and $\nu(c) \in \R$ compute the
biped's average walking velocity, then the structure of the query is: Given the
manifold in $\G$ that contains the gait $a$, find a gait $c_0$ that walks on
flat ground ($\sigma(c_0) = 0$) at 0.7~m/s.  The constraint function $H =
[\sigma(c), \nu(c) - 0.7]^T = 0$ is used to encode the query as a set of
equality constraints.  We now integrate the GHM map into the rest of our
framework.

%%%%%%%%%%%%%%%%%%%%%%%%%%%%%%%%%%%%%%%%
\begin{comment}
\begin{algorithm}[t]
\caption{Constructing multi-dimensional manifolds}
\begin{algorithmic}[1]
\Require $\ceq \in E_0$.
\State \textbf{Populate $\Gm$ with gaits}:
\State Call Algorithm~\ref{alg:gm} with $\ceq$.
\State Resize $\Gm$ to a $(k+1)$-dimensional array.
\State \textbf{Iteratively build $(k+1)$-dimensional manifold}:
\For {$i := 1..k$ and each $g = (y_0, t, \upsilon) \in \Gm$} \label{alg:multi:for1}
\State Define map $\Phi_i(c) = [\tau - t, \mu_1 - \upsilon_1, \ldots,$
\State $\quad \mu_{i-1} - \upsilon_{i-1}, \mu_{i+1} -  \upsilon_{i+1}, \ldots,
\mu_k - \upsilon_k]^T$.
\State Define map $M_i(c) = [P^T(c), \Phi_i^T(c)]^T.$
\State Call Algorithm~\ref{alg:en} with map $M_i$ and gait $g$
\State Add gaits to $\Gm$.
\EndFor \label{alg:multi:for2}
\State \textbf{return} $\Gm$
\end{algorithmic}
\label{alg:multi}
\end{algorithm}
\end{comment}
%%%%%%%%%%%%%%%%%%%%%%%%%%%%%%%%%%%%%%%% GHM Algorithm
\begin{algorithm}[t]
\caption{A modified continuation method for use with the map $M_a$ of
Equation~\eqref{eq:Ma}}
\begin{algorithmic}[1]
\Require $\fun{M_a}{\R^{\nc}}{\R^{n_G}}$, $a \in \pre{M_a}$, $\alpha \in \R$,
and $\beta \in \R$.
\Function{ghm}{$c$}
\State \textbf{Assume}: $M_a(c) = 0$
\State Define merit function: $f(x) = \frac{1}{2} p(x)^2$.
\State \textbf{Newton Step}:
\State Solve for $\pd{c}{s}$ such that $\pd{M_a}{c}(c) \pd{c}{s} = 0$ and
\State $\pd{c}{s}^T\pd{c}{s} = \id_{k_G}$, where $\id_{k_G}$ is a $k_G \times k_G$ identity matrix.
\State $\Delta s = -\left[\pd{p}{c}^T(c) \pd{c}{s} \right]^\dagger p(c)$
\State \textbf{Perform Line Search}:
\State Set $m = 0$ and $\dot{c} = \pd{c}{s} \Delta s$.
\Repeat
\State Set $\lambda = \beta^m$, $m = m + 1$, and $h = \lambda \Delta s$
\State $z = $ \Call{cmstep}{$c$, $\dot{c}$, $h$} (see Algorithm~\ref{alg:en})
\Until {$f(z) - f(c) \leq -\alpha \lambda \pd{f}{s}^T(c, \dot{c}) \Delta s$}
\State \textbf{return} $z$
\EndFunction
\State \textbf{Generate Curve}:
\State Set $c[0] = a$.
\For {$i := 1..N$} 
\State $c[i] = $ \Call{ghm}{$c[i-1]$}
\EndFor
\State \textbf{return} the solution curve $c$
\end{algorithmic}
\label{alg:ghm}
\end{algorithm}
%%%%%%%%%%%%%%%%%%%%%%%%%%%%%%%%%%%%%%%%

Let $n_G = 2n + n_h$ be the number of total constraints and $k_G = k + 1 - n_h$
be the number of expected freedoms (as singularities on a connected component
can cause the number of freedoms to increase).  Given the GHM, define the map
$M_a: \G \rightarrow \R^{n_G}$ as 
\begin{equation}
\label{eq:Ma}
M_a(c) = [P^T(c), G^T(c)]^T.
\end{equation}
The space $\pre{M_a}$ is comprised of $k_G$-dimensional manifolds in $\G$.  The
goal is to find a path from a known gait $a \in \Gm$ to a gait $c \in \G$ such
that $H(c) = 0$.  From Equation~\eqref{eq:ghm}, this is equivalent to finding a
root of $p$.  In order to find a root of $p$, we modify Algorithm~\ref{alg:en}
so that we simultaneously generate $\Gm$ and search for a gait in $\G$ that is a
root of $p$.  The modified algorithm is summarized in Algorithm~\ref{alg:ghm}.

\begin{myprop}
\label{prop:Ma}
Given a point $a \in \Gm - \pre{H}$ sufficiently close to a root of $p$ and the
map $M_a$, if, at every iteration $i$ ($1 \leq i \leq N$) of
Algorithm~\ref{alg:ghm}, the tangent vector $\sd{c}{s}(s_i) \in
T_{c(s_i)}\pre{M_a}$ is chosen such that
\[
\sd{c}{s}(s_i) = -\pd{c}{s}(s_i) \left[\pd{p}{c}^T(c(s_i)) \pd{c}{s}(s_i)\right]^\dagger
p(c(s_i)), 
\]
where $\left[ \cdot \right]^{\dagger}$ is the Moore-Penrose inverse and
$\pd{c}{s}(s_i) \in \R^{(\nc) \times k_G}$ is a matrix whose $k_G$ columns are
basis vectors for $T_c\pre{M_a}$ at $c \in \pre{M_a}$, then tracing the vector
field $\sd{c}{s}(s_i)$ starting from $a$ simultaneously defines a
one-dimensional curve of fixed points $c(s_i) \in \pre{M_a} \subseteq \G$ and a
sequence of Newton iterates that converges to a root of $p$.
\end{myprop}

\begin{proof}
We prove the proposition in two steps.  First, we derive the direction of a
Newton step $\Delta s \in \R^{k_G}$ for numerically solving $p(c(s)) = 0$, where
$s \in \R^{k_G}$ is some parameterization of the manifold $M_a(c(s)) = 0$.  We
then project $\Delta s$ onto the tangent space $T_{c(s)}\pre{M_a}$ using
$\pd{c}{s}(s)$.

As the system of equations $p(c(s)) = 0$ is not square, we use the Moore-Penrose
inverse to define a Newton step $\Delta s$ \cite{Ben-Israel1966} from a Taylor
approximation of $p$ about a root; that is from $p(c(s+\Delta s)) \approx
p(c(s)) + \pd{p}{c}^T(s) \pd{c}{s}(s) \Delta s = 0$, we get $\Delta s =
-\left[\pd{p}{c}^T(c(s)) \pd{c}{s}(s)\right]^{\dagger} p(c(s))$.

At iteration $i$ of Algorithm~\ref{alg:ghm}, we project the Newton step onto the
tangent space $T_{c(s_i)}\pre{M_a}$ at $c(s_i) \in \pre{M_a}$, which yields the
tangent vector
\begin{equation*}
\sd{c}{s}(s_i) = -\pd{c}{s}(s_i) \underbrace{\left[\pd{p}{c}^T(c(s_i))
\pd{c}{s}(s_i)\right]^\dagger p(c(s_i))}_{-\Delta s}.
\end{equation*}
This choice leads to a curve being traced in $\pre{M_a}$ with points on the
curve that converge to a root of $p$.% \cite{Ben-Israel1966}.
\end{proof}

Algorithm~\ref{alg:ghm} contains the function \textproc{ghm} for computing a new
point $z \in \pre{M_a}$ from $c \in \pre{M_a}$ based on
Proposition~\ref{prop:Ma}.  In order to ensure that we are making progress
towards a root, we take a step of magnitude $h$ in the direction of
$\sd{c}{s}(s) = \pd{c}{s}(s) \Delta s$ based on an Armijo line search
\cite{Kelley1999} using the merit function $f(c(s)) = \frac{1}{2}p(c(s))^2$.
This leads to an adaptive step size strategy for $h$.  Typical values used for
$\alpha$ and $\beta$ in Algorithm~\ref{alg:ghm} are $10^{-4}$ and $0.5$,
respectively \cite{Kelley1999}.

\begin{comment}
\begin{myrem}
The homotopy parameter $p$ is usually defined as a free parameter.
Equation~\eqref{eq:ghm} uses the definition in \cite{Keller1978} and defines $p$
as a function of $c$ and $a$.  This reduces the number of free parameters we
have to consider during the continuation.
\end{myrem}
\end{comment}

\begin{myrem}
Algorithm~\ref{alg:ghm} differs from Algorithms~\ref{alg:en}--\ref{alg:multi} in
that the input map $M_a$ does not have to define one-dimensional level sets
(e.g., $M_0(c) = 0$ of Algorithm~\ref{alg:gm}).  Instead,
Algorithm~\ref{alg:ghm} selects a descent direction in a high-dimensional
tangent space using a merit function similar to many optimization-based methods
\cite{Nocedal1999, Betts2010}.
\end{myrem}

\begin{myrem}
\label{rem:ghm}
For biped models with sufficient control authority, we can seed the gait $a$ of
Algorithm~\ref{alg:ghm} with an EG that is a regular point of $P$.  A biped has
sufficient control authority if any regular EG has a local neighborhood of gaits
in $\G - E$ on the $(k+1)$-dimensional manifold it is on.  If not, we would have
to search for singular EGs as outlined in Section~\ref{ssec:I}.
\end{myrem}

\begin{myrem}
In Section~\ref{sec:ext}, we discussed using the control parameters to define a
family of 2D and 3D biped models.  An important application of the GHM is taking
the parameterized model and, for example, continuously deforming a gait for a
planar version of the model (where the EGs are trivial to specify) into gaits
for a 3D version of the biped model.
\end{myrem}

%%%%%%%%%%%%%%%%%%%%%%%%%%%%%%%%%%%%%%%% Examples

\section{Examples}
\label{sec:ex}
%%%%%%%%%%%%%%%%%%%%%%%%%%%%%%%%%%%%%%%% Figure for Examples
\begin{figure*}
\centering
\subfloat[Curved-foot walker
\cite{Martin2014}]{\includegraphics[width=0.48\textwidth]{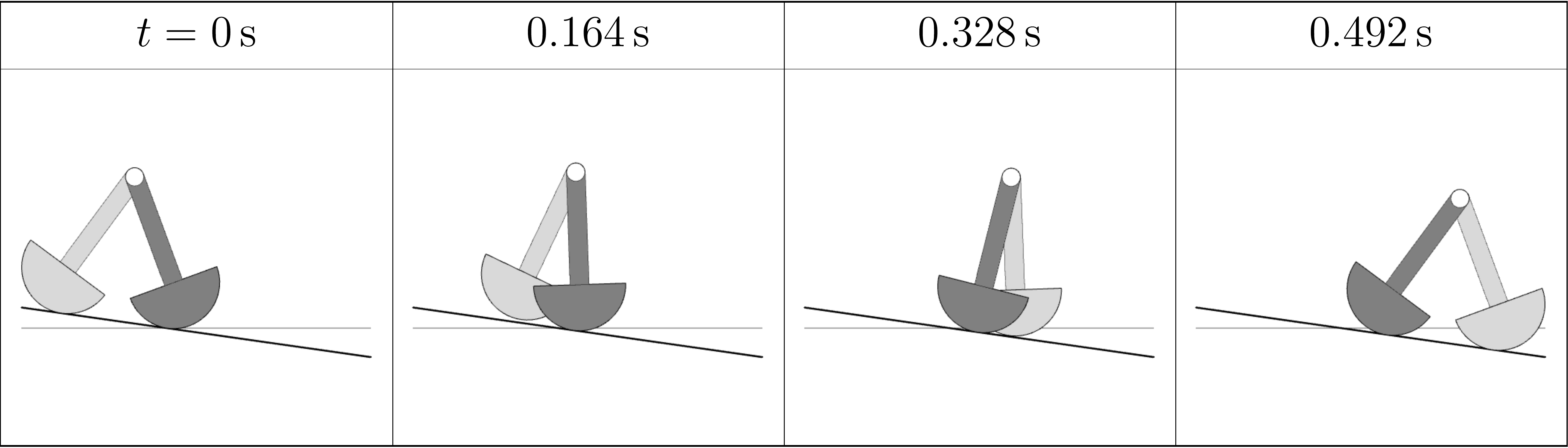}}
\hfill
\subfloat[Compass-gait with torso
\cite{LaHera2013,Rosa2014a}]{
\includegraphics[width=0.48\textwidth]{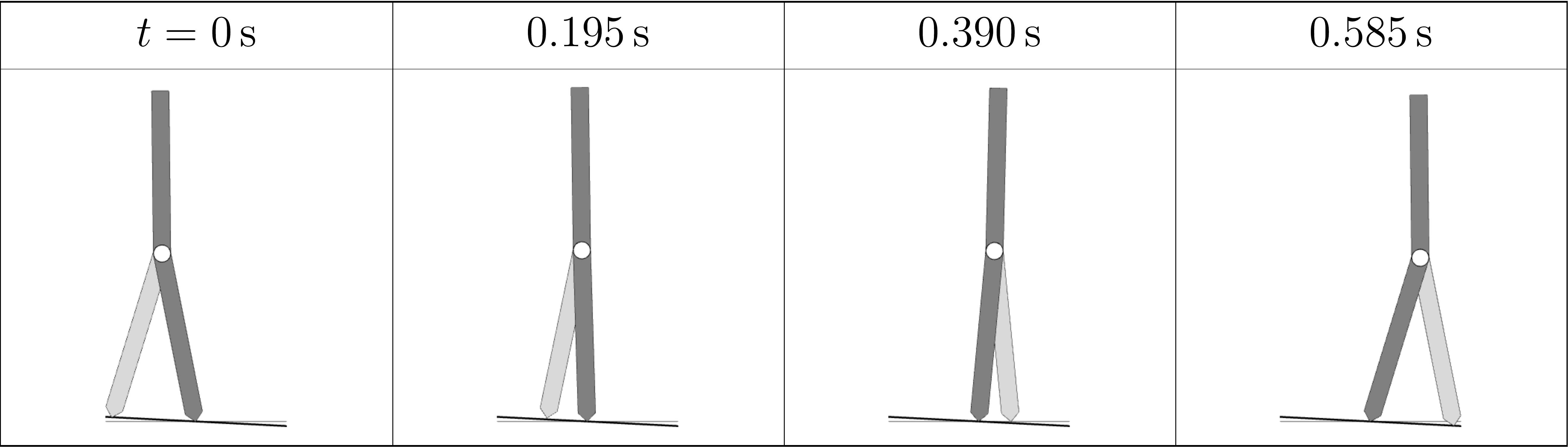}} \\
\subfloat[Point-foot kneed walker
\cite{Chen2007}]{\includegraphics[width=0.48\textwidth]{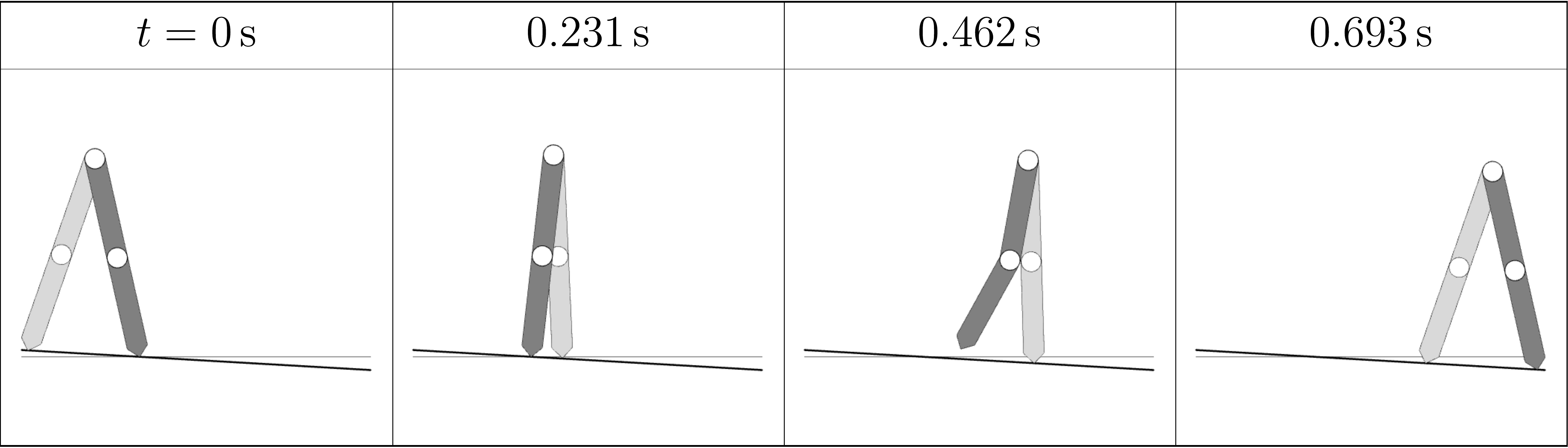}}
\hfill
\subfloat[Humanoid walker
\cite{Dumas2007,Rosa2014a}]{
\includegraphics[width=0.48\textwidth]{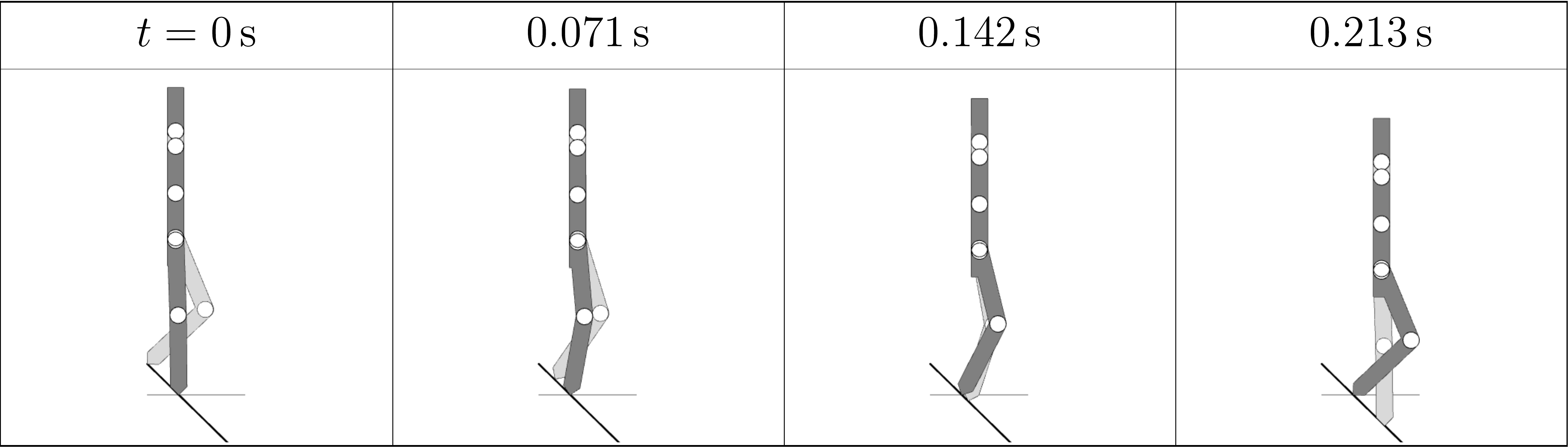}} \\
\subfloat[Five-link 3D walker
\cite{Chevallereau2008}]{\includegraphics[width=0.48\textwidth]{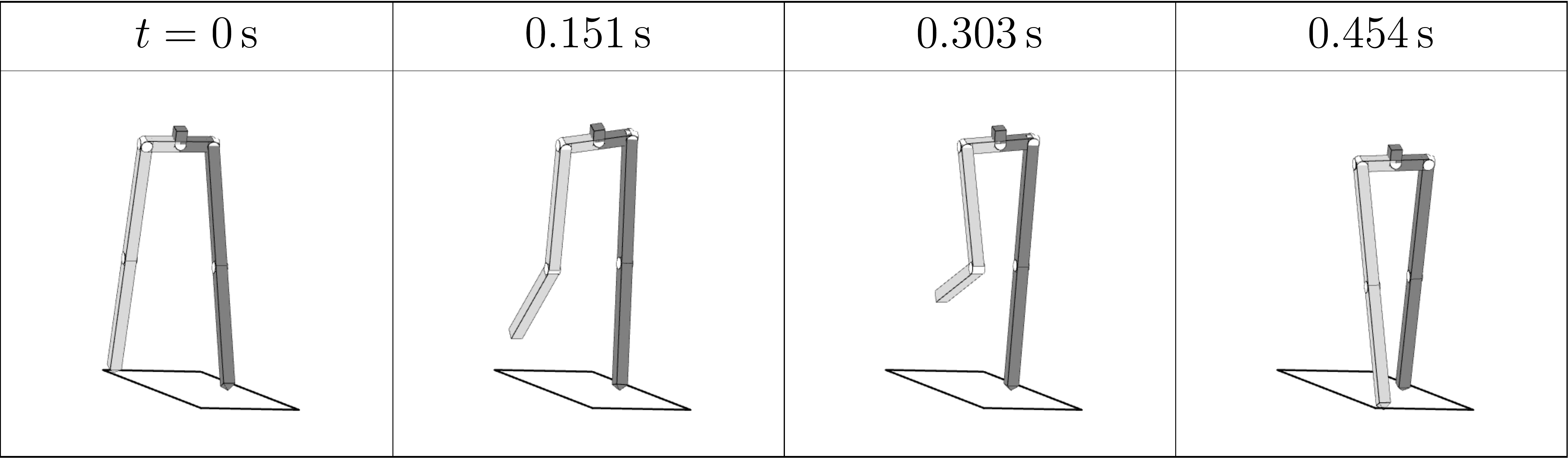}}
\hfill
\subfloat[MARLO biped
\cite{Griffin2015}]{\includegraphics[width=0.48\textwidth,height=72.48032pt]{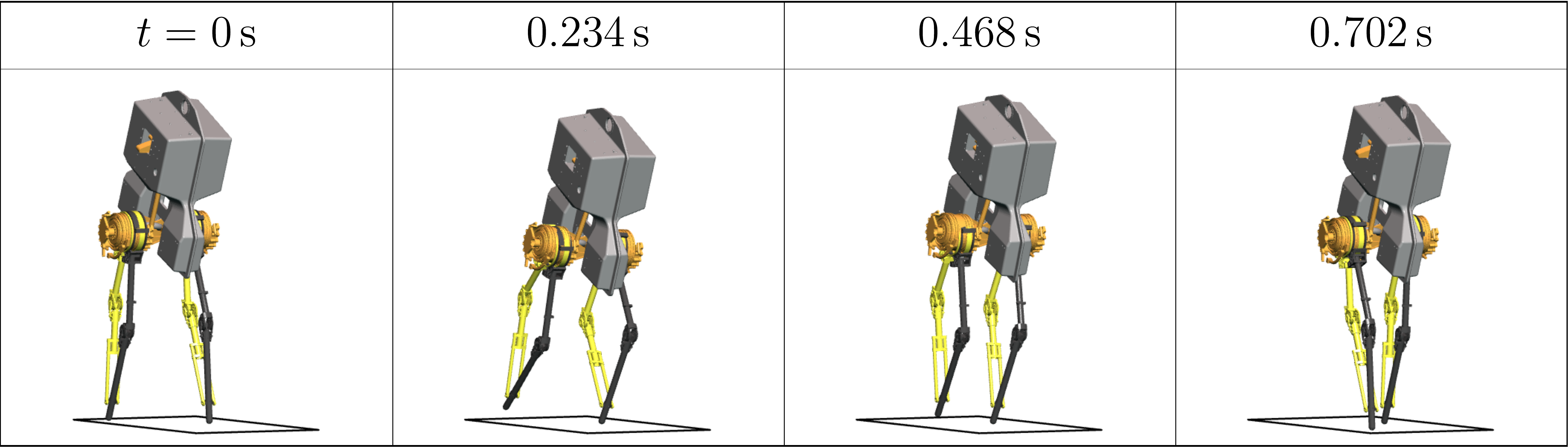}}
\\
\subfloat[Atlas biped
\cite{BostonDynamics2013}]{\includegraphics[width=1.0\textwidth]{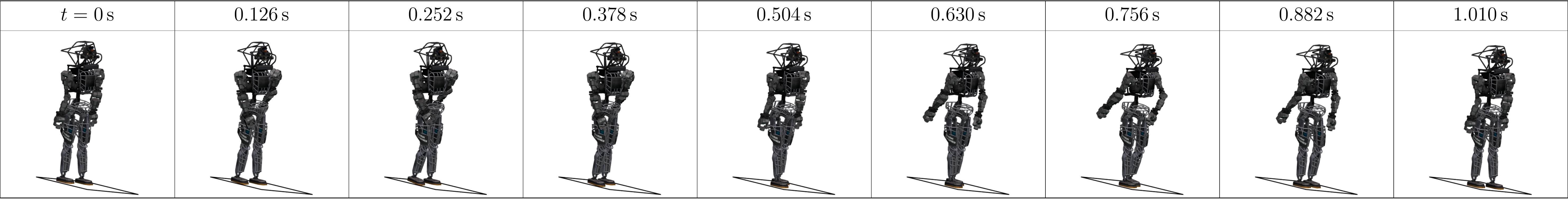}}
\caption{Example period-one gaits for various biped walkers.  The dots are joint
centers.  Gaits pictured in (a)-(d) are walking passively downhill and (e)-(g)
are powered gaits that are subject to virtual holonomic constraints; (e)-(f) are
walking on flat ground; (g) is walking downhill.}
\label{fig:models}
\end{figure*}
%%%%%%%%%%%%%%%%%%%%%%%%%%%%%%%%%%%%%%%%
We have applied our framework to several bipeds taken from the literature
(Figure~\ref{fig:models}) to confirm its wide applicability.  The biped models range from planar passive dynamic
walkers to high-degree-of-freedom actuated 3D humanoids.  The details of these
bipeds can be found in the Multimedia Material.  In this section, we expand on
using equilibria to generate actuated gaits for the compass gait
\cite{Goswami1998}, MARLO \cite{Ramezani2013} (Figure~\ref{fig:models}(f)), and
Atlas \cite{BostonDynamics2013} (Figure~\ref{fig:models}(g)) bipeds.

All bipeds are modeled as kinematic trees with floating bases attached to their
pelvis \cite{Featherstone2007}.  If the biped is planar the floating base has
three configuration variables, otherwise the floating base has six configuration
variables.  Unless otherwise noted, 
\begin{enumerate}
\item physical quantities are measured in SI units (i.e., meters, kilograms,
seconds),
\item the gaits reported in this paper have a normed error between two
consecutive pre-impact states $x_0 = (q, \dot{q})$ of less than $10^{-8}$ (with
$q$ having units of radians and $\dot{q}$ radians per second),
\item the search window for singular equilibrium gaits for
Algorithm~\ref{alg:sfp} is $\tau \in [0.1, 1]$ divided into 100 steps,
\item the step size $h$ of the NCM of Algorithm~\ref{alg:en} is $\pm
\frac{1}{20}$ in order to trace both sides of a curve.  We attempt to generate
$N = 250$ gaits per function call,
\item no constraints are placed on the ground reaction force,
\item \reply{joints that do not appear as free variables in the periodicity map $P$ have been solved for explicitly because they appear linearly in a VHC or PHC:}
\begin{itemize}
    \item \reply{for joints subject to VHCs, the feedback control laws are written so that the joint trajectories are periodic and a function of the free parameters in $\Space$ \cite{Westervelt2007}; and}
    \item \reply{for joints subject to PHCs, especially the Cartesian floating base joints, the values of these joints can be determined from the other parameters in the PHCs,}
\end{itemize}
\item \reply{boundary \Bez coefficients in a VHC are written in terms of $x_0$ and satisfy periodicity constraints},
\item the $x$, $y$, and $z$ axes of the world and local frames of our 3D biped
models are labeled with blue, red, and green arrows, respectively, and
\item gaits are computed on a $2.7$~GHz Intel Core i7-4800MQ CPU laptop running $64$-bit Ubuntu $18.04$ LTS.
\end{enumerate}

The Multimedia Material contains an implementation of our framework as a Mathematica library.  Given
model-specific information, like the biped's PHCs and VHCs, the code is capable
of finding entire families of walking gaits using only equilibria of the biped
models.  The code computes items like the map $P$, the flow $\varphi$, and their
derivatives from the model.

\begin{comment}
 in Mathematica v11.3.0; certain pieces of Mathematica code were compiled to C using Mathematica's compiler interface and a gcc 7.4.0 C compiler

We briefly mention that we constructed $\Gm$ in a very similar fashion for each
of the passive dynamic walkers in Figure~\ref{fig:models}(a)--(d).  The only
exception is the point-foot kneed walker (Figure~\ref{fig:models}(c)), which has
two switching times.  The details are outside the scope of this paper; the code
can be found in the Multimedia Material.

Our general approach to generating gaits for 3D bipeds
(Figure~\ref{fig:models}(e)--(g)) is a two-step process:
\begin{enumerate}
\item First we generate gaits for a planar version of the biped using
Algorithm~\ref{alg:en}.
\item Then we apply a GHM (Section~\ref{ssec:ghm}) to continuously deform a gait
of the planar model into a gait of the 3D model using Algorithm~\ref{alg:ghm}.
\end{enumerate}
The first step is primarily for the case where the 3D biped model does not have
the control authority to achieve a one-footed equilibrium from which to begin
generating gaits.  The second step in our approach is straightforward: the GHM
enables us to search for a viable gait for the 3D biped model using a gait for
the planar model as the seed value for the GHM.  We provide more specifics on
how we implement these steps with the actual biped models.
\end{comment}

%%%%%%%%%%%%%%%%%%%%%%%%%%%%%%%%%%%%%%%% Examples - Compass Gait Actuated

\subsection{Extending Passive Gaits into a Family of Actuated Gaits}
\label{ssec:acgw}
If we add a motor at the hip joint of the compass-gait robot as described in
Section~\ref{ssec:cgw}, the state-time space $\Space$ can be augmented with
control dimensions.  In general, increasing the dimension of $\Space$ will also
increase the dimension of $\G$.  The original $\G$ is just the $\mu = 0$ slice
of the extended $\G$.  

\begin{figure}[t]
\centering
\subfloat[]{\includegraphics[width=0.40\textwidth]{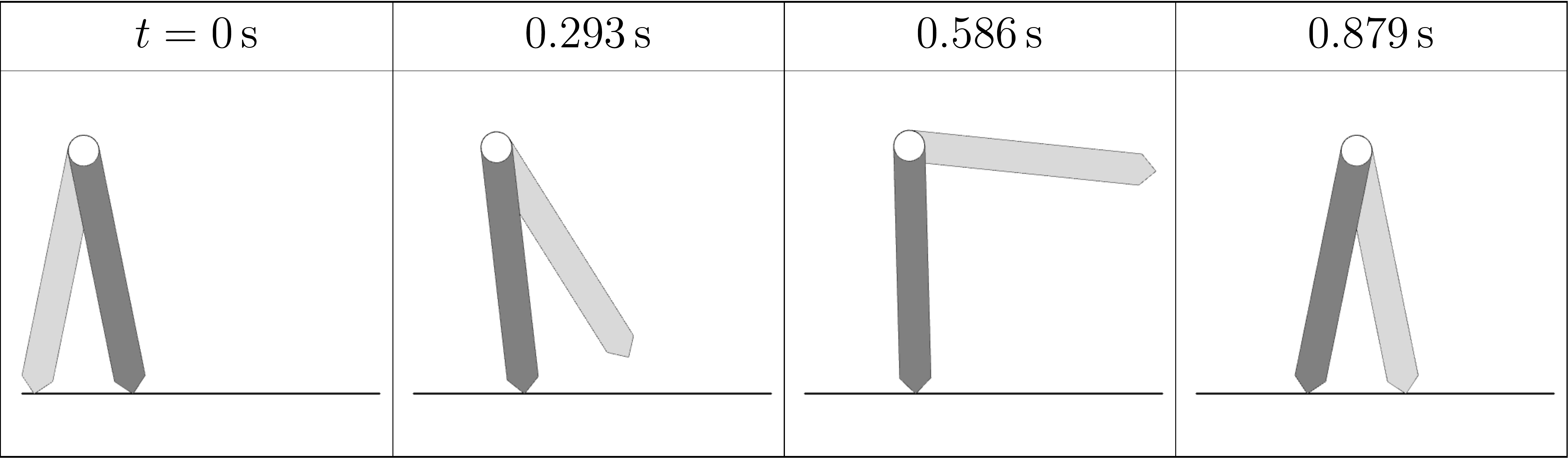}} \\
\subfloat[]{\includegraphics[width=0.45\textwidth]{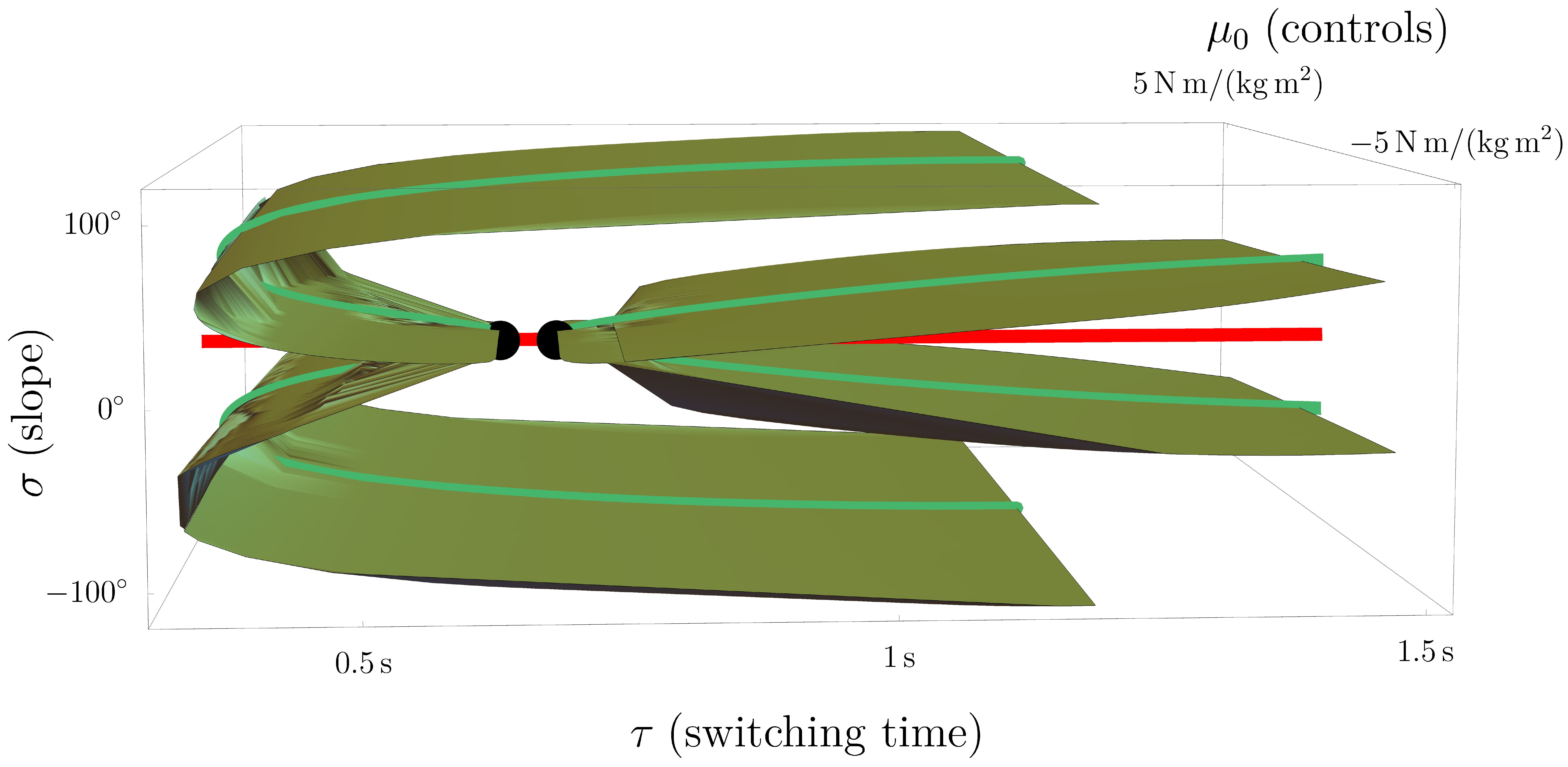}}
\caption{(a) An actuated gait of the compass-gait walker walking on flat ground,
where $u(t) = \mu_0 \sin(\omega t)$ with $\omega = 2 \pi$.  For this step,
$\mu_0 = -5.34$~N\,m/(kg\,m$^2$).  At the start of the next step, $\mu_0 =
5.34$~N\,m/(kg\,m$^2$). (b) A connected component of the compass-gait walker in
a higher-dimensional state-time-control space (see Figure~\ref{fig:S} for
legend).  The curves of passive gaits in Figure~\ref{fig:cgw-cc} are slices of
this space where $\mu_0 = 0$ (green and red curves).}
\label{fig:cgw-actuated-gait}
\end{figure}

We use the motor to drive leg 2 relative to leg 1 with the torque $u_0(t) =
\mu_0 \sin(\omega t)$ \cite{Iida2009}, where the amplitude $\mu_0 \in \R$ is a
control parameter.  The angular frequency is fixed at $\omega = 2 \pi$ to keep
this example low dimensional.  The control dimension is one ($k = 1$) and we
have added an actuator ($n_u = 1$) to the model.  Design parameters could also
be defined.  For example, a parameter defining the curvature of the feet
\cite{Martin2014} or position of the center of mass \cite{McGeer1990a} could be
added.  In this example, however, we only add the control parameter $\mu_0$.
Figure~\ref{fig:cgw-actuated-gait}(a) shows an actuated gait in the extended
gait space.  The value of $\mu_0$ for this step is $-5.34$~N\,m/(kg\,m$^2$) and
becomes $5.34$~N\,m/(kg\,m$^2$) at the start of the next step.  As masses and
lengths are scaled in \reply{our reference model \cite{Goswami1998}}, this corresponds to a maximum output torque of
about $\frac{\mu_0}{4} = 1.34$~N\,m.

In this six-dimensional state-time-control space $\Space$ consisting of points
$(x_0, \tau, \mu_0)$, we have two-dimensional gait manifolds (the six parameters
minus the four periodicity constraints).  The passive gaits of the previous
section are now a slice of this higher-dimensional gait space $\G \subset \R^6$,
where the control parameter $\mu_0$ is zero
(Figure~\ref{fig:cgw-actuated-gait}(b)).

Algorithm~\ref{alg:multi} is used to construct the surface of
Figure~\ref{fig:cgw-actuated-gait}(b).  The algorithm first uses the map $M_0$
of Section~\ref{ssec:I} to generate a set of passive gaits using the seed values
$(\xeq, 0.62, 0)$ and $(\xeq, 0.68, 0)$ (i.e., the singular points of the
original $\G$ space mapped to the extended $\G$ space).  Our Mathematica implementation of the algorithm took $7.2$~minutes to compute the $828$ passive gaits in Figure~\ref{fig:cgw-actuated-gait}(b).
Algorithm~\ref{alg:multi} then uses every gait found in $\pre{M_0}$ as seed
values to Algorithm~\ref{alg:en} using the map $\fun{M_1}{\Space}{\R^{2n+k}}$ of
Equation~\eqref{eq:Mi}, which holds $\tau$ constant and allows $\mu_0$ to vary
during the continuation.  The library took $1.8$~hours to compute the $18,400$ actuated gaits in Figure~\ref{fig:cgw-actuated-gait}(b).

This higher-dimensional example shows that we can grow $\Gm$ from an EG to a set
of passive gaits to an even larger set of actuated gaits by adding extra control
parameters to the state-time-control space $\Space$.  We use this notion of
growing a manifold from lower-dimensional slices for our 3D bipeds as well.  In
this case, we generate gaits for a planar or simplified 3D version of the biped
and use these walking gaits to generate gaits for the full 3D model.

%%%%%%%%%%%%%%%%%%%%%%%%%%%%%%%%%%%%%%%% Examples - Atlas

\subsection{Generating Gaits for a Flat-footed Walking Biped with Arms}
\label{ssec:bda}
\begin{figure}[t]
\centering
\includegraphics[width=0.34\textwidth]{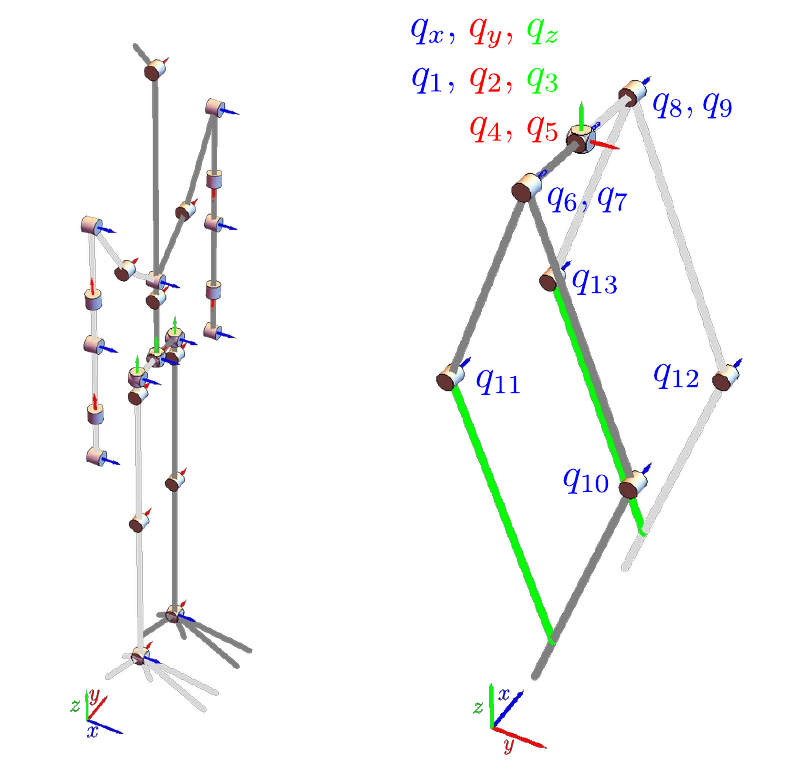}
\caption{Mechanical structure and coordinates for the models of Atlas
(left) and MARLO (right).  The light gray links are on the ``right-side'' of the
bipeds, the cylinders are joints, the arrows emanating from the cylinders are
the (positive) rotation axes, and the $x$-$y$-$z$ frames are world frames.  In
addition, MARLO has joint centers that overlap and are not visible, and green
links to emphasize the leg's four-bar structure.}
\label{fig:biped-models3D}
\end{figure}

%%%%%%%%%%%%%%%%%%%%%%%%%%%%%%%%%%%%%%%%
\begin{comment}
%%%%%%%%%%%%%%%%%%%%%%%%%%%%%%%%%%%%%%%%
\begin{figure}[t]
\centering
\includegraphics[width=0.48\textwidth]{cgw-traj-u-1}
\caption{An actuated gait of the compass-gait walker walking on flat ground,
where $u(t) = \mu_0 \sin(\omega t)$ with $\omega = 2 \pi$.  For this step,
$\mu_0 = -5.34$~N\,m/(kg\,m$^2$).  At the start of the next step, $\mu_0 =
5.34$~N\,m/(kg\,m$^2$).}
\label{fig:cgw-actuated-gait}
\end{figure}
%%%%%%%%%%%%%%%%%%%%%%%%%%%%%%%%%%%%%%%%
\begin{figure}[t]
\centering
\includegraphics[width=0.48\textwidth]{cgw-cc-powered}
\caption{A connected component of the compass-gait walker in a
higher-dimensional state-time-control space (see Figure~\ref{fig:S} for legend).
The curves of passive gaits in Figure~\ref{fig:cgw-cc} are slices of this space
where $\mu_0 = 0$ (green and red curves).}
\label{fig:cgw2D}
\end{figure}
%%%%%%%%%%%%%%%%%%%%%%%%%%%%%%%%%%%%%%%%
\end{comment}
\reply{We have tested our technique on Boston Dynamics'
Atlas}, a 3D flat-footed walking biped
(Figure~\ref{fig:biped-models3D}(a)).  The use of flat-footed walking
constraints is inspired by the reduced models of \cite{Posa2016, Hereid2018},
which only consider the legs of Atlas as their biped model.  In this example,
we generate gaits for a biped with fewer actuators than internal degrees of
freedoms ($n_u < n - 6$).  This example demonstrates that we
can generate gaits for complicated underactuated bipeds subject to VHCs.

The model we use for Atlas is from a DARPA Robotics Challenge ``.cfg''
file available online \cite{OSRF2014}.  This source file generates the biped's
URDF file for use in ROS and Gazebo.  We use the full model with 3D
dynamics and a 6D floating base at the pelvis.  The only changes to the model
are
\begin{enumerate}
\item zeroing the $y$ position of the center of mass of two links (see
Multimedia Material) to make the model physically symmetric
(Assumption~\ref{as:sym}), and 
\item defining the biped's home position with the arms pointing downwards
(Figure~\ref{fig:biped-models3D}(a)), so that $\xeq = 0 \in \X$ can be an
equilibrium point.
\end{enumerate}

In the end, our model of Atlas has 34 degrees of freedom when unconstrained
(Figure~\ref{fig:biped-models3D}(a)).  The model has 15 VHCs and 15 actuators.
However, the VHCs and actuators are not all active during a step.  When a VHC is
not active, an actuator is considered off and the joint attached to the inactive
actuator is unactuated during the step.

During a step, the biped walks in its sagittal plane and is subject to six PHCs
that fix the stance foot to the ground ($n_p=6$), 13 active VHCs ($n_v = 13$),
and 13 active actuators to enforce the VHCs ($n_u = 13$).  Atlas' remaining 15
internal joints that are not subject to VHCs as well as the $6$ joints of the
floating base are unactuated.

The set of active VHCs depends on whether the left or right leg is the stance leg.  At the
post-impact time $t = 0^+$, we assume the left leg is the stance leg.  The
configurations and velocities of the joints about the $x$ and $z$ axes at the
hips, $y$ axis at the neck, $x$ and $y$ axes at the ankles, $y$ axis at the
right elbow, and $x$ axis at the wrists are subject to VHCs that force the
joints to track third-order \Bezs[.] The configuration of the left knee joint is
subject to the final VHC that keeps the relative knee angle locked at zero
degrees throughout the motion post-impact.  Further details, like the active set
of VHCs when the right leg is the stance leg, can be found in the Multimedia
Material.
%%%%%%%%%%%%%%%%%%%%%%%%%%%%%%%%%%%%%%%%
%%%%%%%%%%%%%%%%%%%%%%%%%%%%%%%%%%%%%%%%

From this data, the biped's pre-impact state is $x_0 \in \X \subset \R^{68}$ ($n
= 34$).  In our code, we use the PHCs and VHCs to reduce the number of
independent states in $x_0$ down to 33 state variables and define $x_0$ as a
function of a reduced state vector $\bar{x}_0 \in \R^{33}$ such that $x_0 =
x_0(\bar{x}_0)$.

In addition to the states, we define two control parameters $\omega_1, \omega_2
\in [0, 1]$, making $\mu = [\omega_1, \omega_2]^T \in \Mu$.  These dimensionless
control parameters modify the biped's physical parameters to create a 3D model
that can balance on one foot at $x_0 = 0$.  The parameter $\omega_1$ affects
every link's center of mass position and relative distance from the pelvis along
the link's $x$ axis (i.e., if the distance of link $i$ from the pelvis along the
$x$ axis was $\delta x_i$, then the model would contain the product $\omega_1
\delta x_i$).  The parameter $\omega_2$ affects the center of mass position
along the $y$ axis of every link's center of mass.  When both parameters are
zero, they, along with the active VHCs, eliminate the non-zero moments about the
joints due to the gravity vector making the configuration depicted in
Figure~\ref{fig:biped-models3D}(a) an equilibrium for the model.  A value of one
for each parameter gives the original biped model.

In total, the biped has a 71-dimensional state-time-control space $\Space$ (68
states, 2 control parameters, and 1 switching time).  We define $M_0$ of
Equation~\eqref{eq:M0} so that $\omega_1 = \omega_2 = 0$ throughout the
continuation making $\mu_0 = 0$ in $M_0$.  At these fixed values of the control
parameters, we generate gaits for Atlas using the same process as we did with
the compass gait.  

For the parameterized model at $\omega_1 = \omega_2 = 0$, a singular EG found in
the constant-control slice occurs at $\tau = 0.396$~s.  Algorithm~\ref{alg:gm}
performs a NCM using the map $M_0$ starting at the singular EG at $\tau =
0.396$~s.  Our Mathematica code took $1.5$~hours to compute $250$ gaits.  We then apply Algorithm~\ref{alg:ghm} to find a gait with $\omega_1 =
\omega_2 = 1$.  The library took $1.8$~hours to find a gait with the desired values and computed $48$ gaits in the process.  The desired gait from this continuation is shown in
Figure~\ref{fig:models}(g).  The biped is shown taking two steps.

%%%%%%%%%%%%%%%%%%%%%%%%%%%%%%%%%%%%%%%% Examples - MARLO

\subsection{Incorporating Inequality Constraints}

Our final example gives an in-depth overview of using the GHM and
Algorithm~\ref{alg:ghm}.  In this example, we use the University of Michigan's MARLO \cite{Ramezani2013, Griffin2015} (Figure~\ref{fig:biped-models3D}(b)) to
demonstrate how to incorporate inequality constraints into a continuation.  The
biped is part of a line of ATRIAS bipeds developed at Oregon State
University \cite{Hubicki2016}.  The hybrid dynamics of the model is detailed in
\cite{Ramezani2013}.  We do not model the series-elastic actuators, but do take
into account the mass of the actuators.  The physical parameters for our model
are taken from the source code in \cite{Griffin2016}, which is used in
\cite{Griffin2015}.

MARLO has 16 degrees-of-freedom (DOFs) when no constraints are
applied.  The biped walks in its sagittal plane with gravity
pointing downward.  Referring to
Figure~\ref{fig:biped-models3D}(b), the joints of the biped are a 6-DOF floating
base (where $q_1$--$q_3$ are the roll, pitch, and yaw angles, respectively), two
hip joints for out-of-plane leg rotation ($q_4$--$q_5$), and eight joints for
the two four-bar mechanisms serving as legs for the biped ($q_6$--$q_{13}$).
When constraints are applied, MARLO has seven PHCs ($n_p = 7$).  Three of the
constraints keep the stance foot (a point) stationary and the other four
constraints are the four-bar linkage constraints on each leg:
\begin{equation*}
\begin{aligned}
q_6 + q_{10} - q_7 = 0 & & q_7 + q_{11} - q_6 = 0 \\
q_8 + q_{12} - q_9 = 0 & & q_9 + q_{14} - q_8 = 0.
\end{aligned}
\end{equation*}
Given these constraints, we can describe a pre-impact state $x_0 \in \X$ of the
biped using 18 numbers, the nine joint angles $q_1$--$q_9$ and their respective
velocities ($n = 9$).
%%%%%%%%%%%%%%%%%%%%%%%%%%%%%%%%%%%%%%%%
\begin{figure*}[t]
\centering
\subfloat[$\omega =
0$]{\includegraphics[width=0.48\textwidth]{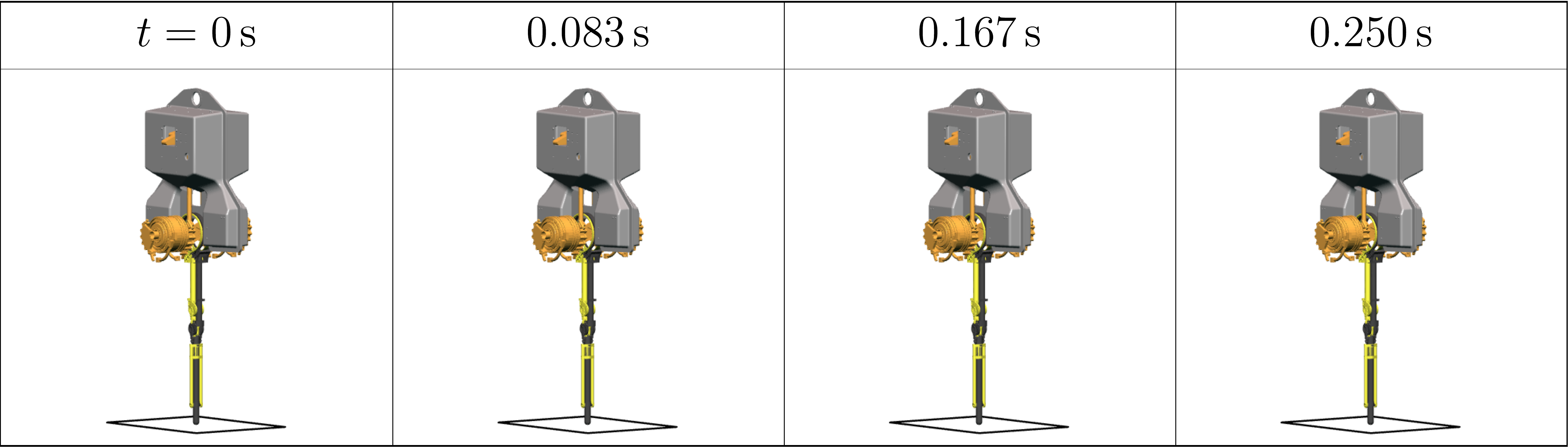}} \hfill
\subfloat[$\omega =
0.37$]{\includegraphics[width=0.48\textwidth]{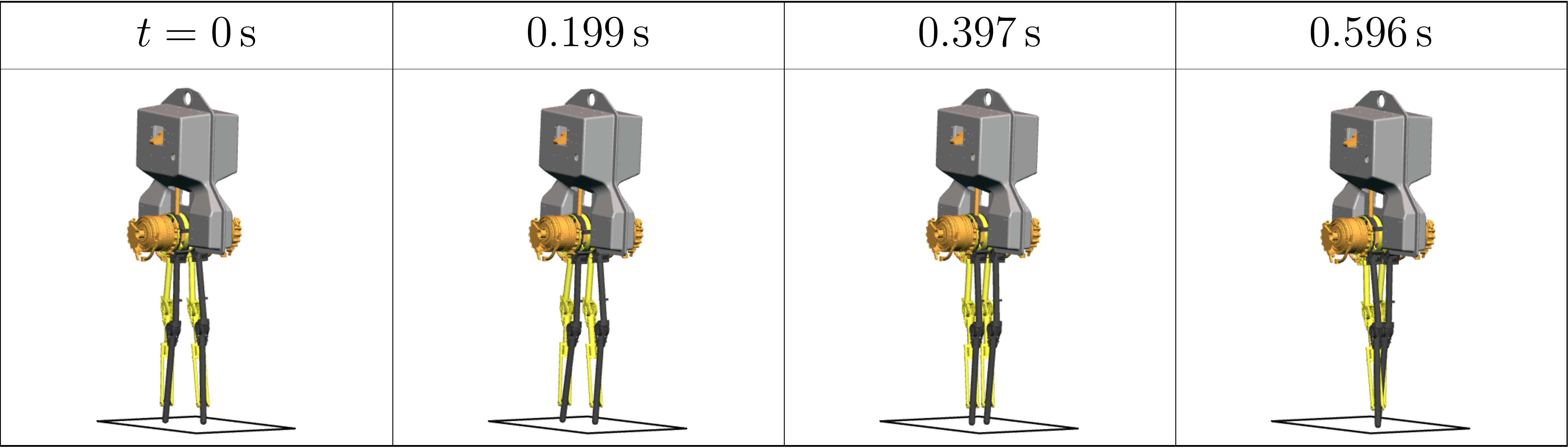}} \\
\subfloat[$\omega =
0.67$]{\includegraphics[width=0.48\textwidth]{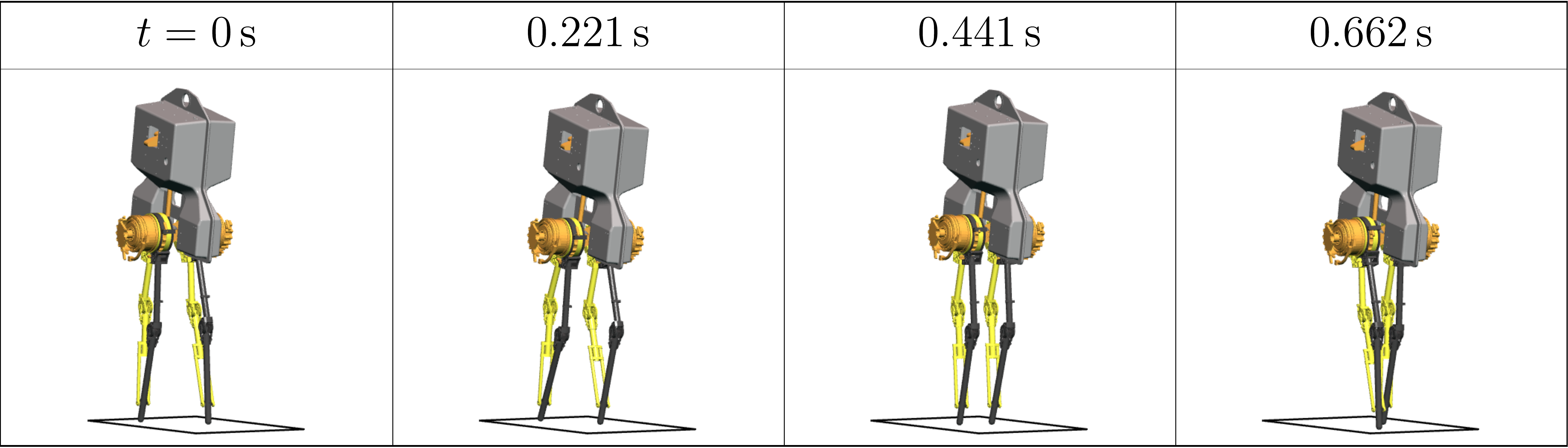}} \hfill
\subfloat[$\omega =
1$]{\includegraphics[width=0.48\textwidth]{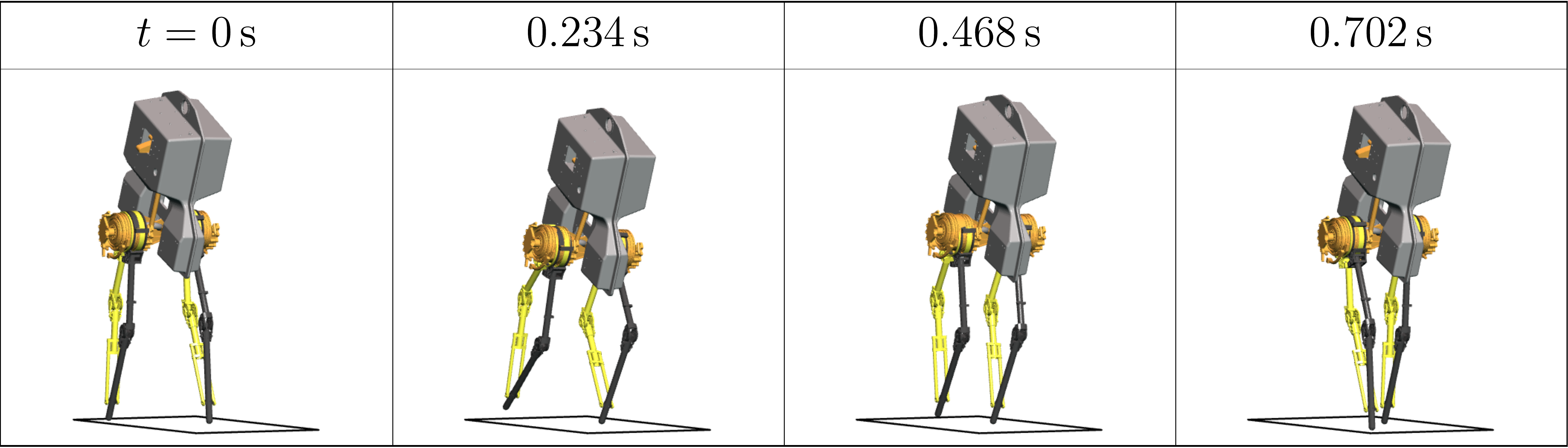}}
\caption{The effect of the parameter $\omega$ on the hip width of MARLO.  The
parameter also affects the position of the center of mass of each link (not
shown).}
\label{fig:marlo-w}
\end{figure*}

The biped has six actuators that drive the robot's leg joints $q_4$--$q_9$ ($n_u
= 6$) and is subject to six virtual holonomic constraints ($n_v = 6$).  When the
left leg is in stance, the constraints are
\begin{equation*}
\label{eq:marlo-vhc}
\begin{aligned}
q_4 - b_4^3(\theta, a) = 0 & & q_5 - b_5^3(\theta, a) = 0  & & q_6 - b_6^3(\theta, a) = 0 \\
q_{10} - b_{10}^3(\theta, a) = 0 & & q_8 - b_8^4(\theta, a) = 0  & & q_{12} - b_{12}^4(\theta,
a) = 0,
\end{aligned}
\end{equation*}
and similarly when the right leg is in stance.

During a step, the VHCs force the hip, stance thigh, and lower leg to track
third-order \Bezs and the swing thigh and lower leg to track fourth-order
\Bezs[.]  The two fourth-order polynomials $b_8^4(\theta, a)$ and
$b_{12}^4(\theta, a)$ each have a free coefficient that is not determined by the
periodicity boundary constraints.  The two free coefficients, say $\alpha_1,
\alpha_2 \in \R$, correspond to control dimensions in $\Space$.  \reply{The continuation will determine values for these coefficients for each gait generated.}

Overall, there are six control parameters $\mu = [\omega, s_1, s_2, s_3,
\alpha_1, \alpha_2]^T \in \Mu \subseteq \Rmu$ ($k = 6$), explained below.  The
dimensionless parameter $\omega \in \R$ continuously deforms the physical
parameters of the biped from a planar model into a 3D model by controlling the
hip width and the position of the center of mass of each link on the biped (in
other words, if the hip width is defined by the physical parameter
$\ell_\text{hip}$, then the model would have its hip width defined as the
product $\omega \ell_\text{hip}$).  Figure~\ref{fig:marlo-w} depicts how
$\omega$ affects the biped's hip width; it does not show the effects on the
center of mass of each link.  At $\omega = 0$, the biped has zero hip width and
all of the center of masses are projected onto their respective links.  When
$\omega = 1$, the original values for the biped's hip width and link center of
masses are restored.

Finally, the vector $\mu$ has three slack variables $s_1, s_2, s_3 \in \R$ with
lower bounds $s_1, s_2, s_3 \geq 0$; there are no upper bounds on the variables.
These constraints are treated as box constraints \cite{Nocedal1999}.  We use the
slack variables $s_1$ and $s_2$ so that we only search for gaits where the knee
joints $q_{10}$ and $q_{12}$ are nonnegative pre-impact.  This is sufficient to
avoid knee hyperextension as the biped takes its step ($q_{10}(t), q_{12}(t)
\leq 0$ for all $t$).  The third slack variable $s_3$ is used to make the biped
walk from left to right by imposing a forward velocity constraint on the biped's
floating base $q_y \in \R$ coordinate.

We also impose integral inequality constraints.  Let $p(t) = p(\flow[t]) \in
\R^3$ represent the location of the swing foot in space, and $\dist(p(t)) \in
\R$ be a distance function that is positive when the swing foot is above the
surface, zero when the swing foot is on the surface, and negative when the swing
foot is below the surface.  The goal is for the swing foot $p(t)$ to never go
below the walking surface.  To avoid foot penetration with the ground, we
require $\dist(p(t)) \geq 0$ for all $t \in \R$.  This is equivalent to finding
the zeros of $\ad{0}{\tau}{[\dist(p(t))]^-}{t}$, where $[x]^-$ returns $x$ if $x
\leq 0$ and zero otherwise.

Given the model and its PHCs and VHCs, the resulting state-time-control
space $\Space$ is 25-dimensional (18 state variables, six design and control
parameters, and one switching time).  The biped's periodicity map $P$ is
\begin{equation*}
\begin{aligned}
P(c) &= \begin{bmatrix} Q_1 \\ \dot{Q}_1  \end{bmatrix}, &
Q_1 &= \begin{bmatrix} q_1(\tau^-) - q_1(0^-) \\  q_2(\tau^-) - q_2(0^-) \\
q_3(\tau^-) - q_3(0^-) \\  \end{bmatrix},
\end{aligned}
\end{equation*}
which states that the roll, pitch, and yaw angles of the floating base have to
be periodic.  These are the only angles that are unactuated; all other angles
are subject to VHCs which we can design to satisfy the periodicity condition
of the (virtually) constrained joints \cite{Westervelt2007}.  \reply{We note that while we do not write periodicity constraints in $P$ for joint angles $q_4$-$q_9$ and their velocities, they are still free parameters in $\Space$ that affect the motion of the base and the four-bar linkage.  The numerical continuation determines their values for each gait generated.}

The manifolds in the gait space $\G$ are 19-dimensional (25 state-time-control
variables minus six periodicity constraints).  We cannot apply
Algorithm~\ref{alg:sfp} because $\pd{P}{x_0}(c) \in \R^{6 \times 18}$ is not a
square matrix.  For MARLO, we demonstrate the utility of searching for
locomoting gaits on a high-dimensional manifold using a GHM.  
As an additional benefit of the GHM, we can start from any EG with an
arbitrarily chosen switching time $\tau$ provided the equilibrium gait is not a
root of Equation~\eqref{eq:ghm}.

The primary goal of using the GHM is to generate a gait of the 3D model starting
from an EG of the planarized version of biped model ($\omega = 0$).  To achieve
this goal, we instantiate the map of Equation~\eqref{eq:Ma} as
$\fun{M_a}{\Space}{\R^{18}}$ such that
\begin{equation*}
\begin{gathered}
M_a(c) = [P^T(c), \Phi^T(c), H^T(c)-p(c)H^T(a)]^T, \\
\Phi(c) = \left[\begin{smallmatrix} q_6(0^-)-q_7(0^-) - s_1 \\ q_8(0^-)-q_9(0^-) - s_2
\\ \dot{q}_y(0^-) - s_3 \\ q_1(0^-) \\ q_2(0^-) \\ q_4(0^-) \\ q_5(0^-)
\end{smallmatrix}\right], H(c) = \left[\begin{smallmatrix} s_1 - s_{1, \text{des}} \\ s_2 - s_{2,
\text{des}} \\ \sigma(c) - \sigma_\text{des} \\ q_y(\tau^-) - q_y(0^-) - q_{y,
\text{des}} \\ \ad{0}{\tau}{[\dist(p(t))]^-}{t} \\ \omega - 1
\end{smallmatrix}\right],
\end{gathered}
\end{equation*}
where the map $P$ is the biped's periodicity map; $p(c) \in \R$ is the
homotopy parameter (Equation~\eqref{eq:ghm}); $\fun{\Phi}{\Space}{\R^7}$ is a map
of three inequality constraints and four equality constraints, where the first two
constraints ensure the left and right knee joints bend inward in an
anthropormorphic manner, the third constraint ensures the biped travels from
left to right with a positive forward velocity, and the last four constraints
ensure the out-of-plane angles (including those of the floating base) start and
end at zero degrees; and, finally, the map $\fun{H}{\Space}{\R^6}$ is used to
find a gait with a desired knee bend ($s_{1, \text{des}} = s_{2, \text{des}} =
20^\circ$), walking surface ($\sigma_\text{des} = 0 \in \R^2$, i.e., a flat
ground), step length ($q_{y, \text{des}} = 0.5$\,m), minimum height above the
ground for the swing foot during a step ($\dist(p(t)) > 0$), and original
physical parameters (e.g., a gait in $\G$ with $\omega = 1$).  These are common
equality and inequality constraints encountered in the literature
\cite{Ramezani2013, Hereid2018, Posa2016, Bessonnet2005, Westervelt2007}.

Figure~\ref{fig:marlo-w} shows the deformation of a gait starting from an EG of
$a = (0, 0.25) \in E$, where $a$ corresponds to a planar biped model in its
equilibrium stance ($\omega = 0$, Figure~\ref{fig:marlo-w}(a)), to a desired
gait of the original 3D model ($\omega = 1$) using the map $M_a$ and
Algorithm~\ref{alg:ghm}.  The Mathematica code found a desired gait after $42$~minutes and computed $58$ gaits.

%%%%%%%%%%%%%%%%%%%%%%%%%%%%%%%%%%%%%%%% FROST

\section{\reply{Comparison to FROST with IPOPT}}
\label{sec:frost}

\reply{We have demonstrated our approach using the examples in Section~\ref{sec:ex} and the bipeds in Figure~\ref{fig:models}.  In this section, we informally compare the numerical continuation approach to FROST \cite{Hereid2017} using IPOPT.  FROST is a Matlab/Mathematica-based library for modeling and simulating hybrid mechanical systems and transcribing them to nonlinear programming (NLP) problems.  The FROST library ships with IPOPT as the underlying solver for the NLP.  IPOPT uses interior point methods (a form of numerical continuation \cite{Nocedal1999}) to minimize a cost function subject to box constraints.  We view FROST as a library interface that simplifies the process of generating an NLP for bipedal robots (similar to our library) and IPOPT as a suite of algorithms that performs the search for an optimal gait (like our algorithms).}

\reply{Our library has several features in common with FROST.  For example, they both use Mathematica (MMA) to model and simulate rigid-body mechanical systems subject to impacts, PHCs, and VHCs; compute derivatives; and compile MMA expressions into C code.}
%We expect the C code that our library generates to be as fast as FROST's as we both use MMA's internal compilation tools.  We have not verified this assertion.}

\reply{In terms of distinct features, FROST can output its C code to be used in other environments such as Matlab.  A unique feature of our library is the ability to differentiate and compile user-defined algorithms written in a subset of MMA into C executables, including the Recursive Newton-Euler and Composite Rigid Body algorithms \cite{Featherstone2007}.  This approach scales better than FROST's approach, which symbolically computes the Euler-Lagrange equations \cite{Choset2005}.  As an example, on our test computer, FROST crashes after more than an hour of computing the mass matrix of the Atlas model in Section~\ref{ssec:bda}, while our approach can compile the hybrid equations of motion and their linearized dynamics to C in under 10~minutes.  
%The only objects we need to differentiate are the algorithms and their inputs, which largely consists of 6D spatial quantities.  
However, a limitation of our current implementation is MMA's memory management of its C code, which precludes compiling higher-order derivatives.}

\subsection{\reply{Generating Gaits for RABBIT}}
\reply{For comparison, we ported the FROST model of RABBIT (see Figure~\ref{fig:generic-biped}) and the corresponding NLP constraints into our framework. 
Briefly, the NLP locally minimizes the time-integral of summed and squared input torques subject to a desired average forward velocity and constraints on friction at the stance foot and the minimum step height at the middle of the step.   See the Multimedia Material for details.}

\reply{In order to represent the NLP, FROST uses direct collocation and generates 1270 variables and 1463 constraints to solve the problem at 21 nodes, where the time steps of the integration occur over the first 20 nodes and the last node is where the collision occurs.  The total time for NLP compilation to C and solving is approximately 82~seconds, with IPOPT requiring only 3 seconds, beginning from an initial seed chosen as the average of each decision variable's lower and upper bounds. }
%The optimal gait has a cost of 0.24~(N\,m)$^2$.}

\reply{We use a more compact representation of 19 decision variables (i.e., our state-time-control space is 19 dimensional) to represent the corresponding problem.  While we could have specified the same fixed-step integrator used in FROST, we let MMA automatically choose a variable-step ODE solver.  The total time for model compilation to C and solving using the GHM algorithm (Algorithm~\ref{alg:ghm}) is approximately 213~seconds.}

%Our main script takes 213~seconds to run when we generate our C code from scratch.  We satisfy the constraints across two runs of Algorithm~\ref{alg:ghm}.  The runs take 12.8 and 57~seconds to run, respectively.  We satisfy the constraints across two runs because the step-height-related constraints result in an ill-conditioned matrix when computing a descent direction at our chosen equilibrium point.}
%The cost of the final gait with respect to FROST's objective function is 4.1~(N\,m)$^2$, which we did not try to explicitly minimize.}

\reply{The comparison is not direct, as FROST/IPOPT optimizes over a small number of collocation points while Algorithm~\ref{alg:ghm} is a continuation method using the global homotopy map.  Still, FROST leverages the mature, optimized solver IPOPT for computational efficiency, while Algorithm~\ref{alg:ghm} is not optimized. In general, when FROST/IPOPT succeeded in finding gaits, the search time was shorter than using the GHM method, but we were unable to find feasible gaits for MARLO and Atlas using FROST.}

\reply{Future work could focus on the efficiency of Algorithm~\ref{alg:ghm}, but computational efficiency is not the purpose of this paper.  The focus of this paper is to show that entire families of biped gaits can be systematically constructed from simple initial seeds:  equilibrium stances.}

\section{Conclusion}
\label{sec:conc}
We present a robust method for generating large families of walking gaits for
high-degree-of-freedom bipeds using numerical continuation methods.  Our
approach differs from other gait-generation algorithms in that we transform the
problem of gait generation to mapping a level set in a state-time-control space.
A major advantage is that a one-footed equilibrium stance suffices as a seed to
find nearby walking gaits for bipeds subject to physical and virtual holonomic
constraints.  Furthermore, we prove that
the dimension of the search space for an initial gait is one dimensional and is
independent of the number of states and controls of the biped.  We have applied this approach to find
both passive and actuated gaits for a wide variety of simulated bipeds,
including the compass-gait, Atlas, and MARLO discussed in this paper.  

%We preserve the ``natural'' dynamics of the hybrid system producing passive dynamic walking gaits.  We can also incorporate control and design parameters, which can generate a diverse set of periodic walking motions (e.g., actuated and 3D gaits) using virtual holonomic constraints.  As examples, we show how our algorithm can generate numerous gaits for the compass-gait, Atlas, and MARLO bipeds.

% CUT FOR LENGTH
\begin{comment}
A few benefits of the algorithms presented in this paper, unlike other
gait-generation routines, are that we do not rely on randomly generated seed
values, a particular integration scheme to solve for the hybrid dynamics (as is
the case with direct collocation methods), and can reliably generate gaits using
only first-order derivative information of the hybrid dynamics (which can be a
useful feature when computing the Hessian is prohibitively expensive, as can be
the case for high-degree-of-freedom humanoid robots).  A free implementation of
our framework can be found online \cite{Rosa2014}.  It can be used with any
biped model subject to physical and virtual holonomic constraints that can be
represented as a kinematic tree of rigid bodies \cite{Featherstone2007}.
Equilibria of the robots serve as reliable seeds for finding entire families of
walking gaits.  
\end{comment}
% CUT FOR LENGTH

%%%%%%%%%%%%%%%%%%%%%%%%%%%%%%%%%%%%%%%% Thanks
\section*{Acknowledgment}
%\appendix
We thank Jian Shi, Zack Woodruff, Paul Umbanhowar, and the reviewers for their feedback on this paper.

\begin{comment}
\section*{Index to Multimedia Material}
%\appendix
\begin{tabular}{ccc}
\hline
Material & Media Type & Description \\ 
\hline
TRO2020CompanionVideo & .mp4 & Video animation of all gaits in paper \\
Code/ & .nb & Framework library and biped models \\
Code/ & .js/.html & Node.js gait animation library \\
\hline
\end{tabular}
\end{comment}

%%%%%%%%%%%%%%%%%%%%%%%%%%%%%%%%%%%%%%%% Appendix

\appendices

%%%%%%%%%%%%%%%%%%%%%%%%%%%%%%%%%%%%%%%% Appendix - Hybrid Dynamics

\section{Bipeds as Constrained Mechanical Systems}
\label{app:model}
In Section~\ref{sec:mp}, we defined the hybrid dynamics $\Sigma$ as the tuple
$\Sigma = (\X, f, \Delta, \phi)$, where $\fun{f}{\X \times \Ru}{T\X}$ is a
vector field on $\X$, $\fun{\Delta}{\X}{\X}$ is a jump map mapping pre- to
post-impact states, and $\fun{\phi}{\R \times \X}{\R}$ is a switching function
which determines when an impact occurs.  The goal of this section is to derive
$f$ and $\Delta$ in terms of the Euler-Lagrange equations for constrained
mechanical systems \cite{Liu2002, Aghili2005, Choset2005}.

%%%%%%%%%%%%%%%%%%%%%%%%%%%%%%%%%%%%%%%% Appendix - Vector Field

\subsection{The Vector Field $f$}
We model the continuous dynamics of an $n$-degree-of-freedom biped as a
constrained mechanical system subject to $n_p$ PHCs and $n_v$ VHCs ($n_p, n_v
\geq 0$).  For a biped's continuous dynamic regime, we assume that the state of
the biped $x = (q, \dot{q}) \in \X$ is known and that the biped is subject to a
set of physical and virtual holonomic constraints $h_p(q) = 0 \in \Rp$ and
$h_v(q) = 0 \in \Rv$, respectively.  The state $x$, accelerations $\ddot{q} \in
\R^n$, constraint forces $\lambda \in \R^{n_p}$, and control inputs $u \in \Ru$
($n_u \geq n_v$) of the continuous dynamics satisfy
\begin{equation}
\begin{gathered}
M(q) \ddot{q} + b(q, \dot{q}) = J_p^T(q) \lambda + B_v(q) u, \\
\dot{J}_p(q) \dot{q} + J_p(q) \ddot{q} = 0, \quad
\dot{J}_v(q) \dot{q} + J_v(q) \ddot{q} = -v(q, \dot{q}),
\end{gathered}
\label{eq:cd}
\end{equation}
where $M(q) \in \R^{n \times n}$ is the mass matrix, $b(q, \dot{q}) \in \Rq$ is
a vector of the centrifugal, Coriolis, and gravitational forces, $B_v(q) \in
\R^{n \times n_u}$ is a transmission matrix, $J_p(q) = \pd{h_p}{q}(q) \in
\R^{n_p \times n}$ and $J_v(q) = \pd{h_v}{q}(q) \in \R^{n_v \times n}$ are the
constraint Jacobian for the physical and virtual constraints, respectively, and
$v(q, \dot{q}) \in \R^{n_v}$ is a linear feedback controller for stabilizing the
virtual constraints.  An example PD control law for $v(q, \dot{q})$ used in
\cite{Westervelt2007, Hereid2018} is
\begin{equation*}
\label{eq:v}
v(q, \dot{q}) = \frac{1}{\epsilon} K_D J_v(q) \dot{q} + \frac{1}{\epsilon^2} K_P
h_v(q),
\end{equation*}
where $h_v(q) \in \Rv$ are VHCs, $K_P, K_D \in \R^{n_v \times n_v}$ are positive-definite matrices and $\epsilon \in \R$ is a positive scalar tuning parameter
which can speed up convergence to the origin, $h_v(q) = J_v(q) \dot{q} = 0$.

Given Equation~\eqref{eq:cd}, the vector field $f$ is then $f(x, u) =
(\dot{q}, \ddot{q})$, where $\ddot{q}$ is the solution to 
\begin{equation*}
\label{eq:cms}
\begin{bmatrix} 
M(q) & -J_p^T(q) & -B(q) \\
J_p(q) & 0 & 0 \\
J_v(q) & 0 & 0 
\end{bmatrix} 
\begin{bmatrix} \ddot{q} \\ \lambda \\ u \end{bmatrix} 
= - \begin{bmatrix} b(q, \dot{q}) \\ 0 \\ v(q, \dot{q}) + J_v(q) \dot{q}
\end{bmatrix},
\end{equation*}
which is a linear system of equations with $n+\npv$ equations in $n+\npu$
unknowns.  A solution exists with a generalized right inverse (we use the
Moore-Penrose inverse) as long as the matrix on the left-hand side has maximal
rank $n+\npv$.

\begin{myrem}
The differential form of the virtual constraints $\dot{J}_v(q) \dot{q} + J_v(q)
\ddot{q} = -v(q, \dot{q})$ can model virtual holonomic \emph{and} nonholonomic
constraints.  The same is true of the differential form of the physical
constraints.
\end{myrem}

\begin{myrem}
A linear stabilizing feedback controller $v_p(q, \dot{q})$ can also be defined
for the $n_p$ physical constraints such that $\dot{J}_p(q) \dot{q} + J_p(q)
\ddot{q} = -v_p(q, \dot{q})$ in order to reduce constraint violations during
simulations due to numerical rounding errors.  For example, we can implement
Baumgarte's constraint stabilization technique \cite{Bauchau2008} using the form
of the feedback law for $v(q, \dot{q})$.  Other options also exist
\cite{Chiou1998}.
\end{myrem}

\begin{myrem}
The solution to $u(t)$ of Equation~\eqref{eq:cd} is equivalent to the solution of
a stabilizing feedback linearizing control law used to enforce VHCs in the
Hybrid Zero Dynamics framework \cite{Westervelt2007} with $v(q, \dot{q})$ as the
linear control law.
\end{myrem}

%%%%%%%%%%%%%%%%%%%%%%%%%%%%%%%%%%%%%%%% Appendix - Jump Map

\subsection{The Jump Map $\Delta$}
We model collisions as a set of impulsive algebraic equations, namely the
impulse-momentum equations in generalized coordinates along with a set of
plastic impact equations needed to uniquely solve for the impulse and
post-impact velocities.  During a collision at time $t \in \R$, we assume the
pre-impact state of the biped $x(t^-) = (q, \dot{q}) \in \X$ is known.  The
pre-impact state $x(t^-)$, post-impact state $x(t^+) = (q^+, \dot{q}^+) \in \X$,
and impulses $\iota \in \R^{n_\iota}$ of the impulse equations satisfy
\begin{equation*}
\begin{aligned}
q^+ &= q, &
M(q) (\dot{q}^+ - \dot{q}) &= J_\iota^T(q) \iota, &
J_\iota(q) \dot{q}^+ &= 0,
\end{aligned}
\label{eq:ie}
\end{equation*}
where $J_\iota \in \R^{n_\iota \times n}$ is the constraint Jacobian that maps
the post-impact velocities to contact velocities.

The jump map $\Delta$ is then $\Delta(x) = (q, \dot{q}^+)$, where $\dot{q}^+$ is
the solution to 
\begin{equation*}
\begin{bmatrix} 
M(q) & -J_\iota^T(q) \\
J_\iota(q) & 0
\end{bmatrix} 
\begin{bmatrix} \dot{q}^+ \\ \iota \end{bmatrix} 
= \begin{bmatrix} M(q) \dot{q} \\ 0
\end{bmatrix},
\end{equation*}
which is a linear system of equations with $n+n_\iota$ equations in $n+n_\iota$
unknowns.  A unique solution exists as long as $J_\iota$ has maximal rank
$n_\iota$.

For most bipeds, $J_\iota = J_p$.  In general, any $J_\iota$ is fine as long as
the set of jump map constraints are more restrictive than the PHC constraints,
i.e., $\{x \in \X:  J_\iota(q) \dot{q} \} \subseteq \{x \in \X: J_p(q)
\dot{q}\}$.

%%%%%%%%%%%%%%%%%%%%%%%%%%%%%%%%%%%%%%%% Appendix - Additional Proofs

\section{Proofs of the Equilibrium Branches of $M_0$}
\label{app:ep}

\begin{proof}[Proof of Proposition~\ref{prop:reg}]
It follows from the implicit function theorem (IFT, \cite{Spivak1965}) that
there exists a unique curve $c$ passing through $c_0$.  What remains to be shown
is that the points on the curve are all in $E_0$.  From the IFT, we conclude
that in an open neighborhood $A \subseteq \R$ containing $\tau_0$ and an open
neighborhood $B \subseteq \R^{2n+k}$ containing the pair $(\xeq, \mu_0)$ that
for each $\tau(s) \in A$ there exists a unique $g(\tau(s)) \in B$ such that
$g(\tau) = (x_0(\tau), \mu_0(\tau))$, $c(s) = (x_0(\tau(s)), \tau(s),
\mu_0(\tau(s)))$ and $M_0(c(s)) = 0$ for some parameterization of $s \in
(-\delta, \delta)$.

We determine $g$ from the Jacobian $J_0$ of $M_0$
\begin{equation}
J_0(c) = \begin{bmatrix} \pd{P}{c}(c) \\ \pd{\Phi_0}{c}(c) \end{bmatrix}  =
\begin{bmatrix} 
\pd{P}{x_0}(c) & \pd{P}{\tau}(c) & \pd{P}{\mu}(c) \\
0 & 0 & \id_k
\end{bmatrix}.
\end{equation}
Given that $\xeq$ is an equilibrium point, we have at $J(c_0)$
\begin{equation*}
\pd{P}{\tau}(p(s_0)) = f(\xeq, u(\tau(s_0)))\pd{\tau}{c}(p(s_0)) = 0.
\end{equation*}
Additionally, because $c_0$ is a regular point of $M_0$, the submatrix
\begin{equation*}
\bar{J} = \begin{bmatrix} \pd{P}{x_0}(c_0) & \pd{P}{\mu}(c_0) \\
0 & \id_k \end{bmatrix}
\end{equation*}
must be full rank, $\det(\bar{J}) \neq 0$, or else $J_0$ cannot have maximal
rank and $c_0$ would be a singular point, which it is not.  From these two
facts, the IFT states that $g$ must be the solution to the IVP
\begin{equation*}
\begin{gathered}
g(\tau_0) = (\xeq, \mu_0), \\
\pd{g}{\tau}(\tau) = -\begin{bmatrix} \pd{P}{x_0}(p(s_0)) & \pd{P}{\mu}(p(s_0)) \\
0 & \id_k \end{bmatrix}^{-1} \begin{bmatrix} \pd{P}{\tau}(p(s_0)) \\ 0
\end{bmatrix} = 0,
\end{gathered}
\end{equation*}
which gives $g(\tau(s)) = (\xeq, \mu_0)$ for all $\tau(s) \in \R$.

To finish describing the curve $c$, we now determine an expression for $\tau =
\tau(s)$ valid for $s \in (-\delta, \delta)$.  Given an arclength
parameterization of the curve $c$, the tangent to the curve $c$ at $c_0$ is
$\sd{c}{s}(0)$ and it is the only vector in the tangent space $T_{c_0} M_0$.  As
the tangent space is equal to the null space of $J_0(c_0)$, we have 
\begin{equation*}
\begin{aligned}
\Null(J_0(c_0)) &= \Null\left(\begin{bmatrix} \pd{P}{x_0}(c) & 0 & \pd{P}{\mu}(c) \\ 0
& 0 & \id_k\end{bmatrix}\right)  \\
                &= \{\begin{bmatrix} 0_{2n}, & 1, & 0_k \end{bmatrix}^T\},
\end{aligned}
\end{equation*}
where $0_k$ and $0_{2n}$ are vectors with $k$ and $2n$ zeros, respectively.   Hence,
$\sd{\tau}{s}(s) = 1$ and for $\tau(0) = \tau_0$, $\tau(s) = \tau_0 + s$.

Therefore, $c(s) = (\xeq, \tau_0 + s, \mu_0)$ for $s \in (-\delta, \delta)$,
which are all in $E_0$ and by the IFT are the only points in a neighborhood
containing $c_0$ that are in $\G_0$.
\end{proof}

\begin{proof}[Proof of Proposition~\ref{prop:sing}]
We prove the first claim by showing that the path $p$ can never have points in
$\G_0 - E_0$ if EGs on the path are regular points of $M_0$.  The second claim
is proven through a direct computation.

Assume there exists a path $p$ with $p(0) \in E_0$ and $p(1) \in \G_0 - E_0$
such that all EGs in $p$ are regular points of $M_0$.  Then by
Proposition~\ref{prop:reg} because $c_0 \in E_0$ is a regular point of $M_0$,
any path $\fun{p}{[0, 1]}{\G_0}$ starting at $c_0$ must coincide locally
with the unique curve $\fun{c}{(-\delta, \delta)}{E_0}$.  As the functions $p$
and $c$ coincide but have different domains, assume that $s \in [0, s_\delta)$
continuously maps to $\alpha(s) \in (-\delta, \delta)$ such that for $s = 0$ we
have $\alpha(0) = 0$.

Consider $p$ at $s = s_\delta$.  Because $p$ is continuous $p(s_\delta)$ must be
equal to the left-sided limit of $p$ at $s_\delta$, that is, $p(s_\delta) =
\lim_{s \nearrow s_\delta} (\xeq, \tau_0 + \alpha(s), \mu_0) = (\xeq, \tau_0 +
\alpha(s_\delta), \mu_0)$.  Therefore, $p(s_\delta)$ is an EG in $E_0$, but by
assumption it is also a regular point of $M_0$, so we can apply
Proposition~\ref{prop:reg} and conclude that the interval over which $p$ and $c$
have to coincide extends beyond $s \in (0, s_\delta)$ for the path $p$.

In other words, the unique curve $c$ passing through $c(0)$ can be extended past
the open interval $(-\delta, \delta)$.  In fact, there is no finite value of
$\delta$ that can contain the entire interval of $c$ as its limit points
$c(\delta) = p(s_\delta) \in E_0$ are regular points of $M_0$.  The curve can
always be extended and as the curve $c$ is unique, the path $p$ has no choice
but to follow it.  The path $p$ however is finite with its points determined by
$c$ over its finite interval, but, for $s \in [0, 1]$, $p(s) \in E_0$.  This
contradicts our main assumption that $p(1) \in \G_0 - E_0$ given that all EGs
are regular points of $M_0$.

Therefore, there must exist at least one singular EG on the path $p$.  In order
for a point $p(s) \in E_0$ in the path $p$ to be singular, we need the submatrix
$\bar{J}$ of the Jacobian $J_0$ evaluated at $p(s)$
\begin{equation*}
\bar{J}(p(s)) =
\begin{bmatrix} \pd{P}{x_0}(p(s)) & \pd{P}{\mu}(p(s)) \\
0 & \id_k \end{bmatrix}
\end{equation*}
to not be invertible so that the IFT (and Proposition~\ref{prop:reg}) does not
apply.  For any $p(s)$ that is a singular EG of $M_0$, this can only happen if
\begin{equation*}
\det\left(\begin{bmatrix} \pd{P}{x_0}(c_s) & \pd{P}{\mu}(c_s) \\
0 & \id_k \end{bmatrix}\right) = \det\left(\pd{P}{x_0}(c_s)\right) = 0.
\end{equation*}
\end{proof}

\begin{proof}[Proof of Proposition~\ref{prop:curves}]
Given that $c_0$ is an isolated singular point in $E_0$, we must have that there
are exactly $\dim(T_{c_0} \G_0) = 2$ curves that intersect transversally at
$c_0$.  Otherwise, we would conclude that $c_0$ is not isolated or $\dim(T_{c_0}
\G_0) \neq 2$, which would be a contradiction.  Hence, it suffices to study
the tangent space $T_{c_0} \G_0$ to locally determine the gaits on each curve
near $c_0$.

From Equation~\eqref{eq:TcG}, a basis for the tangent space $T_{c_0} \G_0$ is
$T_{c_0} \G_0 = \Null(J_0(c_0))$, where $J_0$ is the Jacobian of $M_0$
(Equation~\eqref{eq:J0}).  Because $c_0$ is a singular EG, we have from
Corollary~\ref{cor:reg} and Proposition~\ref{prop:sing} that $\pd{P}{\tau}(c_0)
= 0$ and $\det(\pd{P}{x_0}(c_0)) = 0$.  This implies that the null space of
$J_0(c_0)$ has at least two linearly independent tangent vectors and, by
assumption of the above proposition, it has exactly two.  As $\pd{P}{\tau}(c_0)
= 0$, the Jacobian $J_0$ evaluated at $c_0$ has the form
\begin{equation*}
J_0(c_0) = \begin{bmatrix} \pd{P}{x_0}(c_0) & 0 & \pd{P}{\mu}(c_0) \\ 0
& 0 & \id_k\end{bmatrix}.
\end{equation*}
Assuming coordinates $\xtu$, a basis for the null space is $\Null(J_0(c_0)) =
\{e_0, g_0\}$, where $e_0 = [0_{2n}, 1, 0_k]^T \in \Rc$, $g_0$ satisfies
$g_0^Te_0 = 0$, and $0_{a}$ is a row vector of $a$ zeros.  The tangent vector
$e_0$ is in $T E_0$.  A curve $c_1$ with tangent vector $\sd{c_1}{s}(0) = e_0$
at $c_1(0) = c_0$ can only have points of the form $c_1(s) = (\xeq, \tau_0 + s,
\mu_0)$ for $s \in [-\epsilon, \epsilon]$.  The tangent vector $g_0$ is
orthogonal to $e_0$, hence a curve $c_2$ with $c_2(0) = c_0$ and $\sd{c_2}{s}(0)
= g_0$ has gaits $c_2(s)$ in $\G_0 - E_0$ for small values of $s \neq 0$.
\end{proof}

%%%%%%%%%%%%%%%%%%%%%%%%%%%%%%%%%%%%%%%% Appendix - Debugging Errors

\section{Debugging Error Messages from Algorithm~\ref{alg:gm}}
\label{app:err}
We present a list of potential sources of errors and solutions in the Multimedia Material when a call to Algorithm~\ref{alg:gm} results in an error state.  The issues covered are when

\begin{enumerate}
    \item the plot of $I(\tau)$ is the constant zero line $I(\tau) = 0$, and
    \item the plot has no zero crossings $I(\tau) \neq 0$.
\end{enumerate}

\bibliographystyle{IEEEtran}
\bibliography{MyBib}

\begin{thebibliography}{10}
\providecommand{\url}[1]{#1}
\csname url@samestyle\endcsname
\providecommand{\newblock}{\relax}
\providecommand{\bibinfo}[2]{#2}
\providecommand{\BIBentrySTDinterwordspacing}{\spaceskip=0pt\relax}
\providecommand{\BIBentryALTinterwordstretchfactor}{4}
\providecommand{\BIBentryALTinterwordspacing}{\spaceskip=\fontdimen2\font plus
\BIBentryALTinterwordstretchfactor\fontdimen3\font minus
  \fontdimen4\font\relax}
\providecommand{\BIBforeignlanguage}[2]{{%
\expandafter\ifx\csname l@#1\endcsname\relax
\typeout{** WARNING: IEEEtran.bst: No hyphenation pattern has been}%
\typeout{** loaded for the language `#1'. Using the pattern for}%
\typeout{** the default language instead.}%
\else
\language=\csname l@#1\endcsname
\fi
#2}}
\providecommand{\BIBdecl}{\relax}
\BIBdecl

\bibitem{Gregg2012limping}
R.~D. Gregg, Y.~Y. Dhaher, A.~Degani, and K.~M. Lynch, ``On the mechanics of
  functional asymmetry in bipedal walking,'' \emph{IEEE Transactions on
  Biomedical Engineering}, vol.~59, no.~5, pp. 1310--1318, May 2012.

\bibitem{Goswami1998}
A.~Goswami, B.~Thuilot, and B.~Espiau, ``A study of the passive gait of a
  compass-like biped robot: {S}ymmetry and chaos,'' \emph{Int. J. Robotics
  Research}, vol.~17, no.~12, pp. 1282--1301, 1998.

\bibitem{Garcia1998}
M.~Garcia, A.~Chatterjee, A.~Ruina, and M.~Coleman, ``The simplest walking
  model: {Stability}, complexity, and scaling,'' \emph{ASME J. Biomechanical
  Engineering}, vol. 120, no.~2, pp. 281--288, 1998.

\bibitem{Westervelt2007}
E.~R. Westervelt, J.~W. Grizzle, C.~Chevallereau, J.~H. Choi, and B.~Morris,
  \emph{Feedback control of dynamic bipedal robot locomotion}.\hskip 1em plus
  0.5em minus 0.4em\relax CRC press Boca Raton, 2007.

\bibitem{Crossley2005}
M.~D. Crossley, \emph{Essential Topology}.\hskip 1em plus 0.5em minus
  0.4em\relax Springer-Verlag, 2005.

\bibitem{Goswami1999}
A.~Goswami, ``Postural stability of biped robots and the foot-rotation
  indicator {(FRI)} point,'' \emph{Int. J. Robotics Research}, vol.~18, no.~6,
  pp. 523--533, 1999.

\bibitem{Bessonnet2005}
\BIBentryALTinterwordspacing
G.~Bessonnet, P.~Seguin, and P.~Sardain, ``A parametric optimization approach
  to walking pattern synthesis,'' \emph{The International Journal of Robotics
  Research}, vol.~24, no.~7, pp. 523--536, Jul 2005. [Online]. Available:
  \url{http://dx.doi.org/10.1177/0278364905055377}
\BIBentrySTDinterwordspacing

\bibitem{Grizzle2014}
J.~W. Grizzle, C.~Chevallereau, R.~W. Sinnet, and A.~D. Ames, ``Models,
  feedback control, and open problems of {3D} bipedal robotic walking,''
  \emph{Automatica}, vol.~50, no.~8, pp. 1955--1988, Aug 2014.

\bibitem{Xi2015}
W.~Xi, Y.~Yesilevskiy, and C.~D. Remy, ``Selecting gaits for economical
  locomotion of legged robots,'' \emph{The International Journal of Robotics
  Research}, Nov 2015.

\bibitem{Chen2007}
V.~E.~H. Chen, ``Passive dynamic walking with knees: A point foot model,''
  Master's thesis, Massachusetts Institute of Technology, 2007.

\bibitem{Hereid2018}
A.~Hereid, C.~M. Hubicki, E.~A. Cousineau, and A.~D. Ames, ``Dynamic humanoid
  locomotion: A scalable formulation for {HZD} gait optimization,''
  \emph{{IEEE} Transactions on Robotics}, vol.~34, no.~2, pp. 370--387, apr
  2018.

\bibitem{Posa2016}
M.~Posa, S.~Kuindersma, and R.~Tedrake, ``Optimization and stabilization of
  trajectories for constrained dynamical systems,'' in \emph{Proceedings of the
  International Conference on Robotics and Automation (ICRA)}, May 2016.

\bibitem{Gregg2012}
R.~D. Gregg, A.~K. Tilton, S.~Candido, T.~Bretl, and M.~W. Spong, ``Control and
  planning of 3-d dynamic walking with asymptotically stable gait primitives,''
  \emph{{IEEE} Transactions on Robotics}, vol.~28, no.~6, pp. 1415--1423, dec
  2012.

\bibitem{Motahar2016}
M.~S. Motahar, S.~Veer, and I.~Poulakakis, ``Composing limit cycles for motion
  planning of 3d bipedal walkers,'' in \emph{2016 {IEEE} 55th Conference on
  Decision and Control ({CDC})}.\hskip 1em plus 0.5em minus 0.4em\relax {IEEE},
  dec 2016.

\bibitem{Liu2012a}
C.~Liu, C.~G. Atkeson, and J.~Su, ``Biped walking control using a trajectory
  library,'' \emph{Robotica}, vol.~31, no.~2, pp. 311--322, May 2012.

\bibitem{Saglam2014}
C.~O. Saglam and K.~Byl, ``Robust policies via meshing for metastable rough
  terrain walking,'' in \emph{Proceedings of Robotics: Science and Systems},
  Berkeley, USA, July 2014.

\bibitem{Rosa2014a}
N.~Rosa and K.~M. Lynch, ``Extending equilibria to periodic orbits for walkers
  using continuation methods,'' \emph{2014 IEEE/RSJ International Conference on
  Intelligent Robots and Systems}, Sep 2014.

\bibitem{Rosa2017}
------, ``Using equilibria and virtual holonomic constraints to generate
  families of walking gaits,'' in \emph{Dynamic Walking Conference}, Mariehamn,
  Finland, Jun. 2017.

\bibitem{Gomes2005a}
M.~W. Gomes, ``Collisionless rigid body locomotion models and physically based
  homotopy methods for finding periodic motions in high degree of freedom
  models,'' Ph.D. dissertation, Cornell University, 2005.

\bibitem{McGeer1990a}
T.~McGeer, ``Passive dynamic walking,'' \emph{Int. J. Robotics Research},
  vol.~9, no.~2, pp. 62--82, 1990.

\bibitem{Chatterjee2000}
A.~Chatterjee and M.~Garcia, ``Small slope implies low speed for {McGeer}'s
  passive walking machines,'' \emph{Dynamics and Stability of Systems},
  vol.~15, no.~2, pp. 139--157, 2000.

\bibitem{Garcia2000}
M.~Garcia, A.~Chatterjee, and A.~Ruina, ``Efficiency, speed, and scaling of
  two-dimensional passive-dynamic walking,'' \emph{Dynamics and Stability of
  Systems}, vol.~15, no.~2, pp. 75--99, Jun 2000.

\bibitem{Rosa2012}
N.~Rosa, A.~Barber, R.~D. Gregg, and K.~M. Lynch, ``Stable open-loop
  brachiation on a vertical wall,'' in \emph{IEEE International Conference on
  Robotics and Automation}, May 2012, pp. 1193--1199.

\bibitem{Rosa2013}
N.~Rosa and K.~Lynch, ``The passive dynamics of walking and brachiating robots:
  Results on the topology and stability of passive gaits,'' in
  \emph{Nature-Inspired Mobile Robotics: Proceedings of the 16th International
  Conference on Climbing and Walking Robots and the Support Technologies for
  Mobile Machines}, 2013.

\bibitem{Krauskopf2007}
B.~Krauskopf, H.~Osinga, and J.~Galan-Vioque, \emph{Numerical Continuation
  Methods for Dynamical Systems: Path Following and Boundary Value
  Problems}.\hskip 1em plus 0.5em minus 0.4em\relax Springer, 2007.

\bibitem{Griffin2015}
\BIBentryALTinterwordspacing
B.~Griffin and J.~Grizzle, ``Nonholonomic virtual constraints for dynamic
  walking,'' \emph{2015 54th IEEE Conference on Decision and Control (CDC)},
  Dec 2015. [Online]. Available:
  \url{http://dx.doi.org/10.1109/CDC.2015.7402850}
\BIBentrySTDinterwordspacing

\bibitem{Hamed2016}
K.~A. Hamed, B.~G. Buss, and J.~W. Grizzle, ``Exponentially stabilizing
  continuous-time controllers for periodic orbits of hybrid systems:
  Application to bipedal locomotion with ground height variations,'' \emph{The
  International Journal of Robotics Research}, vol.~35, no.~8, pp. 977--999,
  Jul 2016.

\bibitem{Chevallereau2003}
C.~Chevallereau, G.~Abba, Y.~Aoustin, F.~Plestan, E.~R. Westervelt,
  C.~{Canudas-de-Wit}, and J.~W. Grizzle, ``Rabbit: A testbed for advanced
  control theory,'' \emph{IEEE Control Systems Magazine}, vol.~23, no.~5, pp.
  57--79, 2003.

\bibitem{Ramezani2013}
\BIBentryALTinterwordspacing
A.~Ramezani, J.~W. Hurst, K.~Akbari~Hamed, and J.~W. Grizzle, ``Performance
  analysis and feedback control of {ATRIAS}, a three-dimensional bipedal
  robot,'' \emph{Journal of Dynamic Systems, Measurement, and Control}, vol.
  136, no.~2, p. 021012, Dec 2013. [Online]. Available:
  \url{http://dx.doi.org/10.1115/1.4025693}
\BIBentrySTDinterwordspacing

\bibitem{Saglam2015}
C.~O. Saglam and K.~Byl, ``Meshing hybrid zero dynamics for rough terrain
  walking,'' in \emph{IEEE International Conference on Robotics and
  Automation}.\hskip 1em plus 0.5em minus 0.4em\relax IEEE, May 2015.

\bibitem{Rheinboldt2000}
W.~C. Rheinboldt, ``Numerical continuation methods: a perspective,''
  \emph{Journal of Computational and Applied Mathematics}, vol. 124, no. 1-2,
  pp. 229--244, Dec 2000.

\bibitem{Liu2012}
L.~Liu, K.~Yin, M.~van~de Panne, and B.~Guo, ``Terrain runner: Control,
  parameterization, composition, and planning for highly dynamic motions,''
  \emph{ACM Trans. Graph.}, vol.~31, no.~6, pp. 154:1--154:10, Nov. 2012.

\bibitem{Gan2018}
Z.~Gan, Y.~Yesilevskiy, P.~Zaytsev, and C.~D. Remy, ``All common bipedal gaits
  emerge from a single passive model,'' \emph{Journal of The Royal Society
  Interface}, vol.~15, no. 146, Sep. 2018.

\bibitem{Gill2002}
\BIBentryALTinterwordspacing
P.~E. Gill, W.~Murray, and M.~A. Saunders, ``{SNOPT}: An {SQP} algorithm for
  large-scale constrained optimization,'' \emph{SIAM Journal on Optimization},
  vol.~12, no.~4, pp. 979--1006, Jan 2002. [Online]. Available:
  \url{http://dx.doi.org/10.1137/S1052623499350013}
\BIBentrySTDinterwordspacing

\bibitem{Betts2010}
\BIBentryALTinterwordspacing
J.~T. Betts, \emph{Practical Methods for Optimal Control and Estimation Using
  Nonlinear Programming}, 2nd~ed.\hskip 1em plus 0.5em minus 0.4em\relax
  Society for Industrial \& Applied Mathematics (SIAM), Jan 2010. [Online].
  Available: \url{http://dx.doi.org/10.1137/1.9780898718577}
\BIBentrySTDinterwordspacing

\bibitem{Rosa2020}
N.~Rosa, ``Bipedal gait generation library,''
  \url{https://github.com/nr-codes/BipedalGaitGeneration}, 2020, last accessed
  6/03/20.

\bibitem{Spivak1965}
M.~Spivak, \emph{Calculus on Manifolds. {A} modern approach to classical
  theorems of advanced calculus}.\hskip 1em plus 0.5em minus 0.4em\relax W. A.
  Benjamin, Inc., New York-Amsterdam, 1965.

\bibitem{Allgower1990}
E.~Allgower and K.~Georg, \emph{Numerical Continuation Methods, An
  Introduction}.\hskip 1em plus 0.5em minus 0.4em\relax New York, NY:
  Springer-Verlag New York, Inc., 1990.

\bibitem{Choset2005}
H.~Choset, K.~M. Lynch, S.~Hutchinson, G.~A. Kantor, W.~Burgard, L.~E. Kavraki,
  and S.~Thrun, \emph{Principles of Robot Motion: Theory, Algorithms, and
  Implementations}.\hskip 1em plus 0.5em minus 0.4em\relax Cambridge, MA: MIT
  Press, June 2005.

\bibitem{Collins2005}
S.~H. Collins and A.~L. Ruina, ``A bipedal walking robot with efficient and
  human-like gait,'' in \emph{IEEE Int. Conf. on Robotics and Automation},
  Barcelona, Spain, 2005, pp. 1983--1988.

\bibitem{Bainov1993}
D.~Bainov and P.~Simeonov, \emph{Impulsive Differential Equations: Periodic
  Solutions and Applications (Monographs and Surveys in Pure and Applied
  Mathematics)}.\hskip 1em plus 0.5em minus 0.4em\relax Chapman and Hall/CRC,
  1993.

\bibitem{Press2002}
\BIBentryALTinterwordspacing
W.~H. Press, S.~A. Teukolsky, W.~T. Vetterling, and B.~P. Flannery,
  \emph{Numerical Recipes in C: The Art of Scientific Computing}, 2nd~ed.\hskip
  1em plus 0.5em minus 0.4em\relax Cambridge University Press, 2002. [Online].
  Available: \url{http://apps.nrbook.com/c/index.html}
\BIBentrySTDinterwordspacing

\bibitem{Henderson2002}
M.~E. Henderson, ``Multiple parameter continuation: Computing implicitly
  defined k-manifolds,'' \emph{International Journal of Bifurcation and Chaos},
  vol.~12, no.~03, pp. 451--476, Mar. 2002.

\bibitem{Nocedal1999}
J.~Nocedal and S.~J. Wright, \emph{Numerical Optimization}.\hskip 1em plus
  0.5em minus 0.4em\relax Springer Verlag, 1999.

\bibitem{Bertsekas1982}
D.~P. Bertsekas, ``Projected {Newton} methods for optimization problems with
  simple constraints,'' \emph{{SIAM} Journal on Control and Optimization},
  vol.~20, no.~2, pp. 221--246, Mar. 1982.

\bibitem{Kelley1999}
C.~Kelley, \emph{Iterative Methods For Optimization}.\hskip 1em plus 0.5em
  minus 0.4em\relax Society for Industrial and Applied Mathematics, 1999.

\bibitem{Bhounsule2014a}
P.~A. Bhounsule, ``Foot placement in the simplest slope walker reveals a wide
  range of walking solutions,'' \emph{IEEE Transactions on Robotics}, vol.~30,
  no.~5, pp. 1255--1260, 2014.

\bibitem{Keller1978}
\BIBentryALTinterwordspacing
H.~B. Keller, ``Global homotopies and {Newton} methods,'' \emph{Recent Advances
  in Numerical Analysis}, pp. 73--94, 1978. [Online]. Available:
  \url{http://dx.doi.org/10.1016/B978-0-12-208360-0.50009-7}
\BIBentrySTDinterwordspacing

\bibitem{Ben-Israel1966}
A.~Ben-Israel, ``A {Newton-Raphson} method for the solution of systems of
  equations,'' \emph{Journal of Mathematical Analysis and Applications},
  vol.~15, no.~2, pp. 243--252, aug 1966.

\bibitem{Martin2014}
A.~E. Martin, D.~C. Post, and J.~P. Schmiedeler, ``Design and experimental
  implementation of a hybrid zero dynamics-based controller for planar bipeds
  with curved feet,'' \emph{International Journal of Robotics Research},
  vol.~33, no.~7, pp. 988--1005, 2014.

\bibitem{LaHera2013}
P.~X.~M. La~Hera, A.~S. Shiriaev, L.~B. Freidovich, U.~Mettin, and S.~V. Gusev,
  ``Stable walking gaits for a three-link planar biped robot with one
  actuator,'' \emph{IEEE Transactions on Robotics}, vol.~29, no.~3, pp.
  589--601, Jun 2013.

\bibitem{Dumas2007}
R.~Dumas, L.~Cheze, and J.-P. Verriest, ``Adjustments to {McConville} et al.
  and {Young} et al. body segment inertial parameters,'' \emph{Journal of
  Biomechanics}, vol.~40, no.~3, pp. 543--553, 2007.

\bibitem{Chevallereau2008}
C.~Chevallereau, J.~W. Grizzle, and C.~Shih, ``Asymptotically stable walking of
  a five-link underactuated {3D} bipedal robot,'' \emph{IEEE Trans. Robotics},
  vol.~25, no.~1, pp. 37--50, 2008.

\bibitem{BostonDynamics2013}
{Boston Dynamics}, ``Atlas - {T}he {A}gile {A}nthropomorphic {R}obot,''
  \url{http://www.bostondynamics.com/robot_Atlas.html}, 2013, last accessed
  12/4/14.

\bibitem{Featherstone2007}
R.~Featherstone, \emph{Rigid Body Dynamics Algorithms}.\hskip 1em plus 0.5em
  minus 0.4em\relax Springer, 2007.

\bibitem{Iida2009}
F.~Iida and R.~Tedrake, ``Minimalistic control of a compass gait robot in rough
  terrain,'' in \emph{IEEE International Conference on Robotics and
  Automation}, May 2009, pp. 1985--1990.

\bibitem{OSRF2014}
{Open Source Robotics Foundation}, ``drc{\_}skeleton{\_}v3{\_}mod1.cfg,''
  {DARPA} Robotics Challenge Simulator, default branch commit dfb3482, June
  2014.

\bibitem{Hubicki2016}
C.~Hubicki, J.~Grimes, M.~Jones, D.~Renjewski, A.~Spr\"{o}witz, A.~Abate, and
  J.~Hurst, ``Atrias: Design and validation of a tether-free 3d-capable
  spring-mass bipedal robot,'' \emph{The International Journal of Robotics
  Research}, vol.~35, no.~12, pp. 1497--1521, 2016.

\bibitem{Griffin2016}
B.~Griffin, ``Nhvc3dsim\_bag160728.zip,''
  \url{http://www.griffb.com/s/NHVC3DSim_BAG160728.zip}, 2016.

\bibitem{Hereid2017}
\BIBentryALTinterwordspacing
A.~Hereid and A.~D. Ames, ``Frost: Fast robot optimization and simulation
  toolkit,'' in \emph{IEEE/RSJ International Conference on Intelligent Robots
  and Systems (IROS)}.\hskip 1em plus 0.5em minus 0.4em\relax Vancouver, BC,
  Canada: IEEE/RSJ, Sep. 2017. [Online]. Available:
  \url{https://github.com/ayonga/frost-dev/tree/549d2a6a89102779ebaa1565f0b2809617c99276}
\BIBentrySTDinterwordspacing

\bibitem{Liu2002}
G.~Liu and Z.~Li, ``A unified geometric approach to modeling and control of
  constrained mechanical systems,'' \emph{IEEE Transactions on Robotics and
  Automation}, vol.~18, no.~4, pp. 574--587, Aug. 2002.

\bibitem{Aghili2005}
F.~Aghili, ``A unified approach for inverse and direct dynamics of constrained
  multibody systems based on linear projection operator: Applications to
  control and simulation,'' \emph{IEEE Trans. on Robotics}, vol.~21, no.~5, pp.
  834--849, 2005.

\bibitem{Bauchau2008}
\BIBentryALTinterwordspacing
O.~A. Bauchau and A.~Laulusa, ``Review of contemporary approaches for
  constraint enforcement in multibody systems,'' \emph{Journal of Computational
  and Nonlinear Dynamics}, vol.~3, no.~1, 2008. [Online]. Available:
  \url{http://dx.doi.org/10.1115/1.2803258}
\BIBentrySTDinterwordspacing

\bibitem{Chiou1998}
J.~C. Chiou and S.~D. Wu, ``Constraint violation stabilization using
  input-output feedback linearization in multibody dynamic analysis,''
  \emph{Journal of Guidance, Control, and Dynamics}, vol.~21, no.~2, pp.
  222--228, Mar 1998.

\end{thebibliography}

%%%%%%%%%%%%%%%%%%%%%%%%%%%%%%%%%%%%%%%% Biographies
\vskip -2\baselineskip plus -1fil
\begin{IEEEbiography}
    [{\includegraphics[width=1in,height=1.25in,clip,keepaspectratio]{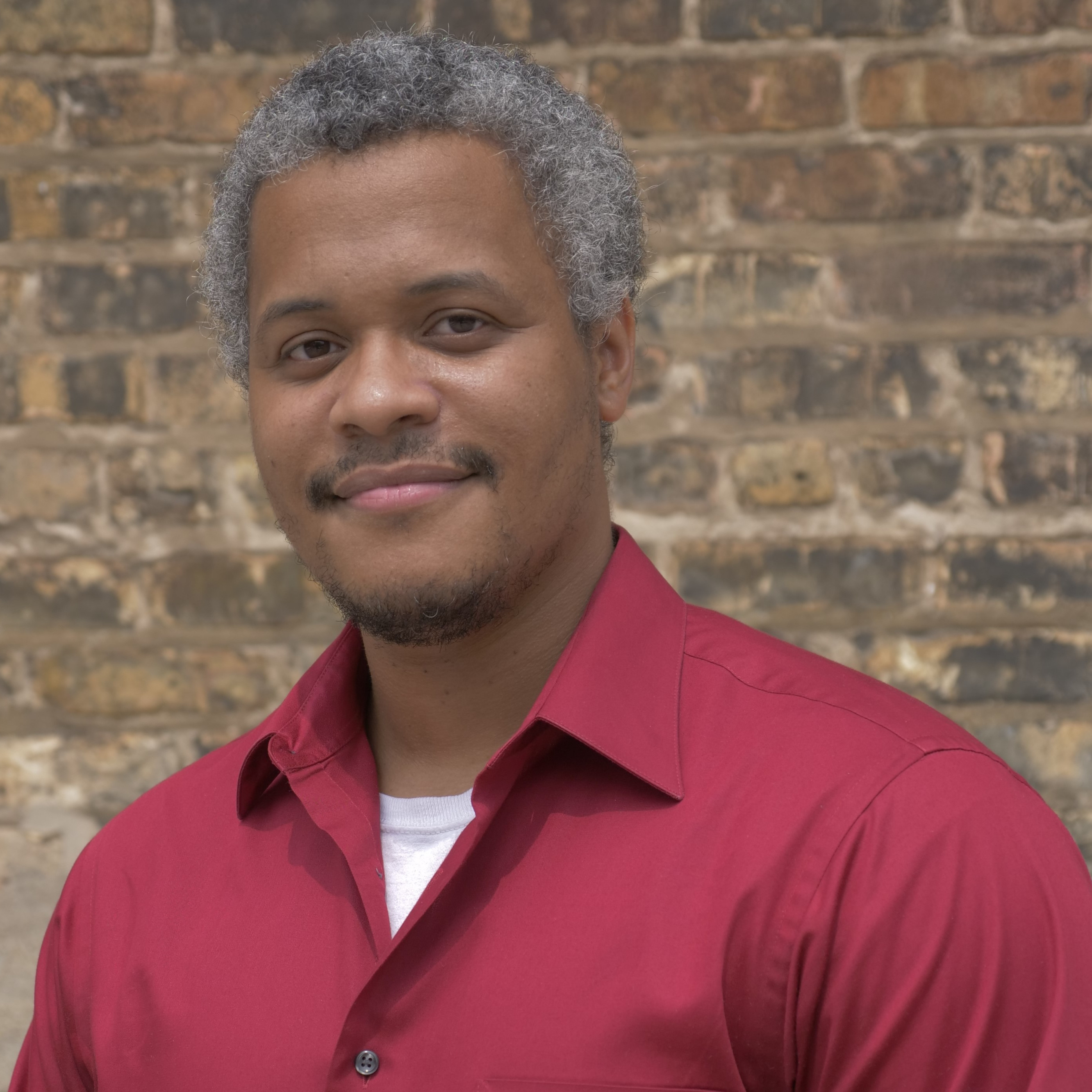}}]{Nelson Rosa Jr.}
    received the B.E. degree in engineering sciences from Dartmouth College, Hanover, NH, USA, in 2008, and the Ph.D. degree in mechanical engineering from Northwestern University, Evanston, IL, USA, in 2018.
    
    Dr. Rosa received the Alexander von Humboldt Postdoctoral Research Fellowship in 2021.  As a fellow, he is pursuing his research at the Institute of Nonlinear Mechanics at the University of Stuttgart, Stuttgart, Germany.  His research interests include hybrid dynamical systems, embedded/computer systems, algorithmic design, mechatronics, and legged robots.
\end{IEEEbiography}

\vskip -2\baselineskip plus -1fil

\begin{IEEEbiography}
    [{\includegraphics[width=1in,height=1.25in,clip,keepaspectratio]{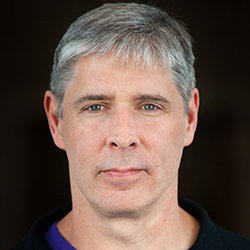}}]{Kevin M. Lynch}
    (S’90–M’96–SM’05–F’10) received the B.S.E. degree in electrical engineering from Princeton University, Princeton, NJ, USA, in 1989, and the Ph.D. degree in robotics from Carnegie Mellon University, Pittsburgh, PA, USA, in 1996.
    
    %He is currently a Professor and the Chair of the Department of Mechanical Engineering, Northwestern University, Evanston, IL, USA.
    He is the Director of the Northwestern Center for Robotics and Biosystems and a member of the Northwestern Institute on Complex Systems. He is a coauthor of the textbooks titled \textit{Principles of Robot Motion} (Cambridge, MA, USA: MIT Press, 2005), \textit{Embedded Computing and Mechatronics} (New York, NY, USA: Elsevier, 2015), and \textit{Modern Robotics: Mechanics, Planning, and Control} (Cambridge, U.K.: Cambridge Univ. Press, 2017). His research interests include dynamics, motion planning, and control for robot manipulation and locomotion; self-organizing multiagent systems; and human–robot systems.
    
    Dr. Lynch is the Editor-in-Chief of the IEEE TRANSACTIONS ON ROBOTICS.
\end{IEEEbiography}

\end{document}